\documentclass[12pt, dvipsnames]{rutgersthesis}

\newcommand{\thesisType}{Dissertation} % Dissertation for PhD and Thesis for Master
\newcommand{\thesisTitle}{\textbf{Efficiently Manipulating Clutter via Learning and Search-Based Reasoning}} % your thesis/dissertation title, e.g., How to Write A Thesis
\newcommand{\yourName}{\textbf{Baichuan Huang}} % your name, e.g., Tim Cook
\newcommand{\yourDegree}{Doctor of Philosophy} % your degree to be obtained, e.g., Master of Science, Doctor of Philosophy
\newcommand{\yourProgram}{Computer Science} %insert your graduate program’s official name
\newcommand{\yourDirector}{Jingjin Yu} % your thesis director name here, his/her title not included
\newcommand{\yourCommitteeNumber}{4} % IMPORTANT! the number of the committee members for your defense
\newcommand{\yourMonth}{May} % your graduation month, not your defense month
\newcommand{\yourYear}{2025} % your graduation year

\DeclareMathOperator*{\argmax}{arg\,max}

\SetKwProg{Fn}{Function}{}{}
\SetKwComment{Comment}{$\triangleright$\ }{}

\bibliography{references}

\begin{document}
% \doublespacing  %set line spacing

\clearpage
\begin{center}

\vspace*{\fill}

\copyright { }{\yourYear}\\
{\yourName}\\
ALL RIGHTS RESERVED\\

\vspace*{\fill}

\end{center}

\pagenumbering{gobble}
\clearpage

\makeTitlePage{Month}{Year}

\begin{frontmatter}
    \begin{my_abstract}

Object rearrangement is a fundamental and highly challenging problem in robotic manipulation, encompassing a diverse set of tasks such as clutter removal and object retrieval. These tasks require robots to intelligently plan and execute sequences of manipulation actions to reorganize objects or extract specific targets from cluttered environments. The significance of object rearrangement extends to numerous real-world applications, including warehouse automation, household assistance, and industrial manufacturing. However, solving this problem efficiently remains difficult due to the high-dimensional configuration spaces, intricate object interactions, and long planning horizons involved. The ability to develop more advanced solutions to these challenges is critical for enabling robots to operate autonomously in unstructured real-world settings.

This dissertation is motivated by the need for more efficient and robust manipulation planning approaches that can be adapted to dynamic and complex environments. To address these challenges, we propose a set of novel algorithms that leverage the strengths of search-based planning, deep learning, and parallelized computation. Our work focuses on improving the prediction of object interactions, integrating these predictions into tree search algorithms, and utilizing high-performance parallel computing to significantly accelerate the planning process.

Our research begins with the development of the Deep Interaction Prediction Network (DIPN), which enables accurate predictions of object motions when subjected to pushing actions. DIPN is trained to model object interactions with high precision, achieving over 90\% accuracy in predicting the final poses of objects after a push. This significantly surpasses existing baseline methods and allows for more reliable decision-making in cluttered environments. Building on this capability, we integrate DIPN with Monte Carlo Tree Search (MCTS) to optimize the planning of non-prehensile actions for object retrieval tasks. This integrated approach enables robots to autonomously determine effective sequences of push actions, leading to a 100\% completion rate in specific, well-defined challenging scenarios where heuristic-based solutions previously struggled.

To further improve computational efficiency, we introduce the Parallel Monte Carlo Tree Search with Batched Simulations (PMBS) framework. This framework leverages GPU-accelerated physics simulations to parallelize planning computations, achieving more than a $30\times$ speed-up compared to traditional serial implementations. Importantly, this acceleration does not compromise solution quality; PMBS maintains or even improves it, demonstrating its effectiveness for real-time robotic planning. Additionally, we combine different manipulation techniques, such as pick-and-place and push, to make our approach more flexible and adaptable to various tasks. By integrating diverse manipulation techniques, our system can tackle a wider range of object rearrangement challenges more effectively.

Extensive experiments conducted in both simulated environments and real-world robotic systems validate the efficacy of our proposed methods. Our findings demonstrate state-of-the-art performance in terms of success rates, solution quality, and computational efficiency across a variety of complex rearrangement tasks. By pushing the boundaries of robotic manipulation capabilities, this work contributes to the advancement of autonomous robotic systems, bringing us closer to deploying intelligent robots capable of handling complex object rearrangement tasks in real-world, unstructured environments.

\end{my_abstract}
    
\begin{acknowledgments}

I would like to express my deepest gratitude to my advisor, Prof. Jingjin Yu, for his inspiration and guidance throughout my research journey. Your expertise and mentorship were instrumental in shaping every aspect of my work during my five-year Ph.D. study, not only in academic matters but also in practical aspects such as building robust robotic systems and in guiding my career planning.
I am also profoundly thankful to my dissertation committee members. In particular, I would like to thank Prof. Abdeslam Boularias, who contributed significantly to much of my works and provided insightful ideas that greatly enhanced my research.
My sincere appreciation goes to all my committee members, not only for their valuable comments and suggestions on this dissertation but also for their broader impact on my academic growth. I am especially honored to have served as a teaching assistant for Prof. Kostas Bekris, from whom I learned invaluable lessons about effective teaching. I extend my heartfelt gratitude to Dr. Bowen Wen, a member of my committee, whose research findings have deeply influenced my own work and this dissertation.

I have been fortunate to collaborate, both formally and informally, with several of my lab members. I would particularly like to thank Shuai Han for his contributions to my first paper at Rutgers. To all my lab mates: thank you for your companionship and unwavering support during times of need.
I am grateful to the funding agencies that supported my work, as well as to my colleagues during my internships. Special thanks to Dr. Siddarth Jain and the team at Mitsubishi Electric Research Laboratories, where I gained invaluable industrial experience. I also appreciate the opportunities provided by Coupang, which further broadened my professional horizons.

Last but certainly not least, I want to express my heartfelt thanks to my parents for their unconditional love and support throughout this journey. Your encouragement has been my foundation and inspiration.

\end{acknowledgments}

    \makeTOC
    \makeListOfTables
    \makeListOfFigures
\end{frontmatter}

\begin{thesisbody}
    
\chapter{Introduction}\label{chap:introduction}
\thispagestyle{myheadings}

Robotic manipulation~\cite{mason2018toward} is a fundamental capability that enables various applications across many industries. In warehouse automation, robots are revolutionizing order fulfillment and inventory management~\cite{azadeh2019robotized}. In healthcare, robotic systems assist in surgeries, patient care, and laboratory tasks~\cite{taylor2016medical}. Human-robot collaboration is enhancing productivity in manufacturing and assembly lines~\cite{ajoudani2018progress}. In disaster response scenarios, search and rescue robots navigate hazardous environments to locate and assist survivors~\cite{jennings1997cooperative}. These diverse applications highlight the critical role of robotic manipulation.

Among these applications, clutter removal is an important task involving the grasping and extracting of objects from cluttered environments such as bins or conveyor belts~\cite{mahler2018dex}. Object retrieval is a closely related but distinct challenge, where the goal is to extract a specific target object from a cluttered scene. This capability is crucial for household robots tasked with retrieving items from a pile, where specific items need to be identified and retrieved~\cite{boroushaki2022fusebot}. 
Beyond removal and retrieval, object rearrangement~\cite{krontiris2015dealing} presents its own challenges, requiring robots to reorganize objects within a given space. Furthermore, the ability to manipulate objects in dynamic environments, where items may be moving or rolling, adds another layer of complexity to these tasks~\cite{marturi2019dynamic}. These challenges, illustrated in~\autoref{fig:intro-imgs}, all fall under the broader category of robot manipulation tasks and represent active areas of research.

\begin{figure}[ht]
    \centering
    \begin{minipage}[b]{0.7\textwidth}
        \centering
        \begin{subfigure}[b]{0.49\linewidth}
            \centering
            \includegraphics[width=\textwidth]{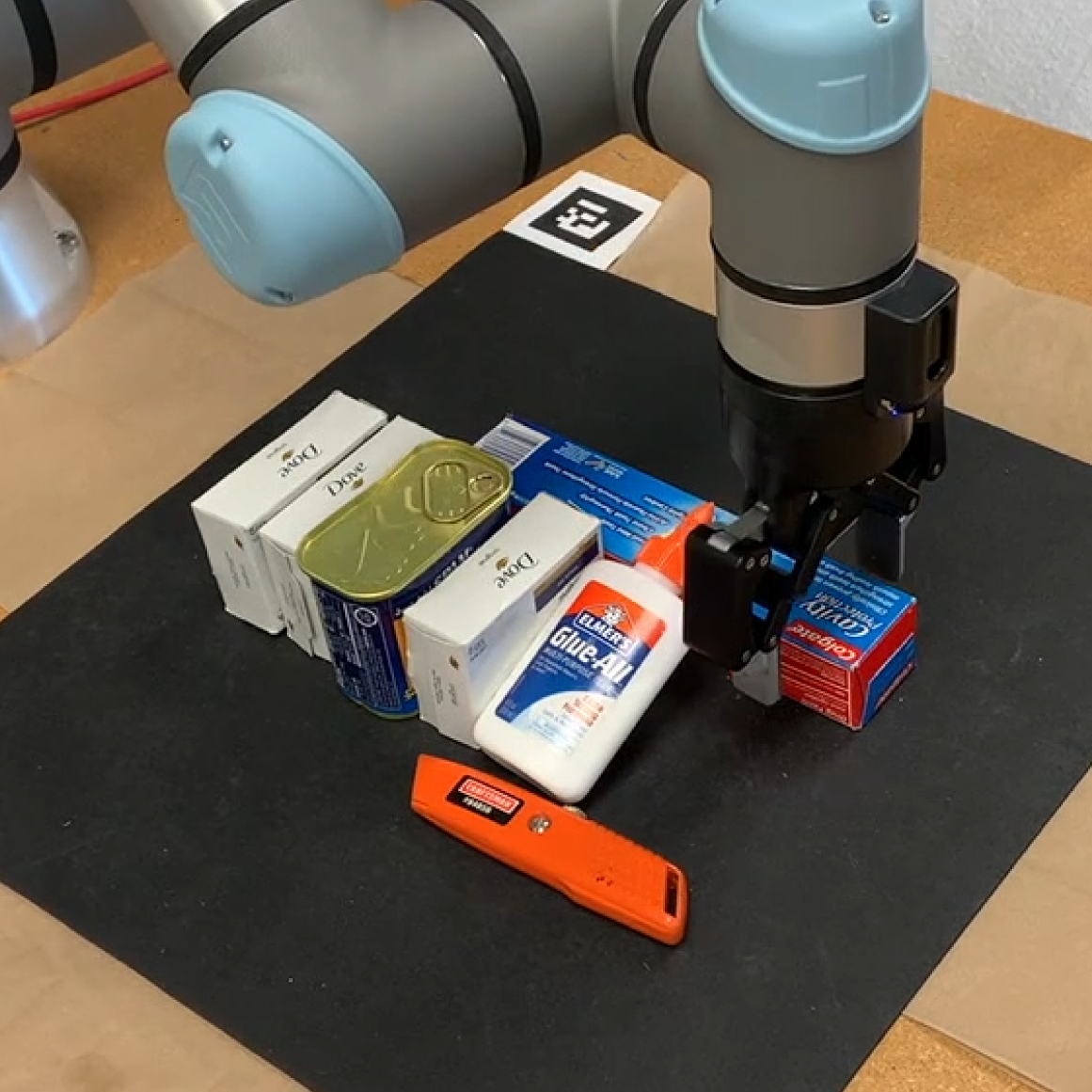}
            \caption{Grasping all from Clutter}
            \label{fig:intro-clutter}
        \end{subfigure}%
        \hfill%
        \begin{subfigure}[b]{0.49\linewidth}
            \centering
            \includegraphics[width=\textwidth, ]{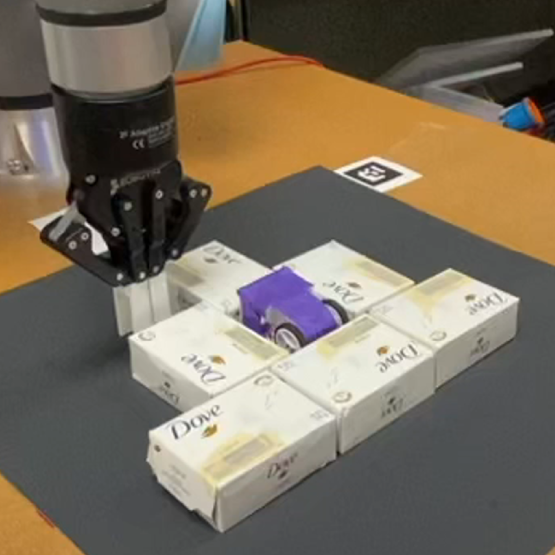}
            \caption{Grasping target from Clutter}
            \label{fig:intro-retrieve}
        \end{subfigure}
    \end{minipage}
    
    \vspace{1px}
    
    \begin{subfigure}[b]{0.7\textwidth}
        \centering
        \includegraphics[width=\textwidth]{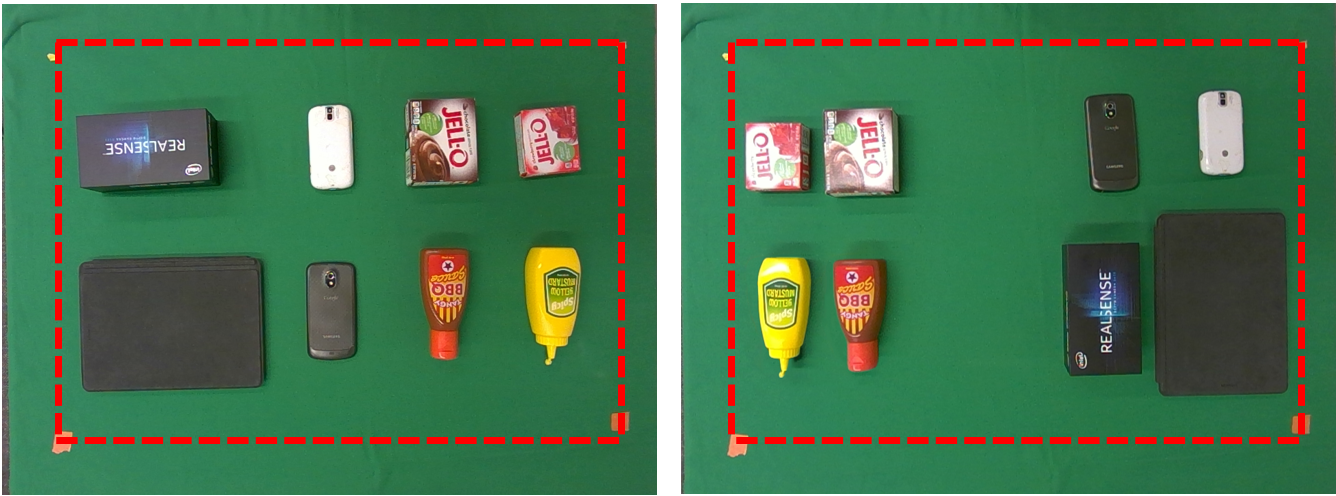}
        \caption{Object Rearrangement: From start (left) to goal (right)}
        \label{fig:intro-rearrangement}
    \end{subfigure}
    \caption{Examples of robot manipulation tasks. (a) Grasping objects from clutter: This can involve grasping all objects or targeting a specific item. The challenge lies in creating sufficient space for the gripper to access objects. (b) Dynamic grasping: Manipulating moving objects, such as receiving an item from a human hand. (c) Object rearrangement: Reorganizing objects to achieve a desired layout, similar to housekeeping. This task requires both high-level planning and precise motion control.}
    \label{fig:intro-imgs}
\end{figure}

The field of robotic manipulation has a rich history of research and development. Approaches to solving these problems have generally fallen into two main categories: learning-based methods and analytical algorithms. Learning-based methods, which rely on neural networks to determine the next action, show promise in performing specific tasks. However, they currently face limitations regarding stability and generalization due to the scarcity of training data in the robotics domain~\cite{andrychowicz2020learning}. 
On the other hand, analytical algorithms, while often more stable, struggle with defining metrics and rules from visual inputs, face challenges in exploring vast solution spaces, and frequently lack the flexibility to adapt to varied problems or changes in the scene. These methods are often sensitive to environmental variations and struggle to generalize across different scenarios~\cite{rodriguez2012caging}.

Given these challenges, developing robust and versatile algorithms for real-world robotic manipulation applications is paramount. Such algorithms must be capable of planning solutions, reasoning about physical interactions between objects, and controlling the robot to complete tasks successfully. 
We aim to create a general framework supporting similar but distinct applications, balancing solution quality and computational efficiency.
This dissertation aims to address these challenges by proposing approaches combining the strengths of learning-based methods and analytical algorithms in robotics. 
We explore using deep learning to predict object interactions, integrate these predictions with tree search algorithms for efficient planning, and leverage parallel computing to accelerate decision-making processes. Our work spans various aspects of robotic manipulation, from object retrieval in cluttered environments to dynamic grasping of moving objects, with the overarching goal of advancing the capabilities of robotic systems in handling complex, real-world manipulation tasks.
Through a series of interconnected studies, we demonstrate how our proposed methods achieve state-of-the-art performance in various challenging manipulation scenarios. By focusing on both the theoretical foundations and practical implementations of these algorithms, we contribute to the broader goal of deploying versatile and efficient robotic systems in unstructured real-world environments.

\begin{figure}[ht]
    \centering
    \includegraphics[width = 0.98\linewidth]{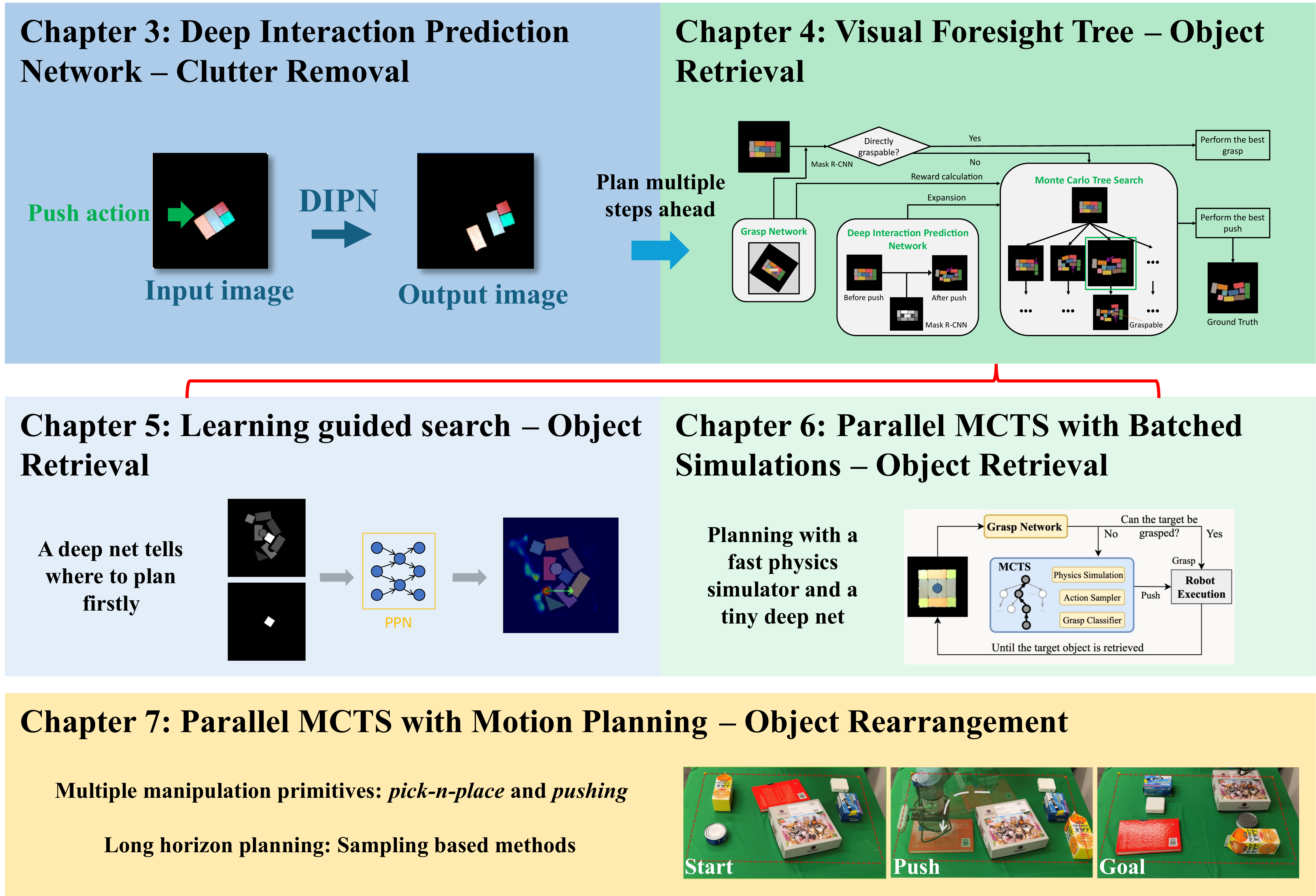}
    \caption{Structure of the dissertation. \autoref{chap:dipn} introduces the Deep Interaction Prediction Network for one-step push prediction in clutter removal. \autoref{chap:vft} extends to multi-step planning for efficient object retrieval using push actions. \Cref{chap:more,chap:pmbs} explore GPU-accelerated Monte Carlo tree search: \autoref{chap:more} focuses on learning a strategic network to guide tree search, while \autoref{chap:pmbs} utilizes Isaac Gym for parallel simulations in real robot execution. \autoref{chap:remp} applies similar concepts to object rearrangement in constrained spaces, incorporating motion planning.}
    \label{fig:intro-overview}
\end{figure}

This dissertation is structured to align with the planned chapters, each corresponding to published research work~\cite{huang2021dipn, huang2021visual, han2021toward, gao2022fast, huang2022interleaving, zhao2022stackelberg, huang2022parallel, huang2023earl, zhang2023learning, gao2023minimizing, huang2024toward, chang2024lgmcts}.

\section{DIPN: Deep Interaction Prediction Network with Application to Clutter Removal}

\autoref{chap:dipn} introduces the Deep Interaction Prediction Network (DIPN), a neural model designed to predict the effects of push actions in cluttered environments. DIPN leverages deep learning to estimate object interactions when a robot manipulator executes push actions, generating accurate synthetic images of potential outcomes. This predictive capability enables the system to make intelligent push-versus-grasp decisions, ultimately facilitating efficient clutter removal. By integrating DIPN with a grasp prediction network, the system achieves robust self-supervised learning, significantly outperforming previous state-of-the-art methods. Remarkably, DIPN demonstrates superior generalization on real hardware compared to simulation, underscoring its practical utility in robotic manipulation tasks.

\section{Visual Foresight Trees for Object Retrieval from Clutter with Nonprehensile Rearrangement}

Building on single-step push predictions, \autoref{chap:vft} extends the scope to multi-step planning for object retrieval in clutter. This chapter presents the Visual Foresight Trees (VFT) framework, which employs a deep predictive model (DIPN) to anticipate object movements resulting from sequential pushing actions. By integrating a tree search algorithm, VFT evaluates various push sequences to determine the optimal strategy for rearranging the environment before grasping a target object. The method significantly improves retrieval success rates and reduces the number of required actions compared to the baseline. Experiments in both simulation and real-world settings confirm that VFT effectively balances prediction accuracy with computational efficiency, paving the way for more intelligent robotic planning.

\section{Interleaving Monte Carlo Tree Search and Self-Supervised Learning for Object Retrieval in Clutter}

\autoref{chap:more} explores a deeper integration of Monte Carlo Tree Search with learning-based strategies for object retrieval in cluttered environments. The Monte Carlo tree search and learning for Object REtrieval (MORE) framework follows a self-supervised approach inspired by Kahneman’s System 2 → System 1 learning paradigm. Initially, MCTS enables a deep neural network to understand object interactions and predict optimal push actions.  Once trained, the network is incorporated into MCTS to accelerate decision-making, significantly reducing computational overhead while maintaining or improving solution quality. MORE represents a key step in closing the loop between classical planning and deep learning for efficient robotic manipulation.

\section{Parallel Monte Carlo Tree Search with Batched Rigid-body Simulations for Speeding up Long-Horizon Episodic Robot Planning}

Unlike \autoref{chap:more}, which focuses on learning-guided Monte Carlo Tree Search (MCTS), PMBS takes a different approach by introducing large-scale parallel simulations to enhance planning efficiency. The Parallel Monte Carlo Tree Search with Batched Simulations (PMBS) method leverages GPU-based parallelism to evaluate multiple action trajectories simultaneously using Isaac Gym. By executing a vast number of physics simulations in parallel with strategic sampling, PMBS significantly accelerates long-horizon planning tasks, such as object retrieval from clutter, achieving over $30\times$ speedups compared to conventional MCTS. PMBS achieves significant computational efficiency and maintains high solution quality while minimizing planning latency, thus making real-time robotic decision-making more viable. Experimental results further demonstrate that PMBS can be seamlessly deployed on real hardware with minimal sim-to-real discrepancies.

\section{Toward Optimal Tabletop Rearrangement with Multiple Manipulation Primitives}

Finally, \autoref{chap:remp} applies these principles to a broader class of robotic manipulation tasks. This chapter presents the Rearrangement with Multiple Manipulation Primitives (REMP) problem, which involves coordinating pick-and-place and push actions to optimally organize objects in constrained spaces. Two complementary algorithms are developed: hierarchical best-first search (HBFS) for fast heuristic planning and parallel MCTS for multi-primitive rearrangement (PMMR) for high-quality, human-like solutions. The integration of push and pick-and-place strategies allows robots to efficiently solve complex rearrangement tasks that require diverse manipulation skills. Extensive evaluations in both simulation and real-world environments highlight the effectiveness of these approaches in optimizing object placement while ensuring task success.

\chapter{Related Works}
\thispagestyle{myheadings}

Robotic manipulation encompasses a broad range of methods and tasks, including grasping, singulation, retrieval, rearrangement, pushing, and combined push-grasping strategies. The following sections review the relevant literature along these dimensions, highlighting both classical (analytical) approaches and more recent learning-based methods. We also highlight how Task and Motion Planning (TAMP), Monte Carlo Tree Search (MCTS), and other advanced approaches have been used to tackle long-horizon challenges such as clutter removal, object retrieval, and object rearrangement.

\section{Prehensile vs. Non-Prehensile Manipulation}
Manipulation actions can be broadly classified into prehensile and non-prehensile actions. Prehensile (or ``grasping'') actions involve lifting or holding objects using a gripper or suction, while non-prehensile actions (e.g., pushing, dragging, toppling) manipulate objects by applying forces without grasping them. Often, these two families of actions are studied separately~\cite{bicchi2000robotic, bauza2017probabilistic, shome2019towards}. However, there is an increasing interest in leveraging both types to tackle challenging tasks more effectively~\cite{zeng2018learning, hang2019pre, dogar2014object}.
Despite being limited in variety, prehensile actions allow a robot to secure an object and move it in tandem with the end-effector. Non-prehensile actions, on the other hand, use contact with the environment to control or guide objects on a surface~\cite{mason2018toward, hou2020robust, doshi2022manipulation}. As we will discuss below, combining the strengths of prehensile and non-prehensile actions (e.g., pushing to create space or orient an object prior to grasping) can significantly improve performance in cluttered or complex environments.

\section{Object Grasping}

\subsection{Analytical vs. Data-Driven Approaches}

Robotic grasping methods can generally be divided into analytical and data-driven categories. Analytical approaches rely on precise object models (often 3D) and mechanical properties (e.g., friction coefficients and mass distributions) to evaluate force-closure or form-closure conditions for grasp stability~\cite{rodriguez2012caging, bicchi2000robotic}. However, building these exact models can be prohibitively difficult in real-world settings due to incomplete scans of objects, unknown friction coefficients, and inaccurate mass estimates.

To address these challenges, data-driven methods~\cite{bohg2013data} learn grasp success probabilities directly from observations. Early data-driven techniques often focused on isolated single objects~\cite{detry2013learning, lenz2015deep, kappler2015leveraging, yan2018learning, mousavian20196}. More recent work has shifted to grasping in clutter, exploring how learned models can effectively handle occlusions, unexpected contacts, and multi-object interactions~\cite{liang2019pointnetgpd, ten2018using, pinto2016supersizing, mahler2017learning, fang2020graspnet, boularias2014efficient, wen2022catgrasp}. A common approach trains convolutional neural networks (CNNs) to propose candidate grasp actions, sometimes producing 6-DoF grasp poses in point clouds, such as Dex-Net~\cite{mahler2017dex, mahler2019learning}. Dex-Net and other approaches also combine suction grippers with fingered jaws to maximize grasp stability~\cite{mahler2018dex, deng2019deep}.

\subsection{Grasping in Dense Environments}
When objects are densely packed, direct grasping may fail due to collisions or partial occlusions. Recent techniques incorporate additional manipulation primitives (e.g., pushing, poking, or top-sliding) to move obstacles or improve grasp accessibility~\cite{zeng2018learning, xu2021efficient}. This synergy motivates us to combine grasping with auxiliary actions.

\section{Pushing}
Pushing is an example of non-prehensile actions that robots can apply on objects. As with grasping, there are two main categories of methods that predict the effect of a push action~\cite{stuber2020let}. \textit{Analytical} methods rely on mechanical and geometric models of the objects and utilize physics simulations to predict the motion of an object~\cite{lynch1993estimating, mason1986mechanics, lynch1996stable, lynch1999dynamic, akella1998posing, mason1986scope}. Notably, Mason~\cite{mason1986mechanics} derived the voting theorem to predict the rotation and translation of an object pushed by a point contact. A stable pushing technique when objects remain in contact was also proposed in~\cite{lynch1996stable}. These methods often make strong assumptions such as quasi-static motion and uniform friction coefficients and mass distributions. To deal with non-uniform frictions, a regression method was proposed in~\cite{yoshikawa1991indentification} for identifying the support points of a pushed object by dividing the support surface into a grid.

The \textit{limit surface} plays a crucial role in the mechanical models of pushing. It is a convex set of all friction forces and torques that can be applied to an object in quasi-static pushing. The limit surface is often approximated as an ellipsoid~\cite{howe1996practical}, or a higher-order convex polynomial~\cite{zhou2016convex, zhou2018convex, zhou2019pushing}. An ellipsoid approximation was also used to simulate the motion of a pushed object to perform a push-grasp~\cite{dogar2011framework}. To overcome the rigid assumptions of analytical methods, \textit{statistical learning} techniques predict how new objects behave under various pushing forces by generalizing observed motions in training examples. For example, a Gaussian process was used to solve this problem in~\cite{bauza2017probabilistic}, but was limited to isolated single objects. Most recent push prediction techniques rely on deep learning~\cite{NIPS2016_6161,byravan2017se3,watters2017visual}, which can capture a wider range of physical interactions from vision. Deep RL was also used for learning pushing strategies from images~\cite{sergey2015learning,levine2016end,finn2017deep,ghadirzadeh2017deep}.

\section{Push-Grasping}

Combining pushing and grasping in a single framework has led to more robust performance in cluttered settings. Two common paradigms are pre-grasp push, where non-prehensile actions reposition or uncover a target object for an ensuing grasp, and push-assisted grasping, where the push directly aids in achieving a stable grasp.
Several techniques implement a push-then-grasp sequence to reduce clutter around a target object~\cite{hang2019pre, dogar2010push, dogar2014object, dogar2012physics, king2013pregrasp}. Others focus on 'push-grasp' actions that simultaneously slide and lift the object~\cite{dogar2011framework}. The Visual Pushing and Grasping (VPG) framework~\cite{zeng2018learning} is a well-known example that uses a model-free Q-learning approach to select push or grasp actions. Extensions to VPG further incorporate predictive models of how objects move under push, allowing for 'look-ahead' planning rather than purely reactive decisions. In such model-based setups, the robot can simulate pushing outcomes before deciding whether and where to push, thereby avoiding unnecessary actions in cluttered environments.

\section{Singulation}
\subsection{Definition and Techniques}
Singulation refers to isolating one or more specific objects from a cluttered collection~\cite{chang2012interactive}. By clearing the target’s immediate surroundings, singulation can make grasping or retrieval more straightforward. A common approach is to use a combination of pushing and grasping actions to move objects that are obstructing the move. The methods in~\cite{eitel2020learning, danielczuk2018linear, tang2021learning} often rely on model-free reinforcement learning (RL) policies that reactively push objects away until the target is exposed.

\subsection{Limitations and Extensions}
These reactive techniques can be effective in lightly cluttered or moderate-density scenes, where a single push often suffices to create sufficient clearance~\cite{chang2012interactive}. However, for more densely packed scenarios or targets that require repeated, strategically chosen pushes, short-sighted or purely reactive methods may underperform. This limitation has spurred research on longer-horizon or model-based push planning to enable more intelligent singulation strategies.

\section{Object Retrieval}
\subsection{Problem Setting and Challenges}
Object retrieval tasks aim to locate and extract a target object from clutter. In many real-world applications (e.g., warehouse order fulfillment, household assistance), objects are stacked or partially occluded, necessitating a sequence of deliberate actions to uncover and grasp the target. This process often involves pushing, rearranging, or even removing other objects from the workspace.

\subsection{Model-Free vs. Model-Based Approaches}
Early methods treat object retrieval as an online planning challenge under partial observability~\cite{dogar2011framework}, sometimes relying on search heuristics to reduce the solution space. Others explore model-free RL to select among push, poke, or grasp actions~\cite{xiao2019online, danielczuk2019mechanical, kurenkov2020visuomotor}. While these methods can learn effective strategies for moderate clutter, long-horizon reasoning is often limited. More recent work integrates predictive models to anticipate how objects will move under certain pushes~\cite{huang2021dipn}, or uses MCTS to systematically explore multi-step action sequences~\cite{song2020multi}. Explicitly modeling future states is especially beneficial in tightly packed scenes, where small changes can substantially affect the feasibility of extracting the target.

\section{Rearrangement Planning}
Rearrangement planning extends object retrieval to more general problems of reconfiguring multiple objects from an initial to a goal arrangement~\cite{song2020multi, HanStiKroBekYu17RSS,han2018complexity,GaoFenYuRSS21,YuRSS21,Yu23IJRR,GaoFenHuaYu23IJRR,tang2023selective, ahn2022coordination}. In tabletop scenarios, rearranging objects often requires carefully allocating free space, deciding which objects to move first, and determining how to move them. Many methods rely on pick-and-place only, treating pushing as secondary or ignoring it entirely~\cite{moll2017randomized, gao2022fast, wang2022efficient}. However, pick-and-place can be inefficient or infeasible when objects are heavy, large, or extremely cluttered.
Some approaches reduce complexity by removing objects from the workspace altogether~\cite{tang2023selective} or by temporarily using external space to hold objects. When dealing with tightly packed configurations, heuristic-guided search has been used~\cite{gao2022fast, moll2017randomized}. Graph-based formulations can capture dependencies among objects that block one another~\cite{HanStiKroBekYu17RSS,han2018complexity}. Beyond classic motion planning frameworks, data-driven or learning-based rearrangement methods have emerged, harnessing deep neural networks or reinforcement learning to guide action selection~\cite{song2020multi}.

\section{Task and Motion Planning (TAMP)}
Task and motion planning (TAMP) deals with orchestrating high-level actions (tasks) while simultaneously ensuring geometric and kinematic feasibility (motion planning)~\cite{kaelbling2011hierarchical, srivastava2014combined, toussaint2015logic, dantam2016incremental}. Compared to discrete, rule-based domains (e.g., board games), TAMP operates over continuous state and action spaces. Traditional TAMP approaches often combine symbolic reasoning with sampling-based motion planning~\cite{garrett2020pddlstream, migimatsu2020object}. More recent work leverages learning—such as learning to guide search, predict outcomes of actions, or estimate feasibility—to navigate the large search space~\cite{chitnis2016guided, kim2019learning, driess2020deep, huang2021dipn, wells2019learning}. In the context of object retrieval or rearrangement, TAMP can formalize the problem: the task layer decides which object to move and how, while the motion planner ensures a collision-free trajectory for each manipulation primitive. When clutter is dense, the space of feasible moves can be large, making efficient search strategies or learned heuristics crucial.

\section{Monte Carlo Tree Search (MCTS) for Manipulation}
Monte Carlo Tree Search (MCTS) has shown promise in multi-step decision-making for manipulation, particularly in cluttered scenes~\cite{song2020multi}. MCTS incrementally expands a lookahead search tree, simulating potential action sequences to estimate their outcomes (e.g., clearing clutter to expose the target). Often, these simulations rely on predictive models—either analytical or learned—to approximate object dynamics, collisions, and future states~\cite{bauza2017probabilistic, watters2017visual, NIPS2016_6161, byravan2017se3, song2020multi}.
When integrated with data-driven push or grasp predictors, MCTS can efficiently explore extended action sequences, balancing exploration of different moves with exploitation of promising trajectories~\cite{song2020multi, kurenkov2020visuomotor}. Recent work also explores network-based predictions of multi-object collisions under pushing to accelerate the simulation phase~\cite{sergey2015learning, levine2016end, finn2017deep, ghadirzadeh2017deep}. Nevertheless, designing accurate predictive networks remains challenging in highly cluttered or diverse object sets~\cite{dengler2022learning}.
Overall, MCTS-based approaches hold substantial potential for manipulation tasks that require long-horizon reasoning, such as object retrieval or complex rearrangement, by combining learned predictive models with systematic search to reduce trial-and-error in the physical environment.

\vspace{10pt}
\noindent
By unifying insights from these diverse research areas—prehensile and non-prehensile actions, pushing and grasping, singulation, object retrieval, rearrangement, TAMP, and MCTS-based planning—we see that integrated strategies are critical for addressing the challenges posed by dense clutter. The remainder of this dissertation builds on these findings, focusing on how to combine model-based predictions, data-driven methods, and efficient planning techniques to enable robust, long-horizon manipulation in real-world settings.
    
\chapter{DIPN: Deep Interaction Prediction Network with Application to Clutter Removal}\label{chap:dipn}
\thispagestyle{myheadings}

\newcommand{\pag}{\textsc{PaG}\xspace}
\newcommand{\dipn}{\textsc{DIPN}\xspace}
\newcommand{\gn}{\textsc{GN}\xspace}
\newcommand{\dipngn}{\textsc{DIPN+GN}\xspace}
\newtheorem{problem}{Problem}

% text of this chapter goes here
\section{Introduction}

We propose a Deep Interaction Prediction Network (\dipn) for learning object 
interactions directly from examples and using the trained network for accurately 
predicting the poses of the objects after an arbitrary push action (\autoref{fig:dipn-intro}). 
To demonstrate its effectiveness, 
We integrate \dipn with a deep Grasp Network (\gn) for completing challenging 
clutter removal manipulation tasks. 
Given grasp and push actions to choose from, the objective is to remove all objects 
from the scene/workspace with a minimum number of actions. 
In an iteration of push/pick selection (\autoref{fig:dipn-system-architecture}), the 
system examines the scene and samples a large number of candidate grasp and push 
actions. Grasps are immediately scored by \gn, whereas for each candidate push 
action, \dipn generates an image corresponding to the predicted outcome. In a sense, 
\dipn ``imagines'' what happens to the current scene if the robot executes a 
certain push. The predicted future images are also scored by \gn; the action with 
the highest expected score, either a push or a grasp, is then executed. 

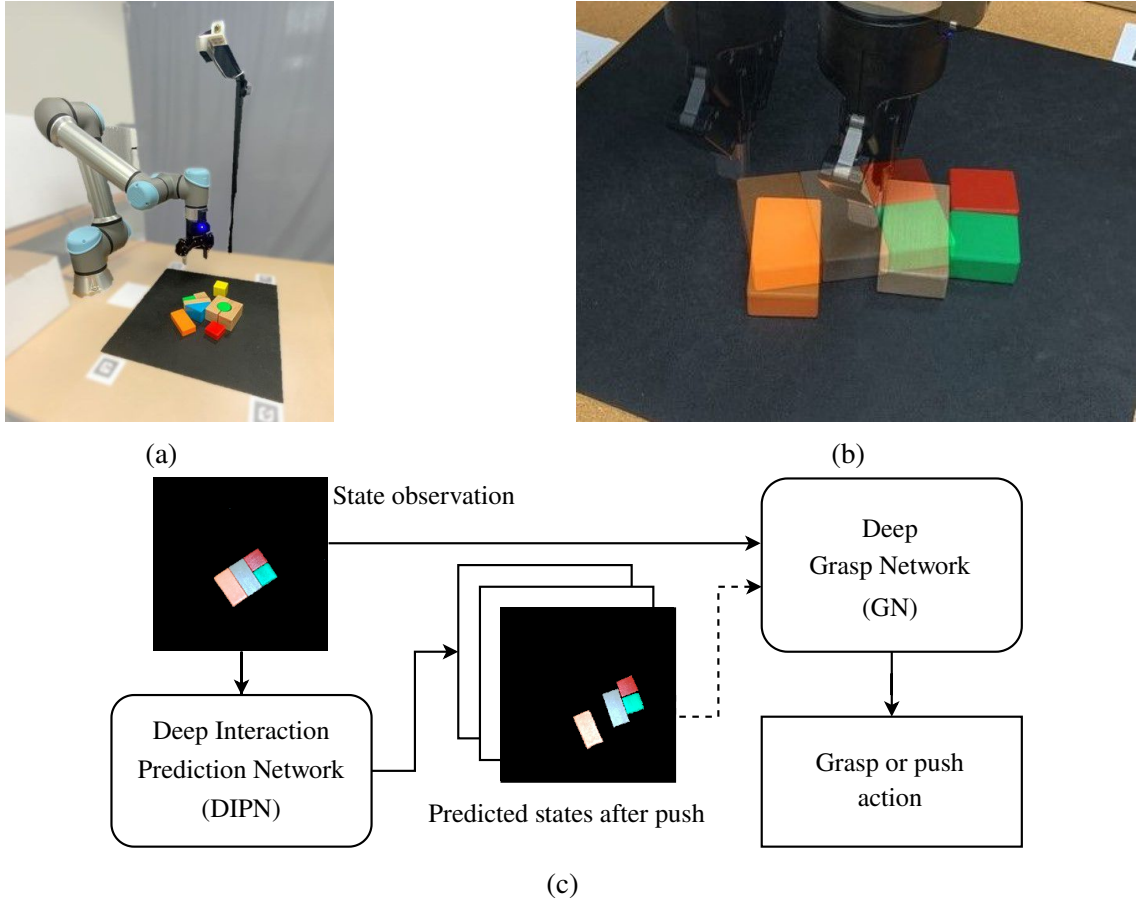
\begin{figure}[ht!]
    \centering
    \subfloat[]{
        \includegraphics[height=2.2in]{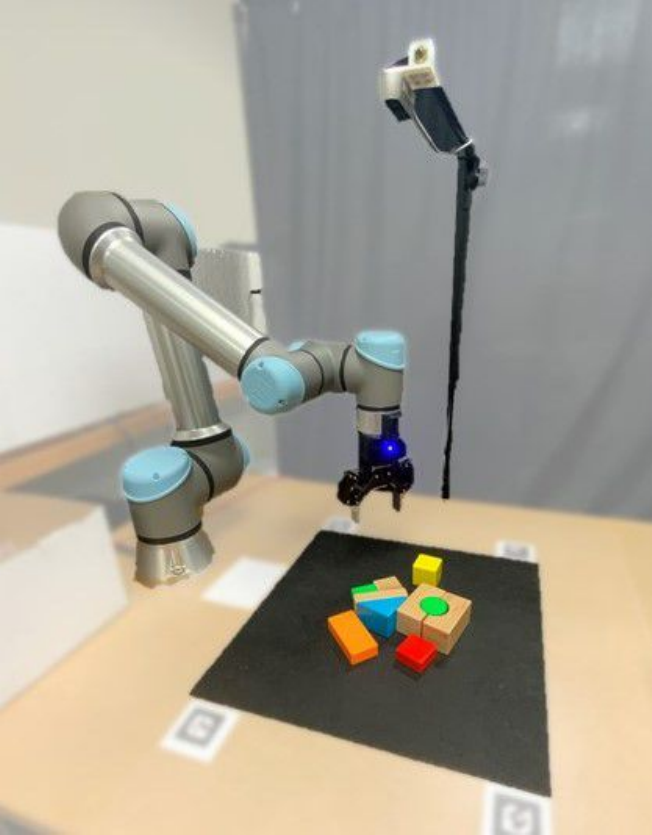}\label{fig:dipn-system-setup}
    }
    \hfill
    \subfloat[]{
        \includegraphics[height=2.2in]{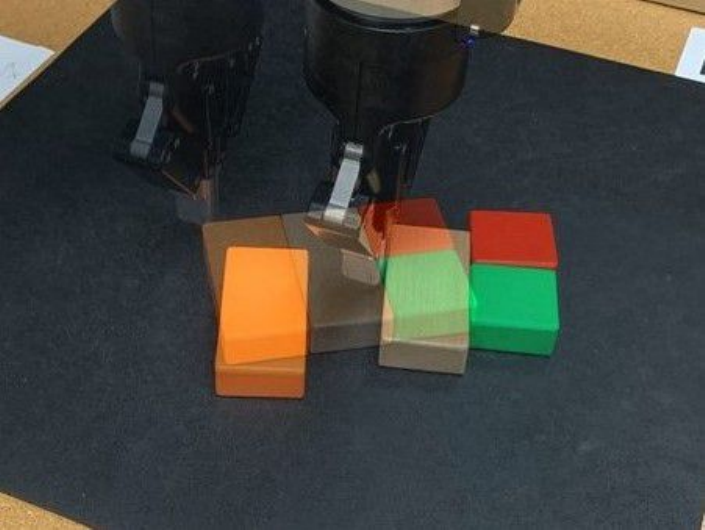}\label{fig:dipn-push-action-example}
    }
        
    \subfloat[]{
        \begin{overpic}[width=0.8\linewidth]{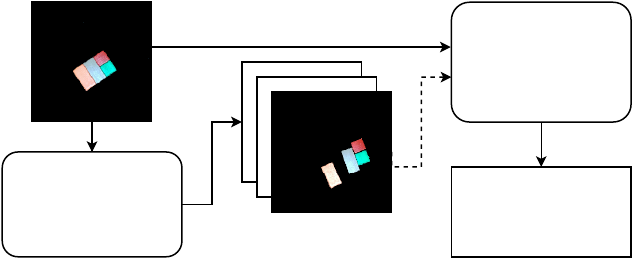}
            \footnotesize
            \put(24.5, 38) {State observation}
            \put(14.5, 13) {\makebox(0,0){Deep Interaction}}
            \put(14.5, 9) {\makebox(0,0){Prediction Network}}
            \put(14.5, 4.5) {\makebox(0,0){(\dipn)}}
            \put(50, 4) {\makebox(0,0){Predicted states after push}}
            \put(85.25, 35) {\makebox(0,0){Deep}}
            \put(85.25, 31) {\makebox(0,0){Grasp Network}}
            \put(85.25, 26.5) {\makebox(0,0){(\gn)}}
            \put(85.25, 9.5) {\makebox(0,0){Grasp or push}}
            \put(85.25, 6) {\makebox(0,0){action}}
        \end{overpic}
        \label{fig:dipn-system-architecture} 
    }
    \caption{\label{fig:dipn-intro} 
    (a) The system setup includes a workspace with objects to remove, a Universal Robots UR-5e manipulator with a Robotiq 2F-85 two-finger gripper, and an Intel RealSense D435 RGB-D camera. 
    (b) An example push action and superimposed images of scenes before and after the push. 
    (c) System architecture of our pipeline, and one predicted image that \dipn can generate for the push shown in (b). Notice the similarity between the predicted synthetic image and the real image resulting from the push action.
    }
\end{figure}

Our extensive evaluation demonstrates that \dipn can accurately predict objects' 
poses after a push action with collisions, resulting in less than $10\%$ average 
single object pose error in terms of IoU (Intersection-over-Union), a significant 
improvement over the compared baselines.
Push prediction by \dipn generates clear synthetic images that can be used by \gn 
to evaluate grasp actions in future states. Together with \gn, our entire pipeline 
achieves $34\%$ higher completion rate, $20.9\%$ higher grasp success rate, and 
$30.4\%$ higher action efficiency in comparison to~\cite{zeng2018learning} on 
challenging clutter removal scenarios. 
Moreover, experiments suggest that \dipn can learn from randomly generated scenarios 
with the learned policy maintaining high levels of performance on challenging tasks 
involving previously unseen objects.
Remarkably, \dipngn achieves even better performance on real robotic hardware than in the simulation environment where it was developed.

\section{Problem Formulation}

We formulate the clutter removal problem (\autoref{fig:dipn-system-setup}) as Pushing 
Assisted Grasping (\pag). In a \pag, the workspace of the manipulator is a 
square region containing multiple objects and the volume directly above it. 
A camera is placed on top of the workspace for state observation. Given camera 
images, all objects must be removed using two basic motion primitives, grasp, 
and push, with a minimum number of actions.

In our experimental setup, the workspace has a uniform background color and the 
objects have different shapes, sizes, and colors. The end-effector is a two-finger 
gripper with a narrow stroke that is slightly larger than the smallest dimension 
of individual objects. Objects are removed one by one, which requires a sequence 
of push and grasp actions.
When deciding on the next action, a state observation is given as an 
RGB-D image, re-projected orthographically, cropped to the workspace's boundary, 
and down-sampled to $224 \times 224$. 

From the down-sampled image, a large set of candidate actions is generated by 
considering each pixel in the image as a potential center of a grasp action or 
initial contact point of a push action.
A grasp action $a^{\text{grasp}} = (x, y, \theta)$ is a vertical top-down grasp 
centered at pixel position $(x, y)$ with the end-effector rotation set to $\theta$
around the vertical axis of the workspace; a grasped object  is subsequently 
transferred outside of the workspace and removed from the scene. Similarly, a 
push action $a^{\text{push}} = (x, y, \theta)$ is a horizontal sweep motion that 
starts at $(x, y)$ and proceeds along $\theta$ direction for a fixed distance.
The orientation $\theta$ can be one of $16$ values evenly distributed between $0$ 
and $360$ degrees. That is, the entire action space includes $2 \times 224 \times 
224 \times 16$ different grasp/push actions. 

The problem studied in this paper is defined as: 
% \vspace*{-2mm}
\begin{problem}
{\normalfont \textbf{Pushing Assisted Grasping (\pag)}.} 
Given objects in clutter within the described system setup, determine a sequence of push and grasp actions, using only visual input from the workspace, to remove all objects while minimizing the total number of actions executed.
\end{problem}

\section{Methodology}
We describe the Deep Interaction Prediction Network (\dipn), the Grasp Network (\gn), 
and the integrated pipeline for solving \pag challenges.

\subsection{Deep Interaction Prediction Network (\dipn)}
The architecture of our proposed \dipn is outlined in~\autoref{fig:dipn-push-prediction-flowchart}.
At a high level, given an image and a candidate push action as inputs, \dipn segments 
the image and then predicts 2D transformations (translations and rotations) for all objects,
and particularly for those affected by the push action, directly or indirectly through a 
cascade of object-object interactions. 
A predicted image of the post-push scene is synthesized by applying the predicted 
transformations on the segments.
We opted against an end-to-end, pixel-to-pixel method as such methods (e.g.,
\cite{NIPS2016_6161}) often lead to blurry or fragmented images, which are not conducive 
to predicting the quality of a potential future grasp action. 

\begin{figure*}[ht!]
    \centering

    \begin{overpic}[width=\linewidth]{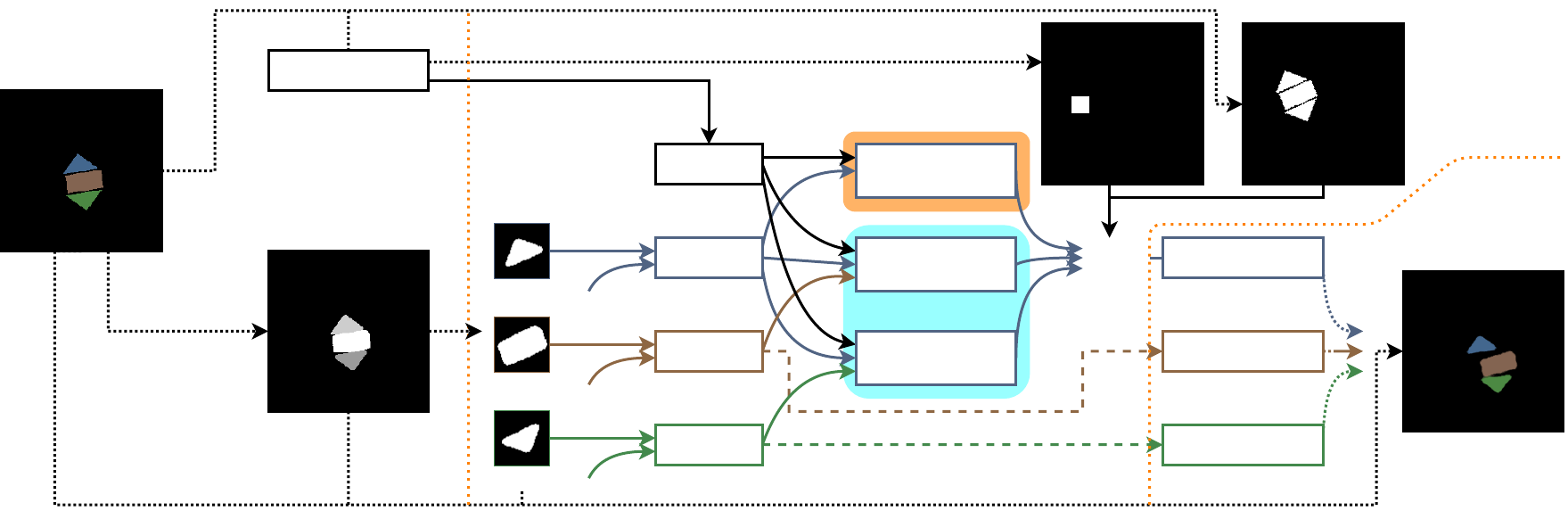}
        \tiny
        \put(0, 29.5) {Input image $I$}
        \put(0, 28) {$224 \times 224 \times 3$}
        \put(17, 19) {Segmentation}
        \put(17, 17.5) {$224 \times 224 \times 1$}
        \put(89.5, 18) {Output image $\hat I$}
        \put(89.5, 16.5) {$224 \times 224 \times 4$}
        \put(47.5, 27.5) {Binary push action image $M_p$}
        \put(54.5, 26) {$224 {\times} 224 {\times} \{0{,} 1\}$}
        \put(90, 27.5) {Binary image $M_I$}
        \put(73.5, 29.5) {\textcolor{orange}{\textbf{$M_p$}}}
        \put(86.5, 29.5) {\textcolor{orange}{\textbf{$M_I$}}}
        \put(90, 26) {$224 {\times} 224{\times} {\{}0{,} 1{\}}$}
        % \put(94.5, 24.5) {${\times} {\{}0{,} 1{\}}$}
        \put(22.3, 28.5) {\makebox(0,0){Push action $p$}}
        \put(22.3, 26.25) {\makebox(0,0){\tiny $(83, 140, 22.5^\circ)$}}
        \put(7, 10) {Mask R-CNN}
        \put(33.5, 21.5) {\makebox(0,0){\parbox{1in}{\centering Mask image$\{M_i\}$, \\ $60 {\times} 60 {\times} \{0{,} 1\}$, \\ center position $\{c_i\}$}}}
        \put(33.5, 14.4) {\makebox(0,0){\tiny $(54, 69)$}}
        \put(33.5, 8.4) {\makebox(0,0){\tiny $(67, 90)$}}
        \put(33.5, 2.4) {\makebox(0,0){\tiny $(73, 112)$}}
        \put(35.2, 15.8) {\textcolor{orange}{\textbf{$M_1$}}}
        \put(35.2, 9.6){\textcolor{orange}{\textbf{$M_2$}}}
        \put(35.2, 3.8) {\textcolor{orange}{\textbf{$M_3$}}}
        \put(36, 17.5) {\tiny ResNet\textsubscript{2}}
        \put(41, 26) {\tiny MLP\textsubscript{1}}
        \put(38.5, 14.5) {\tiny MLP\textsubscript{2}}
        \put(52.5, 24.4) {\tiny MLP\textsubscript{3}}
        \put(52.5, 18.4) {\tiny MLP\textsubscript{4}}
        \put(52.5, 12.4) {\tiny MLP\textsubscript{4}}
        \put(42.7, 22) {Encoded}
        \put(42.7, 16) {Encoded}
        \put(42.7, 10.1) {Encoded}
        \put(42.7, 4.1) {Encoded}
        \put(59.8, 22.2) {\makebox(0,0){\parbox{1in}{\tiny \centering Direct \\ transformation}}}
        \put(59.8, 16.2) {\makebox(0,0){\parbox{1in}{\tiny \centering Interactive \\ transformation}}}
        \put(59.8, 10.2) {\makebox(0,0){\parbox{1in}{\tiny \centering Interactive \\ transformation}}}
        \put(66, 13.5) {\tiny Add}
        \put(73.7, 19.5) {\makebox(0,0){\scriptsize ResNet\textsubscript{1}}}
        \put(69.5, 16.2) {\tiny MLP\textsubscript{5}}
        \put(79.3, 16.7) {\makebox(0,0){\tiny Transformation}}
        \put(79.3, 14.5) {\makebox(0,0){\tiny $(0, 0, 0^{\circ})$}}
        \put(79.3, 10.7) {\makebox(0,0){\tiny Transformation}}
        \put(79.3, 8.5) {\makebox(0,0){\tiny $(18, -3, 6^{\circ})$}}
        \put(79.3, 4.7) {\makebox(0,0){\tiny Transformation}}
        \put(79.3, 2.5) {\makebox(0,0){\tiny $(15, 1, -3^{\circ})$}}
    \end{overpic}
    \caption{\label{fig:dipn-push-prediction-flowchart} 
    \dipn flow with an example. 
    The network components dedicated to an object are color-coded to match the object. 
    We only show the full network for the blue triangle object; the instance-specific 
    structures for the other objects share the same weights and are simplified as dashed lines. Components inside the orange dotted line are the core of the DIPN. The output image is synthesized by applying the predicted transformations to the object segments.
    }
\end{figure*}
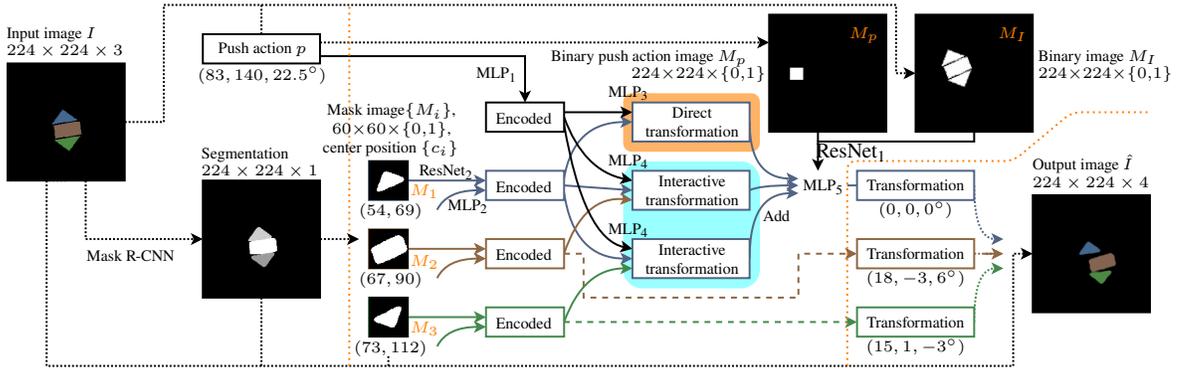

\textbf{Segmentation.} 
\dipn employs Mask R-CNN~\cite{he2017mask} for object segmentation (instance level only, 
without semantic segmentation). The resulting binary masks ($m_i$) and their centers 
($c_i$), one per object, serve as the input to the push prediction module of \dipn. 
Our Mask R-CNN setup has two classes, one for the background and 
one for the objects. 
The network is trained from scratch in a \emph{self-supervised} manner without any human 
intervention: objects are randomly dropped into the workspace, and data is automatically 
collected.
Images that can be easily segmented into separate instances based on color/depth 
information (distinct color blobs) are automatically labeled by the system as 
single instances for training the Mask R-CNN. 
The self-trained Mask R-CNN can then accurately find edges in images of tightly packed 
scenes and even in scenes with novel objects. Note that the data used for training the 
segmentation module are also counted in our evaluation of the data efficiency of our 
technique and the comparisons to alternative techniques.

\textbf{Push sampling.}
\begin{figure}[ht!]
    \centering
  \includegraphics[width=0.5\linewidth]{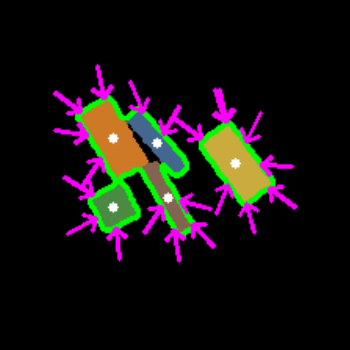}
  \caption{\label{fig:dipn-sample-action-new} 
    Sampled action in purple arrows around each object.}
\end{figure}
Based on foreground segmentation, candidate push actions are generated by uniformly sampling push contact locations on the contour of object clusters (see \autoref{fig:dipn-sample-action-new}). Push directions point 
to the centers of the objects. Pushes that cannot be performed physically, e.g., from 
the inside of an object bundle or in narrow spaces between objects, are filtered out 
based on the masks returned by R-CNN. In the figure, for example, samples between the 
two object clusters are removed. A sampled push is defined as $p = (x, y, \theta) \in 
SE(2)$ where $x$, $y$ are the start location of the push and $\theta$ indicates the 
horizontal push direction. The push distance is fixed.

\textbf{Input to push prediction}. The initial scene image $I$, scene object masks 
and centers $(m_i, c_i)$, and a sampled push action $p = (x, y,\theta)$ are the main 
inputs to the push prediction module. To reduce redundancy, we transform all inputs 
using a 2D homogeneous transformation matrix $T$ such that $pT = (40, 112, 0)$. The position of the push is normalized for easier learning such that a push will always go from left to right in the middle of the left side of the workspace.
From here, it is understood that all inputs are with respect to this updated 
coordinate frame defined by $T$, i.e., $p \leftarrow pT, c_i \leftarrow c_i T$, 
and so on. 
Apart from the inputs mentioned so far, we also generate: (1) one binary 
\emph{push action} image $M_p$ with all pixels black except in a small square with 
top-left corner $(40, 100)$ and bottom-right corner $(65, 124)$ which is the finger movement space, (2) one $224\times 
224$ binary image $M_I$ with the foreground of $I$ set to white, and (3) one $60 \times 
60$ binary mask image $M_i$ for each object mask $m_i$, centered at $c_i$. 
Despite being constant relative to the image transformed by $T$, push image $M_p$ is used 
as an input because we noticed from our experiments that it helps the network focus 
more on the pushing area. 

\textbf{Push prediction.} 
With global (binary images $M_p, M_I$) and local (mask image $M_i$ and the center $c_i$ 
of each object) information, \dipn proceeds to predict objects' transformations. 
To start, a Multi-Layer Perceptron (MLP) and a ResNet~\cite{DBLP:journals/corr/LinDGHHB16} 
(with no pre-training) are  used to encode the push action and the global information, 
respectively: 
\[e_p = \text{MLP}_\text{1}(p),\quad e_{AB} = \text{ResNet}_\text{1}(M_p,M_I).\]
A similar procedure is applied to individual objects. 
For each object $o_i$, its center $c_i$ and mask image $M_i$ are encoded using ResNet 
(again, with no pre-training) and MLP as: 
\[e_i = (\text{ResNet}_\text{2}(M_i),  \text{MLP}_\text{2}(c_i)).\]
Adopting the design philosophy from~\cite{watters2017visual}, the encoded information 
is then passed to 
a {\em direct transformation} (DT) MLP module 
(blocks in~\autoref{fig:dipn-push-prediction-flowchart} with orange background) 
and multiple {\em interactive transformation} (IT) MLP modules 
(blocks in~\autoref{fig:dipn-push-prediction-flowchart} with cyan background): 
\[\forall 1\leq i \leq n:\ \text{DT}_i = \text{MLP}_\text{3}(e_p, e_i), \]
\[\forall 1\leq i, j \leq n, j \neq i:\ \text{IT}_{ij} = \text{MLP}_\text{4}(e_p, e_i, e_j).\]
Here, the direct transformation modules capture the effect of the robot {\em directly} 
touching the objects (if any), while the interactive transformation modules consider 
collision between an object and every other object (if any).
Then, all aforementioned encoding is put together to a decoding MLP to derive the output 
2D transformation for each object $o_i$ in the push action's frame: $\forall 1 {\leq} i {\leq} n$, 
\[(\hat x_i, \hat y_i, \hat \theta_i) = 
\text{MLP}_\text{5}(e_{AB}, \text{DT}_i + \sum_{1\leq j \leq n, j \neq i} \text{IT}_{ij}),\]
which can be mapped back to the original coordinate frame via $T^{-1}$. This yields  
predicted poses of objects. 
From these, an ``imagined'' push prediction image is readily generated. 

In our implementation, both ResNet\textsubscript{1} and ResNet\textsubscript{2} are 
ResNet-50. MLP\textsubscript{1} and MLP\textsubscript{2}, encoding the push action 
and single object position, both have two (hidden) layers with sizes 8 and 16. 
MLP\textsubscript{3}, connecting encoded and direct transformations, has two layers 
of a uniform size 128. MLP\textsubscript{4}, connecting encoded and interactive 
transformations, has three layers with a size of 128 each. 
The final decoder MLP\textsubscript{5} has five layers with sizes [256, 64, 32, 16, 3]. 
The number of objects $n$ varies across scenes. 
The network handles a variable number of objects because the same weight-shared networks (MLP\textsubscript{2-5} and ResNet\textsubscript{2}) process each object and object pair.

\textbf{Training.}
For training in simulation and for real experiments, objects are randomly dropped 
onto the workspace. The robot then executes random pushes to collect training data.
\emph{SmoothL1Loss} (Huber Loss) is used as the loss function. 
Given each object's true post-push transformation $(x_i, y_i, \theta_i)$ and the 
predicted $(\hat x_i, \hat y_i, \hat \theta_i)$, the loss is computed as the sum of 
coordinate-wise SmoothL1Loss between the two.
\dipn performs well on unseen objects and can be completely trained in simulation 
and transferred to the real-world (Sim-to-Real). It is also robust with respect to 
changes in objects' physical properties, e.g., variations in mass and friction coefficients. 

\subsection{The Grasp Network (\gn)}
We briefly describe \gn, which shares a similar architecture to the DQN used 
in~\cite{zeng2018learning}. Given an observed image and candidate grasp actions, \gn finds the optimal policy 
for maximizing a single-step grasp reward, defined as $1$ for a successful grasp 
and $0$ otherwise. 
\gn focuses its \emph{attention} on local regions that are relevant to each single 
grasp and uses image-based self-supervised pre-training to achieve a good
initialization of network parameters. 

The proposed modified network's architecture is illustrated in Fig.~\ref{fig:grasp-network}. 
It takes an input image and outputs a score for each candidate grasp centered at 
each pixel. The input image is rotated to align it with the end-effector frame 
(see the left image in Fig.~\ref{fig:grasp-network}).  
 ResNet-50 FPN~\cite{DBLP:journals/corr/LinDGHHB16} is used as the backbone; 
we replace the last layer with our own customized head structure shown in Fig.~\ref{fig:grasp-network}. 
We observe that our structure leads to faster training and inference 
time without loss of accuracy. 
Given that the network computes pixel-wise values 
in favor of a local grasp region, at the end of the network, we place two 
convolutional layers with kernel size $11 \times 57$, which was determined based 
on the clearance of the gripper.

\begin{figure}[ht!]
    \centering
    \begin{overpic}[width=\linewidth]{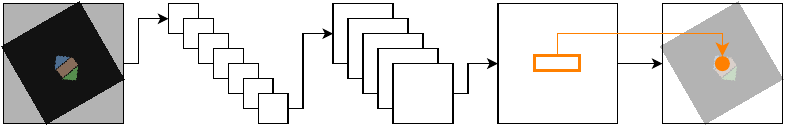}
        \scriptsize
        \put (8 , 18) {\makebox(0,0){\textcolor{NavyBlue}{$320{\times}320{\times}4$}}}
        \put (29, 18) {\makebox(0,0){\textcolor{RubineRed}{$256$}-\textcolor{NavyBlue}{$80{\times}80$}}}
        \put (50, 18) {\makebox(0,0){\textcolor{RubineRed}{$64$}-\textcolor{NavyBlue}{$160{\times}160$}}}
        \put (70, 18) {\makebox(0,0){\textcolor{RubineRed}{$1$}-\textcolor{NavyBlue}{$320{\times}320$}}}
        \put (91.5, 18) {\makebox(0,0){\textcolor{NavyBlue}{$320{\times}320{\times}1$}}}
        \put (18, -1.5) {\makebox(0,0){ResNet-50}}
        \put (18, -4) {\makebox(0,0){FPN P2}}
        \put (37, -1.5) {\makebox(0,0){Conv \textcolor{RubineRed}{128}, \textcolor{ForestGreen}{3$\times$3}}}
        \put (37, -6.5) {\makebox(0,0){\centering ReLU}}
        \put (37, -4) {\makebox(0,0){Batchnorm2d(\textcolor{RubineRed}{128})}}
        % \put (37, -9) {\makebox(0,0){\centering Dropout(0.1)}}
        \put (37, -9) {\makebox(0,0){Conv \textcolor{RubineRed}{128}, \textcolor{ForestGreen}{3$\times$3}}}
        \put (37, -11.5) {\makebox(0,0){Batchnorm2d(\textcolor{RubineRed}{128})}}
        \put (37, -14) {\makebox(0,0){\centering ReLU}}
        % \put (37, -19) {\makebox(0,0){\centering Dropout(0.1)}}
        \put (37, -16.5) {\makebox(0,0){\centering Interpolate 2$\times$}}
        \put (61, -1.5) {\makebox(0,0){Conv \textcolor{RubineRed}{32}, \textcolor{ForestGreen}{3$\times$3}}}
        \put (61, -4) {\makebox(0,0){Batchnorm2d(\textcolor{RubineRed}{32})}}
        \put (61, -6.5) {\makebox(0,0){\centering ReLU}}
        \put (61, -9) {\makebox(0,0){\centering Dropout(0.1)}}
        \put (61, -11.5) {\makebox(0,0){Conv \textcolor{RubineRed}{1}, \textcolor{ForestGreen}{3$\times$3}}}
        \put (61, -14) {\makebox(0,0){\centering Interpolate 2$\times$}}
        \put (83, -1.5) {\makebox(0,0){Conv \textcolor{RubineRed}{1}, \textcolor{ForestGreen}{11$\times$57}}}
        \put (83, -4) {\makebox(0,0){ReLU}}
        \put (83, -6.5) {\makebox(0,0){Conv \textcolor{RubineRed}{1}, \textcolor{ForestGreen}{11$\times$57}}}
    \end{overpic}
    \vspace*{60pt}
    \caption{\label{fig:grasp-network}
    Architecture of \gn.
    Pink, blue, and green text are used for channel count, image size, and kernel 
    size, respectively. 
    }
\end{figure}
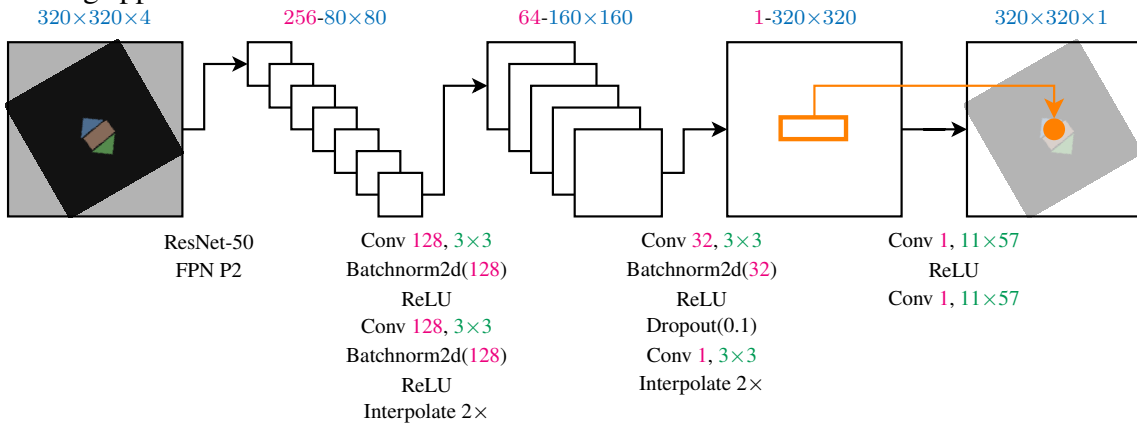

In training \gn, image-based pre-training~\cite{yen2020learning} 
was employed. The pre-training process treats pixel-wise grasping as a vision task 
to obtain a good network initialization. The process automatically labels with $0$ 
or $1$ all the pixels in a small set of arbitrary images, depending on whether grasps 
centered at each pixel would lead to a finger collision with an object, based only on 
color/depth and without actually simulating or executing the grasps physically. 
The pre-training data set does not include objects used for testing.

\subsection{The Complete Algorithmic Pipeline}

The training process of \dipngn is outlined in~\autoref{alg:dipn-train}. In~\autoref{alg:dipn-train-pre}, an image data set is collected for training Mask R-CNN 
(i.e., for push prediction segmentation) and initializing (i.e., pre-training) \gn. 
Note that training data for Mask R-CNN and pre-training data for GN are essentially 
free, with no physics involved. After pre-training, the training process for push 
(\autoref{alg:dipn-train-push}) and grasp (\autoref{alg:dipn-train-grasp}) predictions can 
be executed on \pag scenes in any order.

\SetCommentSty{mycommfont}
\begin{algorithm}
    \small
    \DontPrintSemicolon
    \KwOut{trained \dipngn.}
    \gn, Mask R-CNN${\gets}$GetImageDataSetAndPre-Train\,$()$\label{alg:dipn-train-pre}\;
    \dipn $\gets$ TrainOn{\pag}PushOnly\,$($Mask R-CNN$)$ \label{alg:dipn-train-push}\;
    \gn $\gets$ TrainOn{\pag}GraspOnly\,$($GN$)$ \label{alg:dipn-train-grasp}\; 
    \caption{\label{alg:dipn-train}
    Training \dipngn
    }
\end{algorithm}

The high-level workflow of our framework on \pag is described in ~\autoref{alg:dipn-evaluation}. 
When working on an instance, at every decision-making step $t$, an image $M_t$ is first obtained 
(\autoref{alg:dipn-evaluation-input}). 
Then, the image $M_t$, along with sampled push actions $A^{\text{push}}$, are sent to 
the trained \dipn to generate predicted synthetic images $\hat{M}_{t+1}$ after each imagined 
push $a$ (\autoref{alg:dipn-evaluation-sample}-\autoref{alg:dipn-evaluation-dipn}). 
With $A^{\text{grasp}}$ denoting the set of all grasp actions, 
their discounted average reward on the predicted next image $\hat{M}_{t+1}$ is then 
compared with the average of grasping rewards in the current image 
(\autoref{alg:dipn-evaluation-compare}): recall that \gn takes an image and a grasp action as 
input, and outputs a scalar grasp reward value. If there exists a push action with a higher 
expected average grasping reward in the predicted next image, the best push action is then 
selected and executed (\autoref{alg:dipn-evaluation-push}); otherwise, the best grasp action 
is selected and executed (\autoref{alg:dipn-evaluation-grasp}). %
Because it is desirable to have a single push action that simultaneously renders multiple 
objects graspable, the average grasp reward is used instead of only the maximum.

\begin{algorithm}
    \small
    \DontPrintSemicolon
    \KwIn{trained \gn and \dipn, discount factor $\gamma$}
    \While{\normalfont there are objects in workspace}{
        $A^{\text{push}} \gets \varnothing$, $M_t \gets$ GetImage\,$()$; \label{alg:dipn-evaluation-input} \; 
        \For{\normalfont $a$ in SamplePushActions\,$(M_t)$}{ \label{alg:dipn-evaluation-sample}
            $A^{\text{push}} \gets A^{\text{push}} \cup \{a\}$; $\hat{M}_{t+1} \gets$ \dipn$(M_t, a)$; \label{alg:dipn-evaluation-dipn}\;
              % \tikzmark{rend}
            $Q(M_t, a) = \frac{\gamma}{|A^{\text{grasp}}|} \sum_{a'\in A^{\text{grasp}}} \text{\gn}(\hat{M}_{t+1}, a')$;
        }
        \If{\normalfont $\max_{a \in A^{\text{push}}} Q(M_t, a) > \frac{1}{|A^{\text{grasp}}|} \sum\limits_{a'\in A^{\text{grasp}}} \text{\gn}(M_t, a')$}{ \label{alg:dipn-evaluation-compare}
            Execute $\argmax_{a \in A^{\text{push}}} Q(M_t, a)$;
            \label{alg:dipn-evaluation-push}
        }
        \lElse{
            Execute $\argmax_{ a\in A^{\text{grasp}}  }\text{\gn}(M_t, a)$; \label{alg:dipn-evaluation-grasp}
        }
    }
    \caption{\label{alg:dipn-evaluation}
    Executing \dipngn
    }
\end{algorithm}

The framework contains two hyperparameters. 
The first one, $\gamma$, is the discount factor of the Markov Decision Process. 
For a push action to be selected, the estimated discounted grasp reward after a 
push must be larger than grasping without a push, since the push and then grasp 
takes two actions. In our implementation, we set $\gamma$ to be $0.9$. 
The other {\em optional} hyperparameter is used for accelerating inference: if 
the maximum grasp reward is higher than a threshold, we directly execute the 
grasp action without calling the push prediction. This hyperparameter requires 
tuning. In our implementation, the threshold value is set to be $0.7$. Note that 
the maximum reward for a single grasp is $1$.

\section{Experimental Evaluation}

We first evaluate \gn and \dipn separately and then compare the full system's 
performance with the state-of-the-art model-free RL technique presented 
in~\cite{zeng2018learning}, which is the closest work to ours. 
Apart from evaluating our approach on a real robotic system, we also conducted extensive evaluations in the CoppeliaSim~\cite{rohmer2013v} simulator. 
We use an Nvidia GeForce RTX 2080 Ti graphics card to train and test the
algorithms. 
All simulation experiments are repeated $30$ times; all real experiments 
are repeated for $5$ times to get the mean metrics.

\subsection{Deep Interaction Prediction Network (\dipn)}
To evaluate how accurately \dipn can predict the next image after a push action,  
\dipn is first trained on randomly generated \pag instances in simulation. 
At the start of each episode, randomly generated objects with random colors 
and shapes are randomly dropped from mid-air to construct the scene. 
Up to $7$ objects are generated per scene.  
We calculate the prediction accuracy by measuring the {\it Intersection-over-Union} 
(IoU) between a predicted image and the corresponding ground-truth after pushing.
The IoU calculation is performed at the object level and then averaged.

\begin{figure}[ht!]
    \centering
    \includegraphics[width = \linewidth]{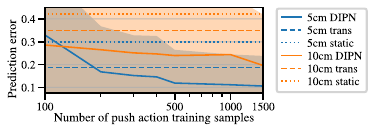}
    \caption{\label{fig:dipn-result-push-prediction}
    \dipn learning curve with standard deviation shown as shaded regions. 
    The $x$-axis is the number of pushes for training \dipn.
    The $y$-axis is the prediction error: $1-\text{IoU}$.
    The dotted and dashed lines are baselines.}
\end{figure}

\dipn is compared with two baselines: 
the first one, called {\em static}, assumes that all objects stay still. 
The second one, called {\em trans}, always assumes that only the pushed object
moves, and that it moves exactly by the push distance along the push direction. 
Both baselines are engineered methods that do not require training. 
The push prediction errors ($1-\text{IoU}$) are illustrated in~\autoref{fig:dipn-result-push-prediction} as learning curves in simulation. 
Mask R-CNN is trained (i.e.,~\autoref{alg:dipn-train},~\autoref{alg:dipn-train-pre}) 
using an additional $100$ images, which is why~\autoref{fig:dipn-result-push-prediction} 
starts from $100$. 
We observe, for different push distances, that \dipn outperforms the baselines 
with a large margin after sufficient training. 
After convergence, the prediction error for \dipn is less than $0.1$ for a $5$cm 
push, which indicates that the predicted pose of an object overlaps $90\%+$ with 
the ground truth. 
As expected, \dipn is more accurate and more sample efficient with a shorter 
push distance. On the other hand, longer push distances generally result in better
overall performance for \pag challenges even though push predictions become 
less accurate, since larger actions are more effective in terms of changing the scene.

\autoref{fig:dipn-push-prediction-result} shows typical predictions by \dipn. 
The network is learned in simulation with randomly shaped and colored objects, and 
directly transferred to the real system. We observe that \dipn can accurately predict 
the state after a push, with good accuracy on object orientation and translation. 

\begin{figure}[ht!]
    % \vspace*{2pt}
    \centering
    \includegraphics[trim={60 60 120 120}, clip, width=0.24\linewidth]{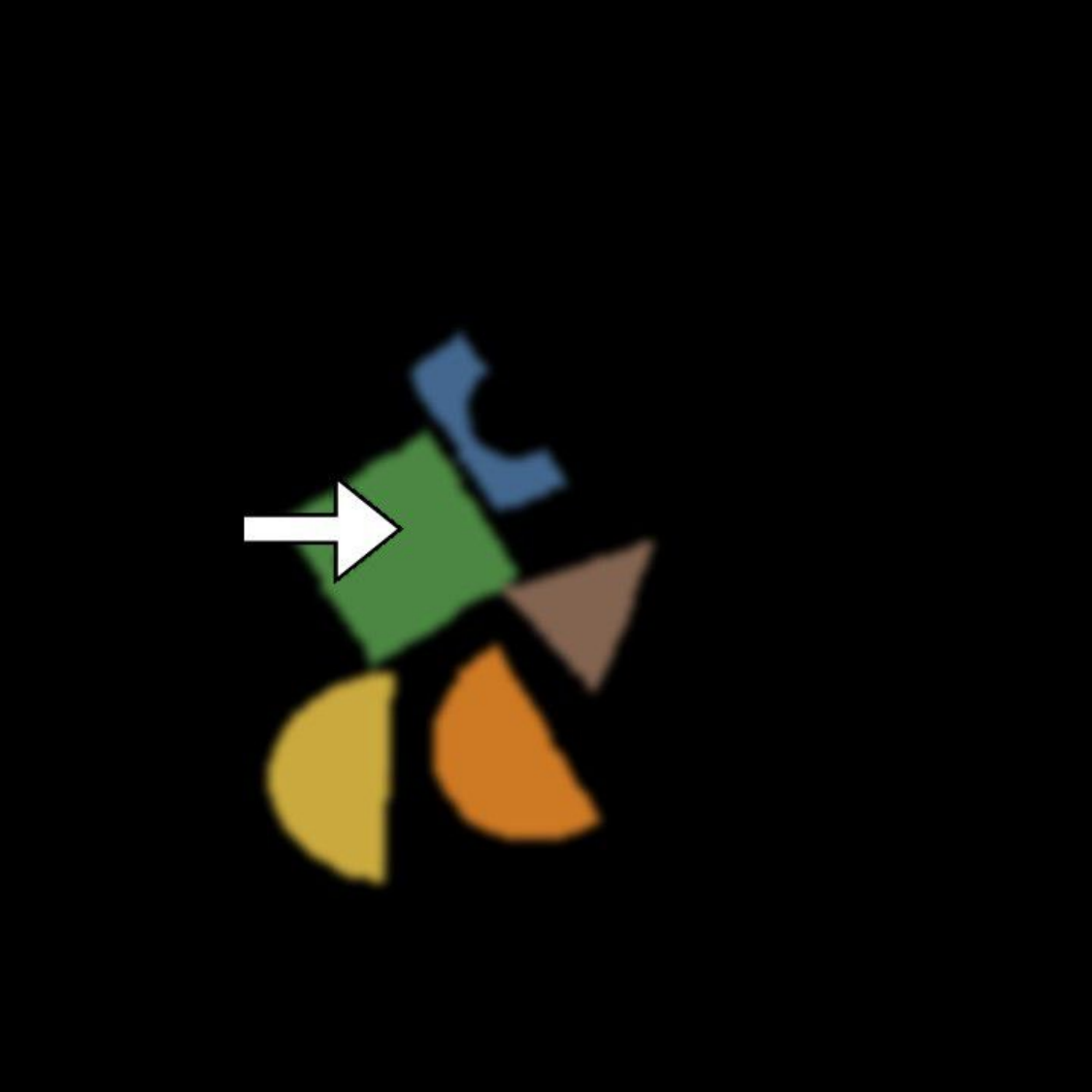}
    \hfill
    \includegraphics[trim={60 60 120 120}, clip, width=0.24\linewidth]{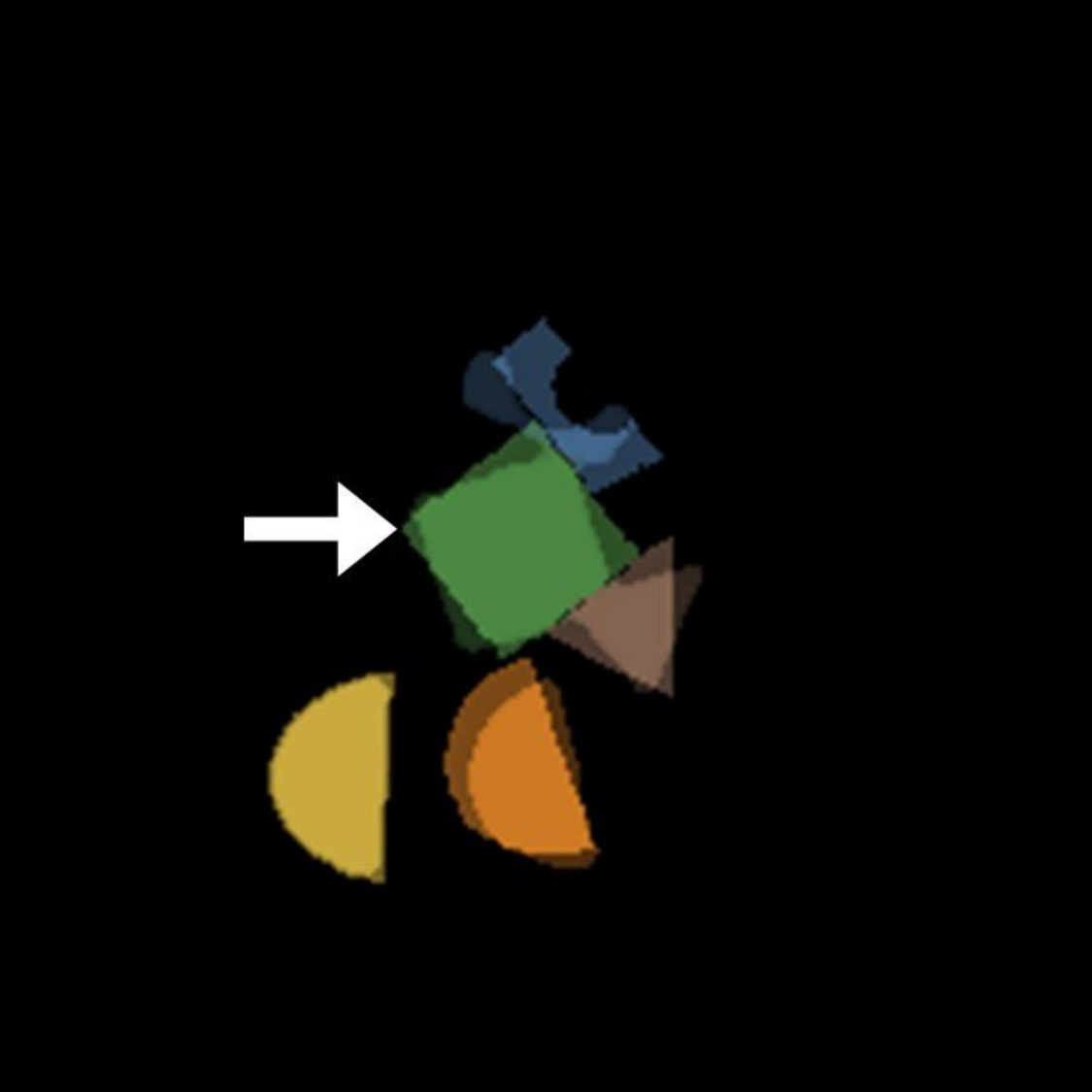}
    \hfill
    \includegraphics[trim={85 120 95 60}, clip, width=0.24\linewidth]{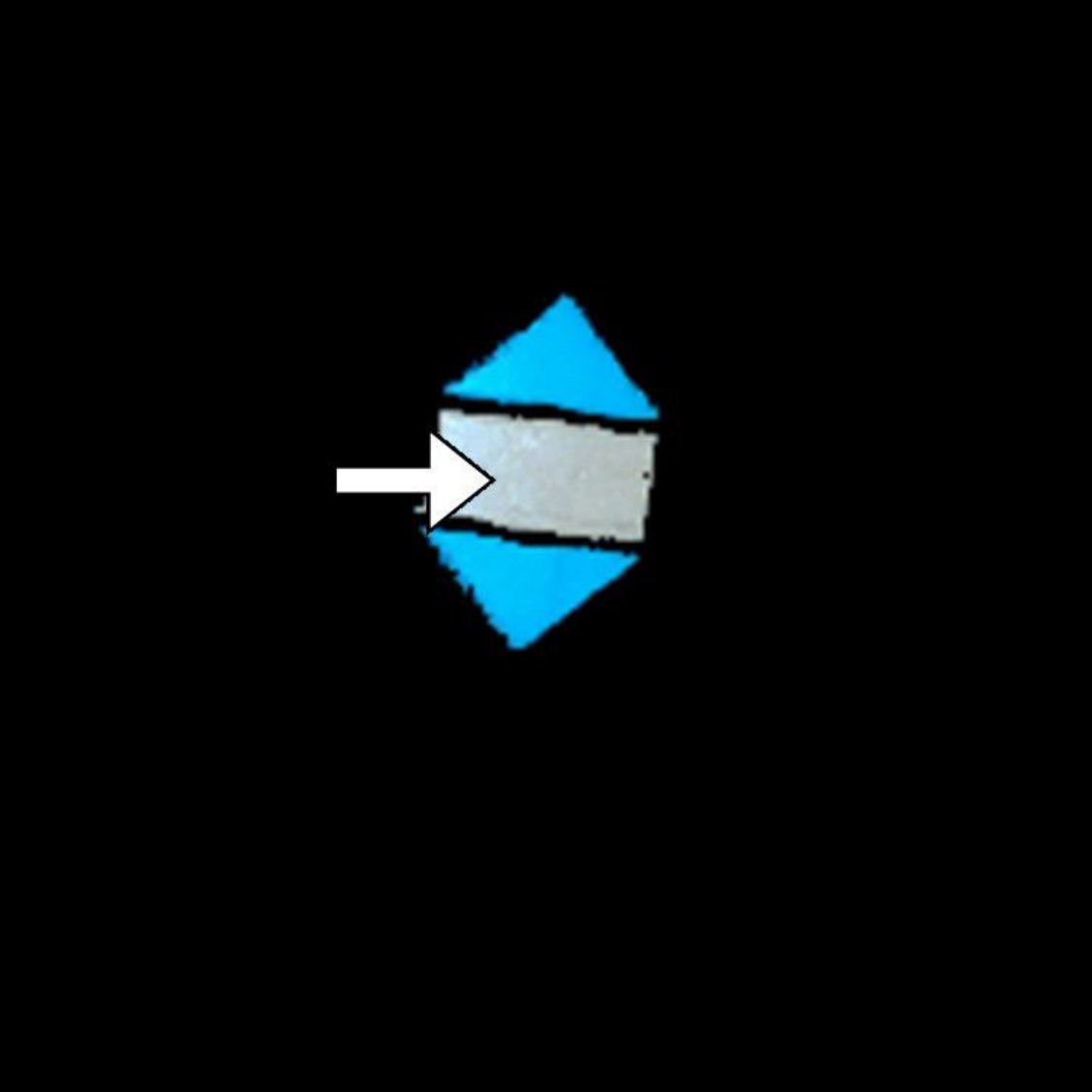}
    \hfill
    \includegraphics[trim={85 120 95 60}, clip, width=0.24\linewidth]{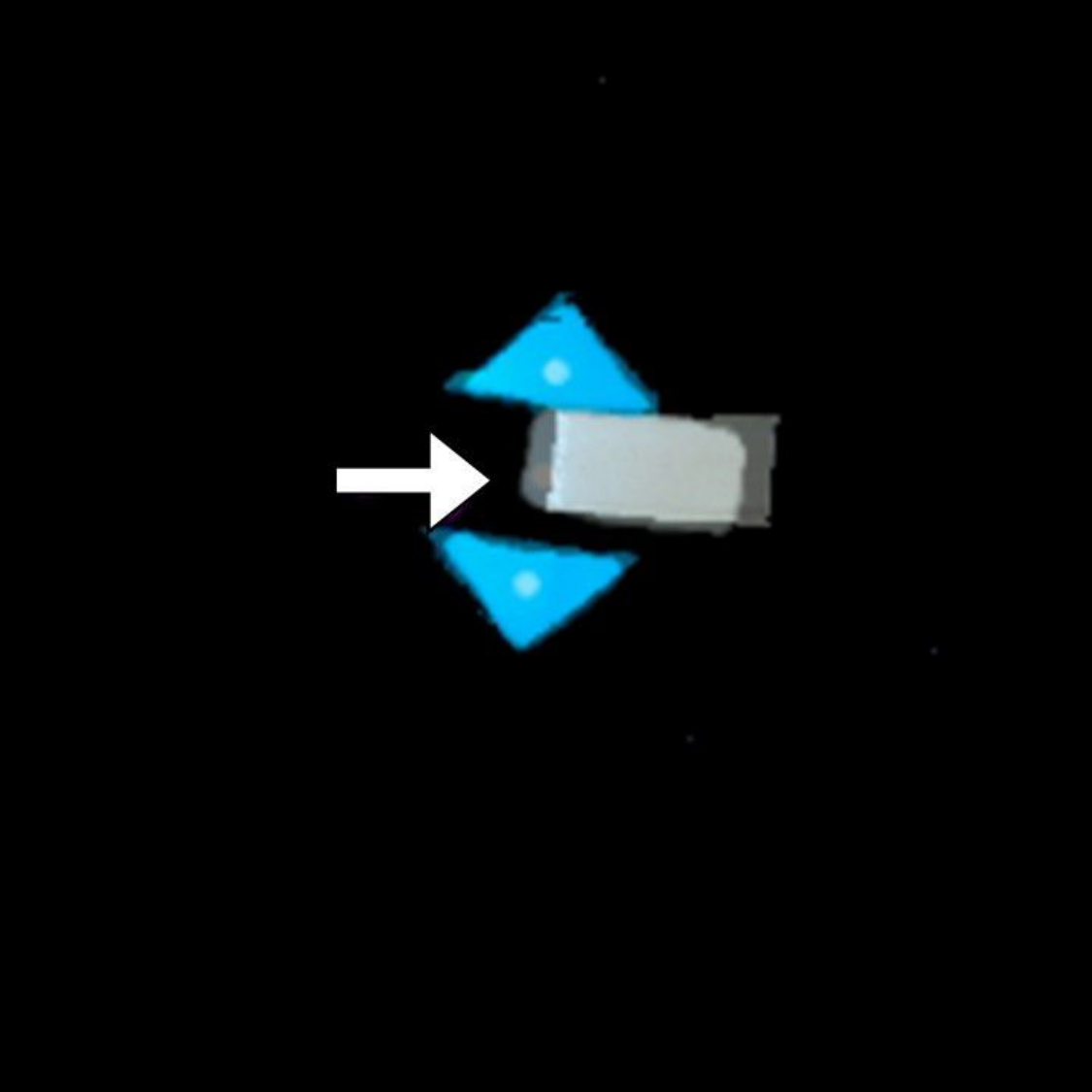}
    \caption{\label{fig:dipn-push-prediction-result} 
    Typical \dipn results. 
    The figures from left to right are: 
    original and predicted images in simulation, and original and predicted images 
    in a real experiment. The ground truth images after a push are overlaid on the
    predicted images with transparency.     The arrows visualize the push actions. 
    }
\end{figure}

\begin{figure*}[ht!]
    \centering
    \includegraphics[width = 0.189 \linewidth, trim = {450, 200, 450, 320}, clip]{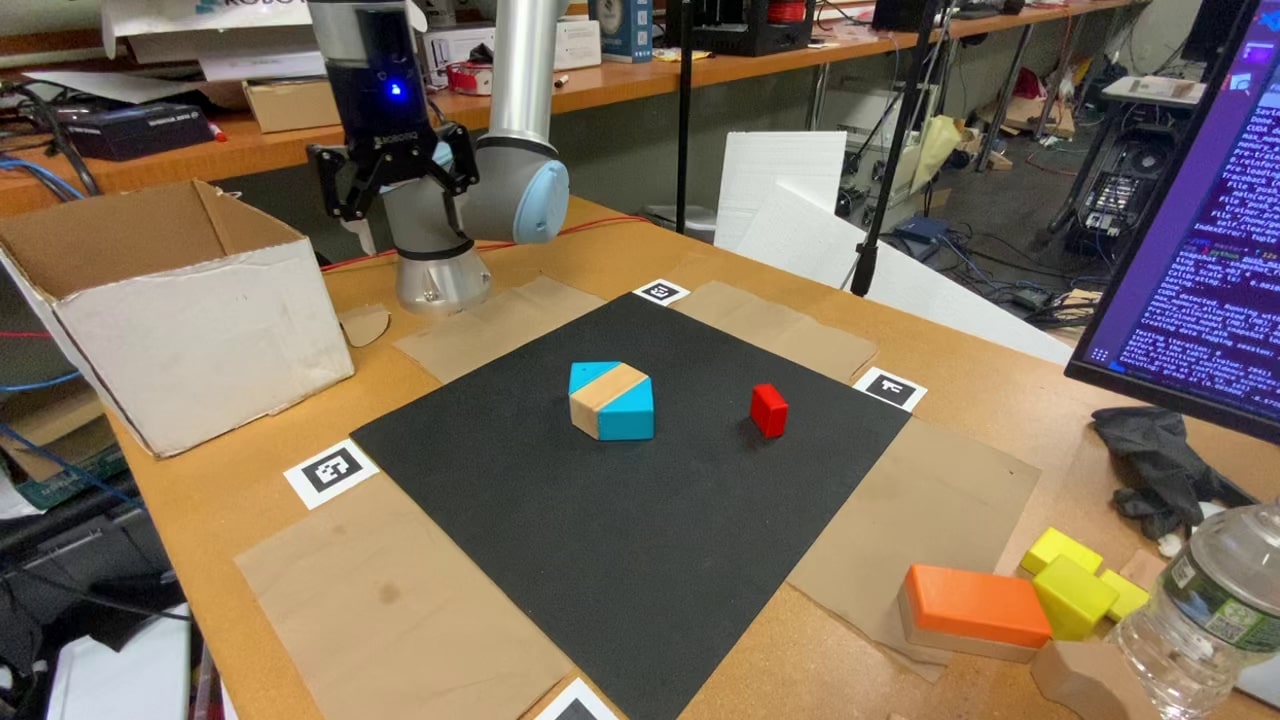} \hfill
    \includegraphics[width = 0.189 \linewidth, trim = {450, 200, 450, 320}, clip]{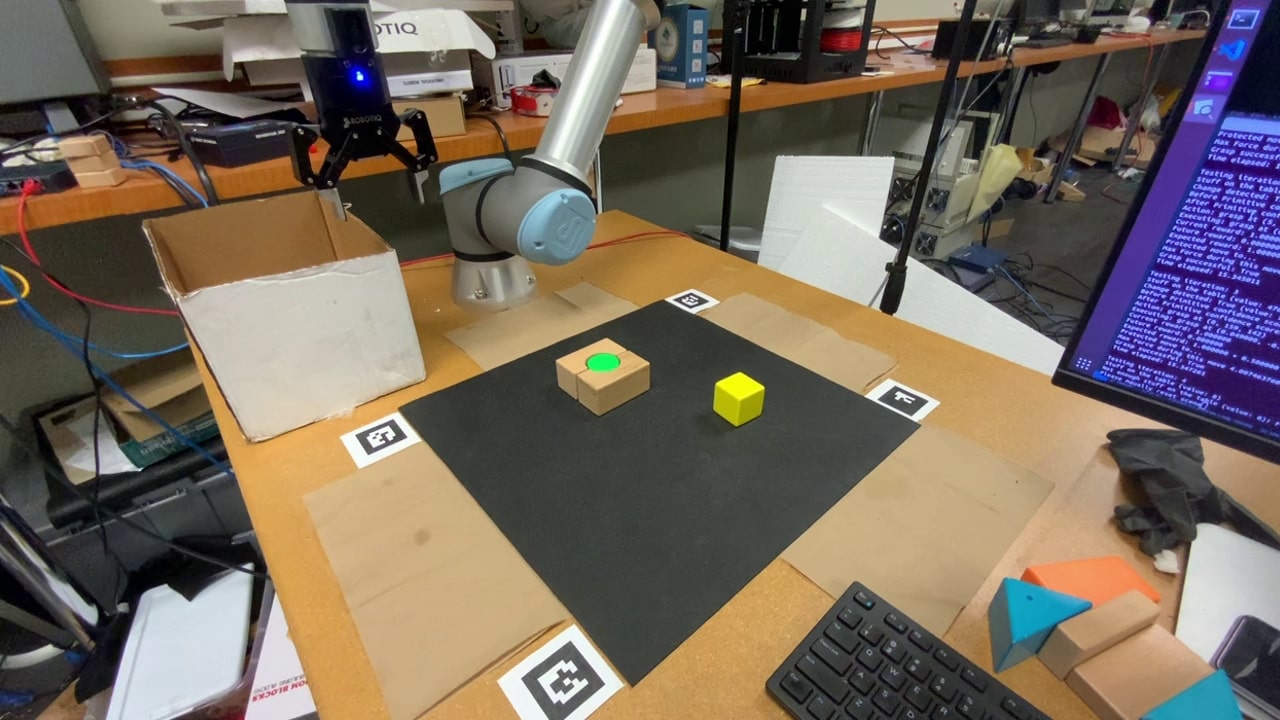} \hfill
    \includegraphics[width = 0.189 \linewidth, trim = {450, 200, 450, 320}, clip]{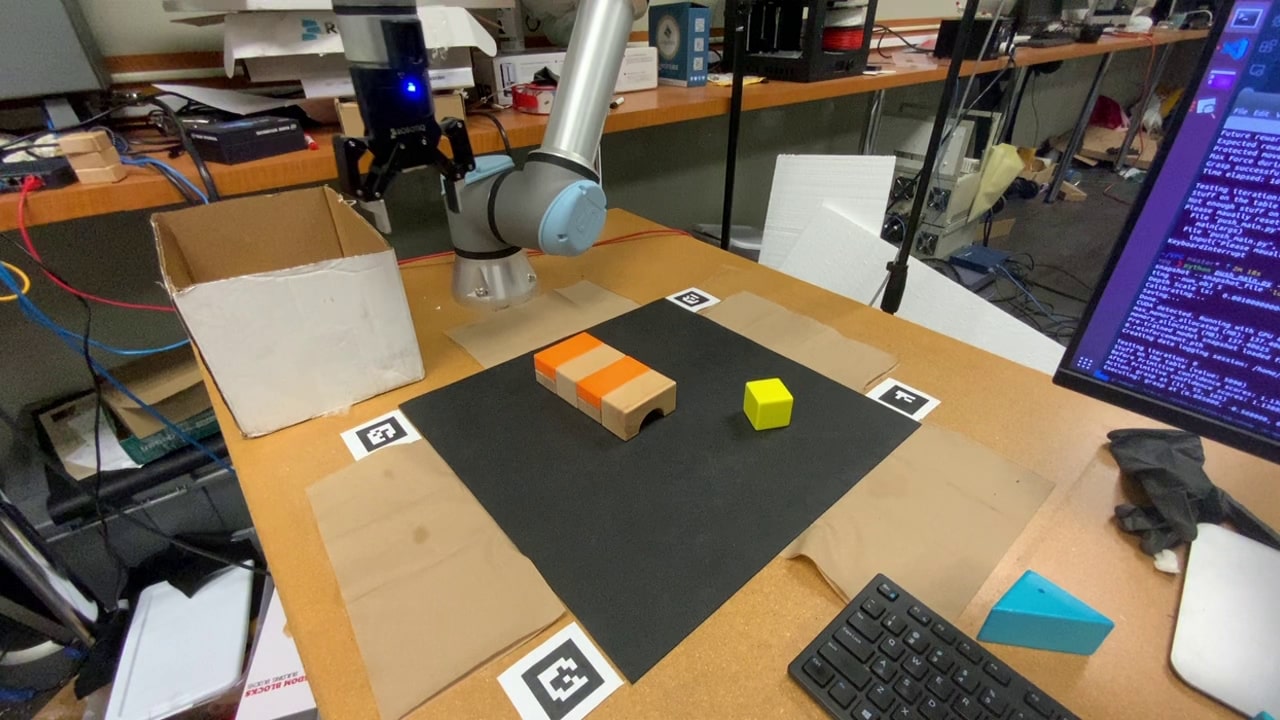} \hfill
    \includegraphics[width = 0.189 \linewidth, trim = {450, 200, 450, 320}, clip]{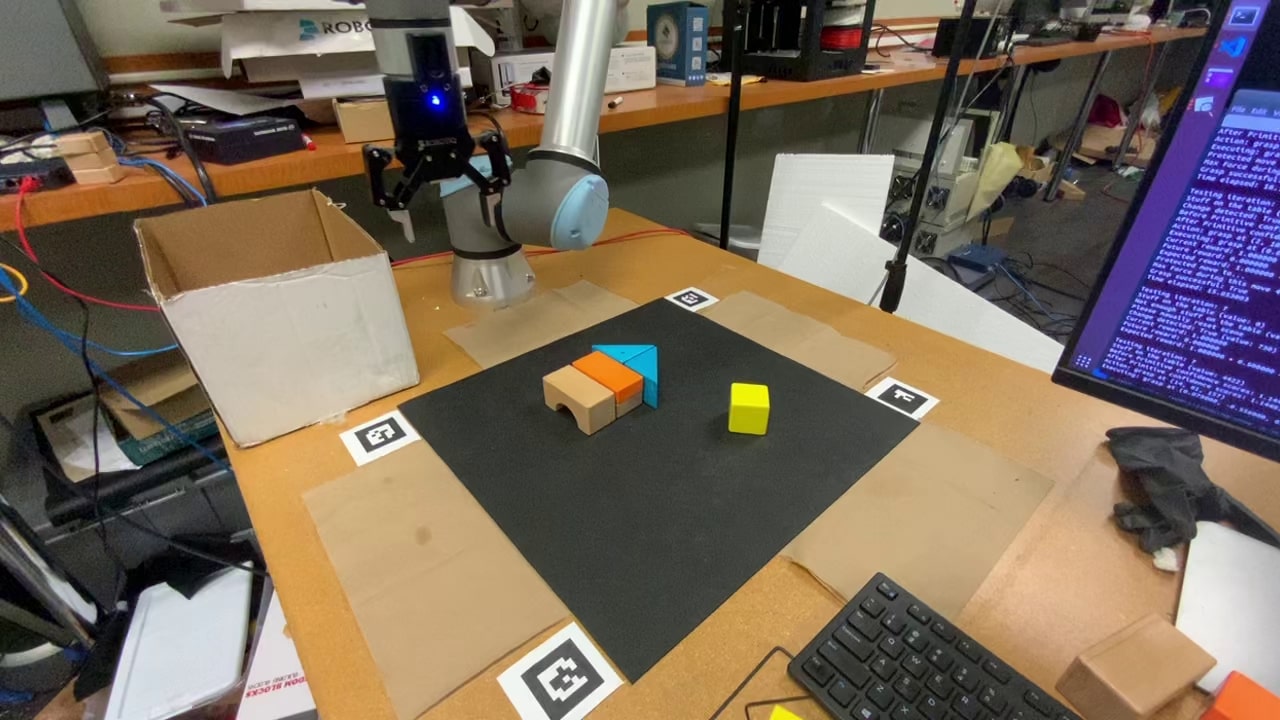} \hfill
    \includegraphics[width = 0.189 \linewidth, trim = {450, 200, 450, 320}, clip]{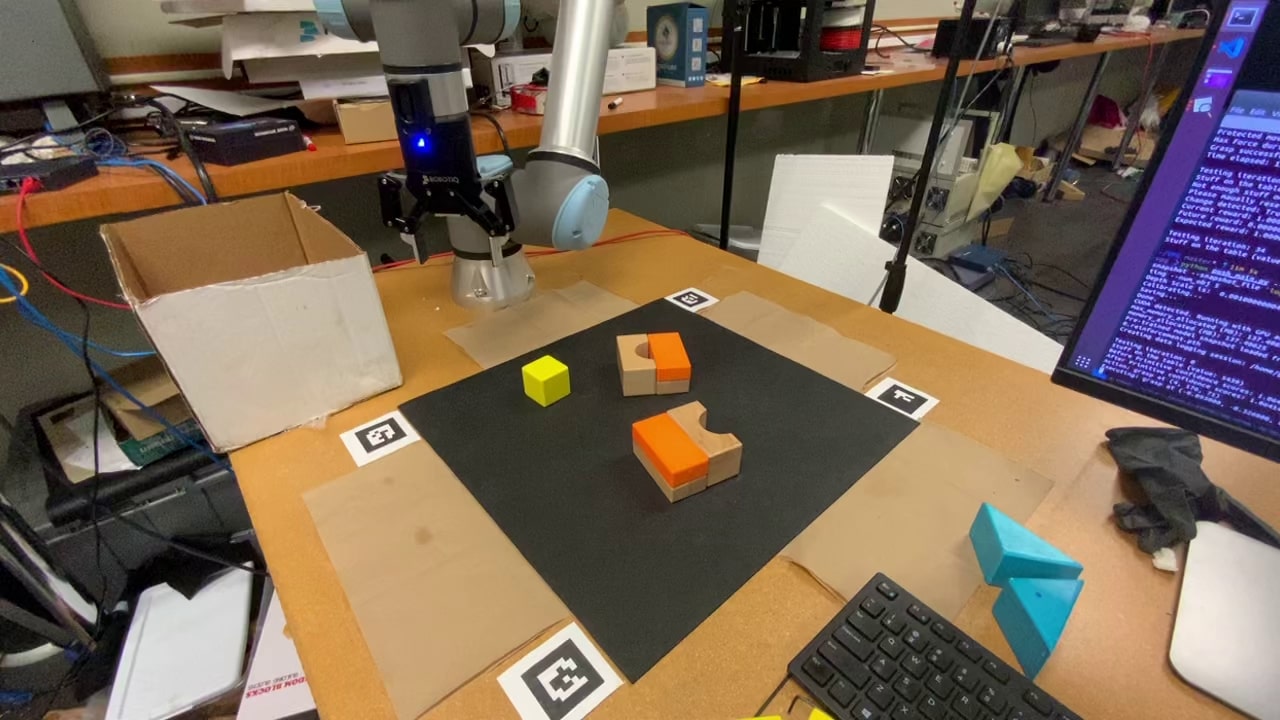}
    
    \vspace*{5pt}
    
    \includegraphics[width = 0.189 \linewidth, trim = {450, 200, 450, 320}, clip]{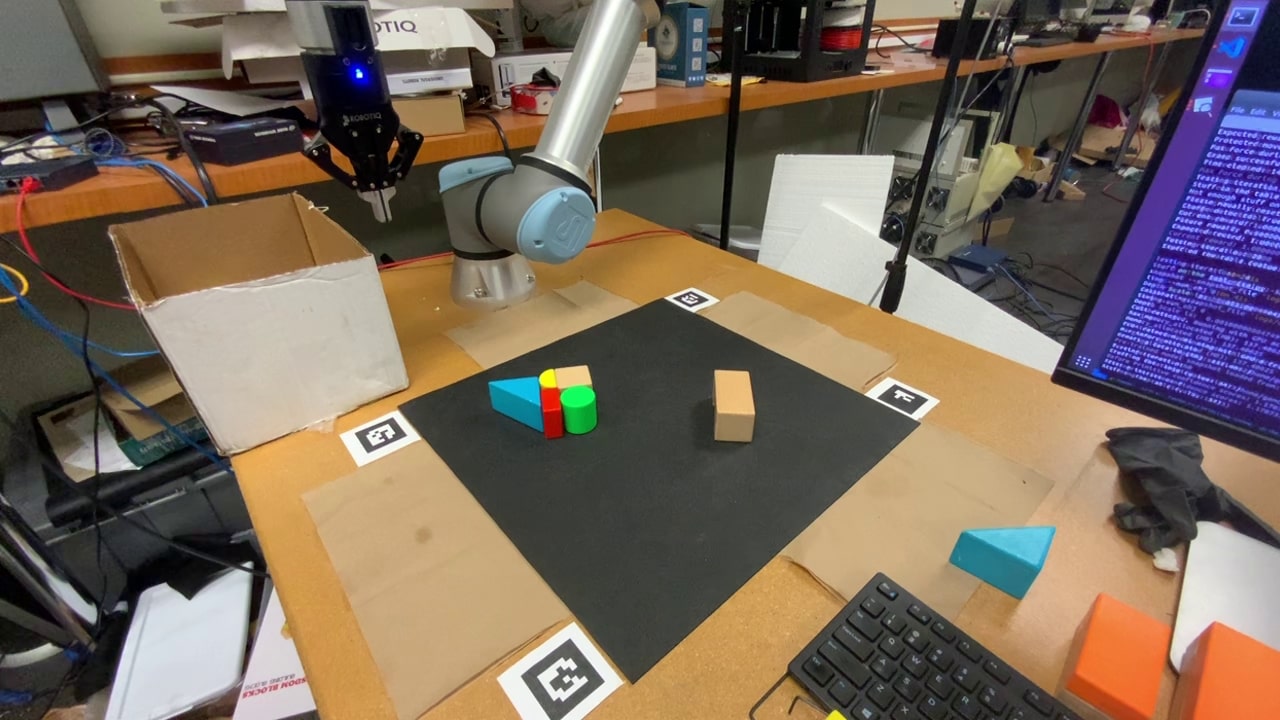} \hfill
    \includegraphics[width = 0.189 \linewidth, trim = {450, 200, 450, 320}, clip]{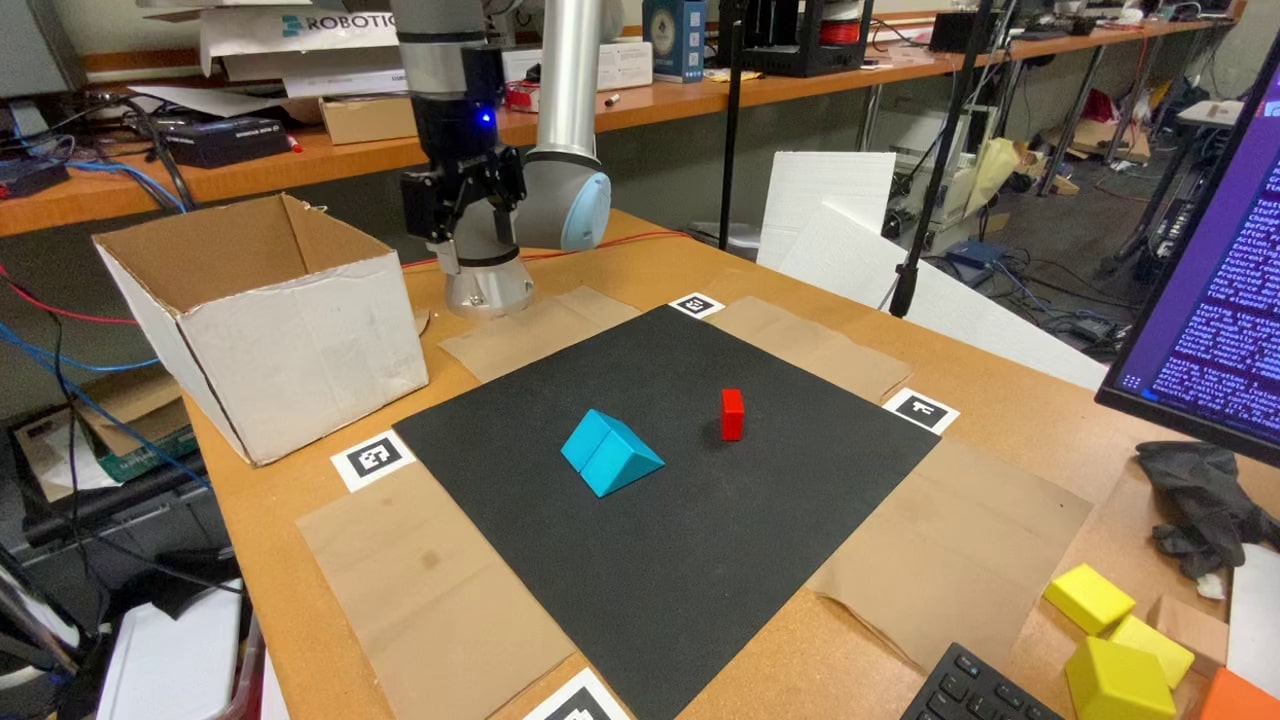} \hfill
    \includegraphics[width = 0.189 \linewidth, trim = {450, 200, 450, 320}, clip]{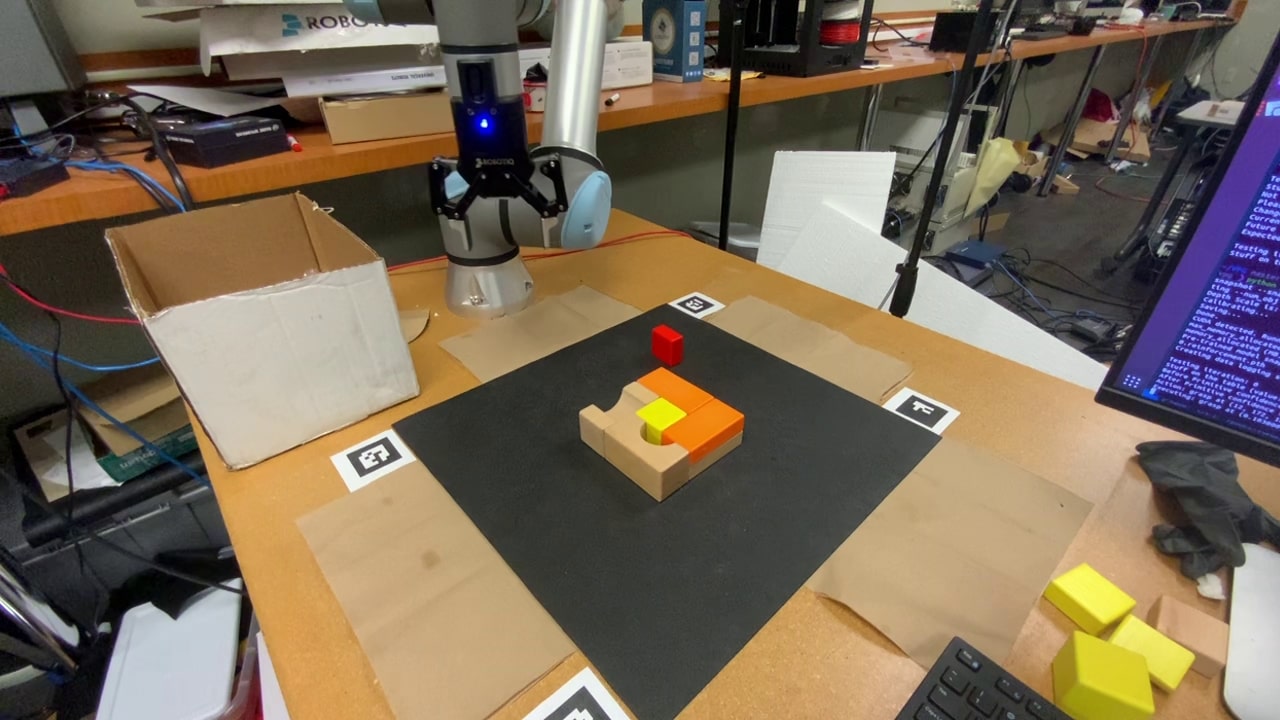} \hfill
    \includegraphics[width = 0.189 \linewidth, trim = {450, 200, 450, 320}, clip]{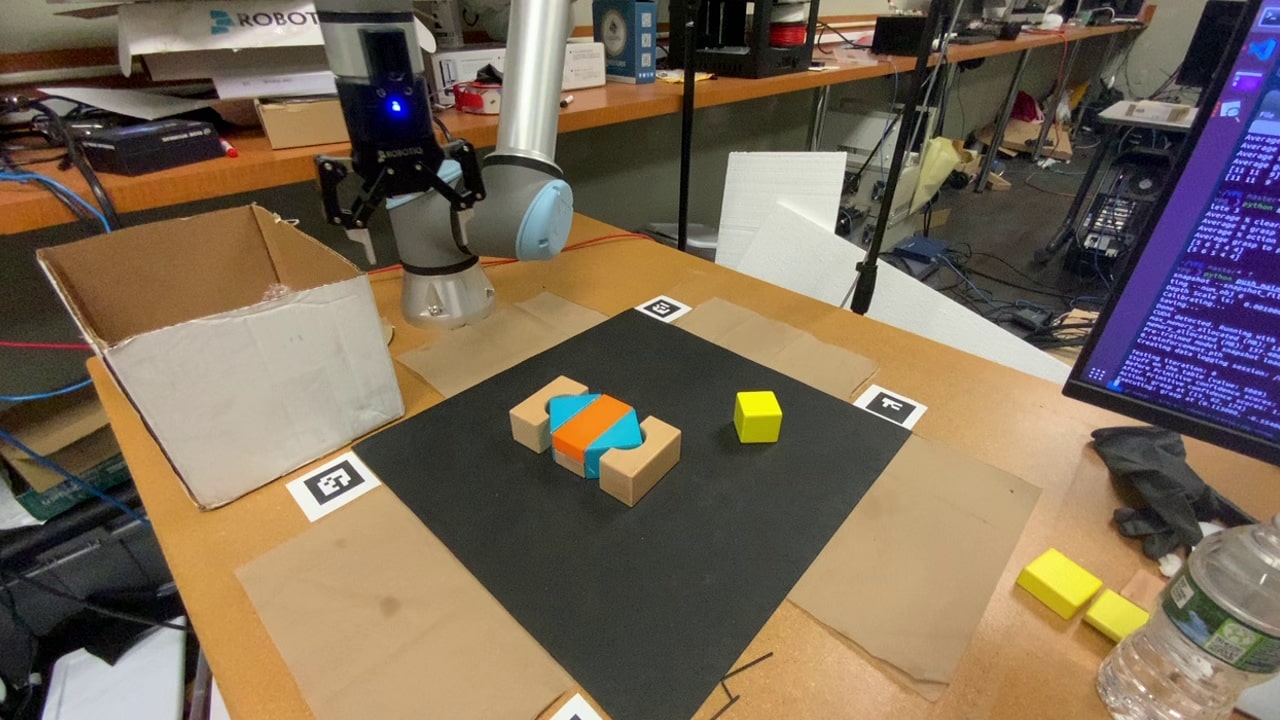} \hfill
    \includegraphics[width = 0.189 \linewidth, trim = {450, 200, 450, 320}, clip]{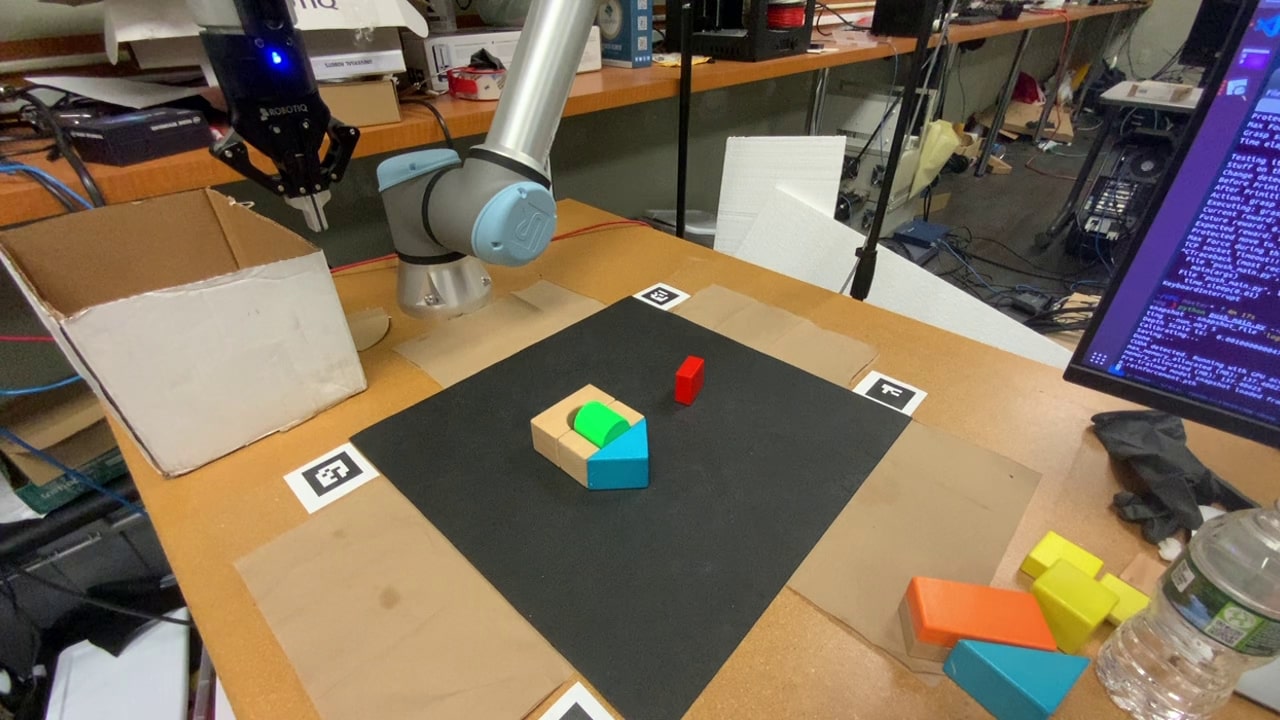}
    \caption{\label{fig:dipn-evaluation-hard-instances}
    Manually generated hard instances largely similar to the ones in \cite{zeng2018learning}. 
    The cases are used in both simulation and real experiment.
    }
\end{figure*}

\subsection{Grasp Network (\gn)}
We train and evaluate \gn as a standalone module and compare it with the state-of-the-art 
DQN-based method known as Visual Pushing and Grasping (VPG)~\cite{zeng2018learning}, 
which learns both grasp and push at the same time. 
Since \gn is only trained on grasp actions, and for a fair comparison, we also tested a
third method that learns both grasp and push actions: this method, denoted by DQN+\gn,
uses \gn for learning grasp actions and the DQN structure of \cite{zeng2018learning} to 
learn push actions. 
The algorithms are compared using the grasp success rate metric, i.e., {\em the number 
of objects removed} divided by {\em the total number of grasps}.
We train all algorithms directly on randomly generated \pag instances with $10$ objects.

The learning curve in simulation is provided in~\autoref{fig:dipn-simulation-learning-curve}. 
The pre-training process (\autoref{alg:dipn-train}, \autoref{alg:dipn-train-pre}, which is also 
self-supervised) for \gn takes $100$ offline images that are not reported in the plot. 
Comparing DQN+\gn which reaches $>90\%$ success rate with less than $300$ (grasp and push) 
samples, and baseline VPG, which converges at $~82\%$ success rate with more than $2000$ 
(grasp and push) samples, it is clear that \gn has significantly higher grasp success 
rate and sample efficiency than the baseline VPG.
As shown by the comparison between \gn and DQN+\gn, when training using only grasp 
actions, \gn can be more sample efficient without sacrificing success rate. 
The result also indicates that for randomly generated \pag, pushing is often \emph{unnecessary}.
 
\begin{figure}[ht!]
    \centering
    \includegraphics[width =  \linewidth]{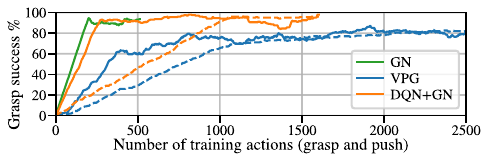}
    % \vspace*{-4pt}
    \caption{\label{fig:dipn-simulation-learning-curve}
    Grasp learning curves of algorithms for \pag in simulation. 
    The $x$-axis is the total number of training steps, i.e., number of actions taken, 
    including push and grasp. The $y$-axis is the grasp success rate.
    The dashed lines denote the success rate for a grasp right after a push action. 
    }
\end{figure}

\subsection{Evaluation of the Complete Pipeline}

We evaluate the learned policies on \pag with up to $30$ objects, first 
in a simulation, then on a real system. 
Four algorithms are tested: VPG~\cite{zeng2018learning}, DQN+\gn, REA+\dipn, and 
\dipngn (our full pipeline). 
Here, REA+\dipn follows~\autoref{alg:dipn-evaluation} but uses the reactive grasp 
network from \cite{zeng2018learning} instead of \gn for grasp reward estimation.
We use three metrics for comparison: 
{\em (i) Completion}, calculated as {\em the number of \pag instances where all 
objects got removed} divided by {\em the total number of instances}; 
incomplete tasks typically occur if objects are pushed out of the workspace, or if the task is not completed within a predefined action limit (e.g., three times the number of objects).
{\em (ii) Grasp success}, calculated as {\em the total number of objects grasped} 
divided by {\em the total number of grasp actions}. 
{\em (iii) Action efficiency}, calculated as {\em the total number of objects 
removed} divided by {\em the total number of actions (grasp and push)}. 
For grasp success rate and action efficiency, we use two formulations: 
one does not count incomplete tasks (reported in gray text), which is the same as 
the one used in~\cite{zeng2018learning}, and the other one counts incomplete tasks, 
which we believe is more reflective. The robot could grasp more than one object at a time. 
We consider it as a successful grasp, as the goal is to clear all objects from the table.

\autoref{table:dipn-pag-sim} reports simulation results on $30$ randomly generated 
\pag instances and $10$ manually placed hard instances (illustrated in~\autoref{fig:dipn-evaluation-hard-instances}) where push actions are necessary. 
The algorithms were not trained on these hard instances. 
The number of training samples for each algorithm are: 
$2500$ actions (grasp and push) for VPG~\cite{zeng2018learning}, 
$1500$ grasp actions and $2000$ push actions for REA+\dipn, 
$1500$ actions (grasp and push) for DQN+\gn,
and $500$ grasp actions and $1500$ push actions for \dipngn (our full pipeline). 
The results show that \dipn and \gn both are sample efficient in 
comparison with the baseline and provide significant improvement 
in \pag metrics; when combined, \dipngn reaches the highest performance 
on all metrics.

We repeated the evaluation on a real system (see.~\autoref{fig:dipn-system-setup}).
Each random instance contains $10$ randomly selected objects; the hard instances are shown 
in~\autoref{fig:dipn-evaluation-hard-instances}. 
\autoref{fig:dipn-real-learning-curve} shows grasp learning curve. 
We compare VPG~\cite{zeng2018learning} (trained with $2000$ grasp and push actions) and the 
proposed \dipngn pipeline (pre-trained with $100$ unlabeled RGB-D images for segmentation, 
trained \gn with $500$ grasp actions and \dipn with $1500$ simulated push actions). 
The evaluation result is reported in~\autoref{table:dipn-pag-real}. Remarkably, our networks, 
while being developed using only simulation based training, perform even better when 
trained/evaluated only on real hardware. 

\begin{figure}[ht!]
    \centering
    \includegraphics[width = \linewidth]{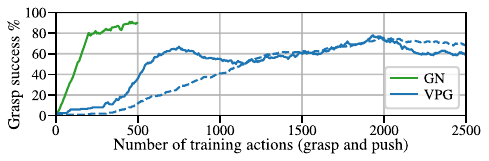}
    % \vspace*{-4pt}
    \caption{\label{fig:dipn-real-learning-curve} 
    Grasp learning curves for \pag in real experiment. 
    Solid lines indicate grasp success rate and dotted lines indicate push-then-grasp 
    success rates over training steps. The GN is trained in a grasp only manner.
    }
\end{figure}

\begin{table}[ht!]
    \centering
    % \footnotesize
    %\captionsetup{justification=centering}
    \caption{\label{table:dipn-pag-sim} Simulation, random and hard instances (mean $\%$)}
    \begin{tabular}{l|l|l|p{0.75cm}|p{0.75cm}|p{0.9cm}|p{0.9cm}}
    \hline
    & Method & Completion & \multicolumn{2}{c|}{Grasp success} & \multicolumn{2}{c}{Action efficiency} \\ \hline \hline
    \multirow{4}{*}{\;Rand\;\;}
    & VPG\cite{zeng2018learning}            & 20.0\footnotemark          
        & \textcolor{gray}{69.0}            & 52.6          
        & \textcolor{gray}{66.3}            & 52.6             \\ \cline{2-7}
    & REA\cite{zeng2018learning}+\dipn      & 83.3          
        & \textcolor{gray}{79.5}            & 77.9           
        & \textcolor{gray}{77.4}            & 76.3            \\ \cline{2-7}
    & DQN\cite{zeng2018learning}+\gn        & 46.7          
        & \textcolor{gray}{85.2}            & 83.9           
        & \textcolor{gray}{83.4}            & 81.7             \\ \cline{2-7}
    & \dipngn                               & \textbf{83.3} 
        & \textcolor{gray}{\textbf{86.7}}   & \textbf{85.2} 
        & \textcolor{gray}{\textbf{84.4}}   & \textbf{83.3}    \\ \hline \hline
    \multirow{4}{*}{\;Hard\;\;}    
    & VPG\cite{zeng2018learning}            & 77.7          
        & \textcolor{gray}{67.4}            & 60.0          
        & \textcolor{gray}{60.8}            & 57.6             \\ \cline{2-7}
    & REA\cite{zeng2018learning}+\dipn      & 90.3          
        & \textcolor{gray}{81.5}            & 76.6           
        & \textcolor{gray}{64.7}            & 62.6             \\ \cline{2-7}
    & DQN\cite{zeng2018learning}+\gn        & 86.0          
        & \textcolor{gray}{91.1}            & 87.1           
        & \textcolor{gray}{70.2}            & 67.9             \\ \cline{2-7}
    & \dipngn                               & \textbf{100.0} 
        & \textcolor{gray}{\textbf{93.3}}   & \textbf{93.3} 
        & \textcolor{gray}{\textbf{74.4}}   & \textbf{74.4}          \\ \hline
    \end{tabular}
\end{table}
\footnotetext{{The low completion rate is primarily due to pushing objects outside of the workspace. }}

\begin{table}[ht!]
    \centering
    % \vspace*{-3pt}
    \caption{\label{table:dipn-pag-real} 
    Real system, random and hard instances (mean $\%$)}
    \begin{tabular}{l|l|l|p{0.75cm}|p{0.75cm}|p{0.9cm}|p{0.9cm}}
    \hline
    & Method & Completion & \multicolumn{2}{c|}{Grasp success} & \multicolumn{2}{c}{Action efficiency} \\ \hline \hline
    \multirow{2}{*}{Rand\;}
    & VPG\cite{zeng2018learning}          & 80.0        
        & \textcolor{gray}{85.5}            & 79.0        
        & \textcolor{gray}{75.3}            & 67.9           \\ \cline{2-7}
    & \dipngn                             & \textbf{100.0} 
        & \textcolor{gray}{\textbf{94.0}}   & \textbf{94.0} 
        & \textcolor{gray}{\textbf{98.2}}   & \textbf{98.2}          \\ \hline \hline
    \multirow{2}{*}{Hard\;}
    & VPG\cite{zeng2018learning}          & 64.0        
        & \textcolor{gray}{75.1}            & 69.0         
        & \textcolor{gray}{51.9}            & 47.8           \\ \cline{2-7}
    & \dipngn  \;\;\;\;\;\;                           & \textbf{98.0\footnotemark} 
        & \textcolor{gray}{\textbf{89.9}}   & \textbf{89.9} 
        & \textcolor{gray}{\textbf{77.6}}   & \textbf{78.2}          \\ \hline
    \end{tabular}
\end{table}
\footnotetext{The single failure was due to an object that was successfully grasped 
but slipped out of the gripper before the transfer was complete.}

With \dipn and \gn outperforming the corresponding components from VPG~\cite{zeng2018learning}, 
it is unsurprising that \dipngn does much better. 
In particular, \dipn architecture allows it to learn intelligent, graded push behavior 
efficiently. In contrast, VPG~\cite{zeng2018learning} has a fixed $0.5$ push reward,
which sometimes negatively impacts performance: VPG could push unnecessarily for 
many times without a grasp when it is not confident enough to grasp. It also risks 
pushing objects outside of the workspace.
Following~\cite{zeng2018learning}, we tested \dipngn with previously unseen objects, such as soapboxes and plastic bottles. Our method maintained a similar level of performance to that reported in \autoref{table:dipn-pag-real}.

\section{Summary}
In this work, we have developed a Deep Interaction Prediction Network (\dipn) for learning 
to predict the complex interactions that occur as a robot manipulator pushes objects in 
clutter. Unlike most existing end-to-end techniques, \dipn is capable of generating accurate 
predictions in the form of clearly legible synthetic images that can be fed as inputs to a 
deep Grasp Network (\gn), which can then predict successes of future grasps. We demonstrated 
that \dipn, \gn, and \dipngn all have excellent sample efficiency and significantly 
outperform the previous state-of-the-art learning-based method for \pag challenges, while
using only a fraction of the interaction data used by the alternative. Our networks are 
trained in a fully self-supervised manner, without any manual labeling or human inputs, 
and exhibit high levels of generalizability. 
The proposed system, initially developed in simulation, also performs effectively when trained and deployed on real hardware with physical objects. \dipngn demonstrates high robustness to variations in object properties such as shape, size, color, and friction.
    
\chapter{Visual Foresight Trees for Object Retrieval from Clutter with Nonprehensile Rearrangement}\label{chap:vft}
\thispagestyle{myheadings}

\newcommand{\pnp}{\textsc{PnP}\xspace}
\newcommand{\orc}{\textsc{ORC}\xspace}
\newcommand{\dqn}{\textsc{DQN}\xspace}
\newcommand{\vft}{\textsc{VFT}\xspace}
\newcommand{\mcts}{\textsc{MCTS}\xspace}
\newcommand{\uct}{\textsc{UCT}\xspace}
\def\gcvpg{gc-\textsc{VPG}\xspace}
\def\gopg{go-\textsc{PGN}\xspace}

\section{Introduction}

In many application domains, robots are tasked with retrieving objects that are surrounded by multiple tightly packed objects. 
To enable the grasping of target object(s), a robot needs to rearrange the scene to create sufficient clearance before attempting a grasp.
Scene rearrangement can be achieved through nested sequential push actions, each moving multiple objects simultaneously.
In this paper, we address the problem of finding the minimum number of push actions to create a scene where the target object can be grasped and retrieved. 

To solve the object retrieval problem, the robot must imagine how the scene would look after any given sequence of pushing actions, and select the shortest 
sequence that leads to a state where the target object can be grasped. 
The huge combinatorial search space makes this problem computationally challenging, 
hence the need for efficient planning algorithms, as well as fast predictive models 
that can return the predicted future states in a few milliseconds. 
Moreover, objects in clutter typically have unknown physical properties such as mass 
and friction coefficients. While it is possible to utilize off-the-shelf physics engines
to simulate contacts and collisions of rigid objects in clutter, simulation is highly sensitive to the accuracy of the provided mechanical parameters. To overcome the 
problem of manually specifying these parameters, and to enable full autonomy of the 
robot, most recent works on object manipulation utilize machine learning techniques 
to train predictive models from data~\cite{Hafner2020Dream,DBLP:journals/corr/abs-1812-00568,8207585}. The predictive models take the state of the robot's environment a control action as inputs and predict the state after applying the control action.

\begin{figure}[t]
    \centering
    \begin{minipage}{.4\linewidth}
        \subfloat[Hardware setup]{\includegraphics[width = \linewidth, trim = 10 110 30 48, clip]{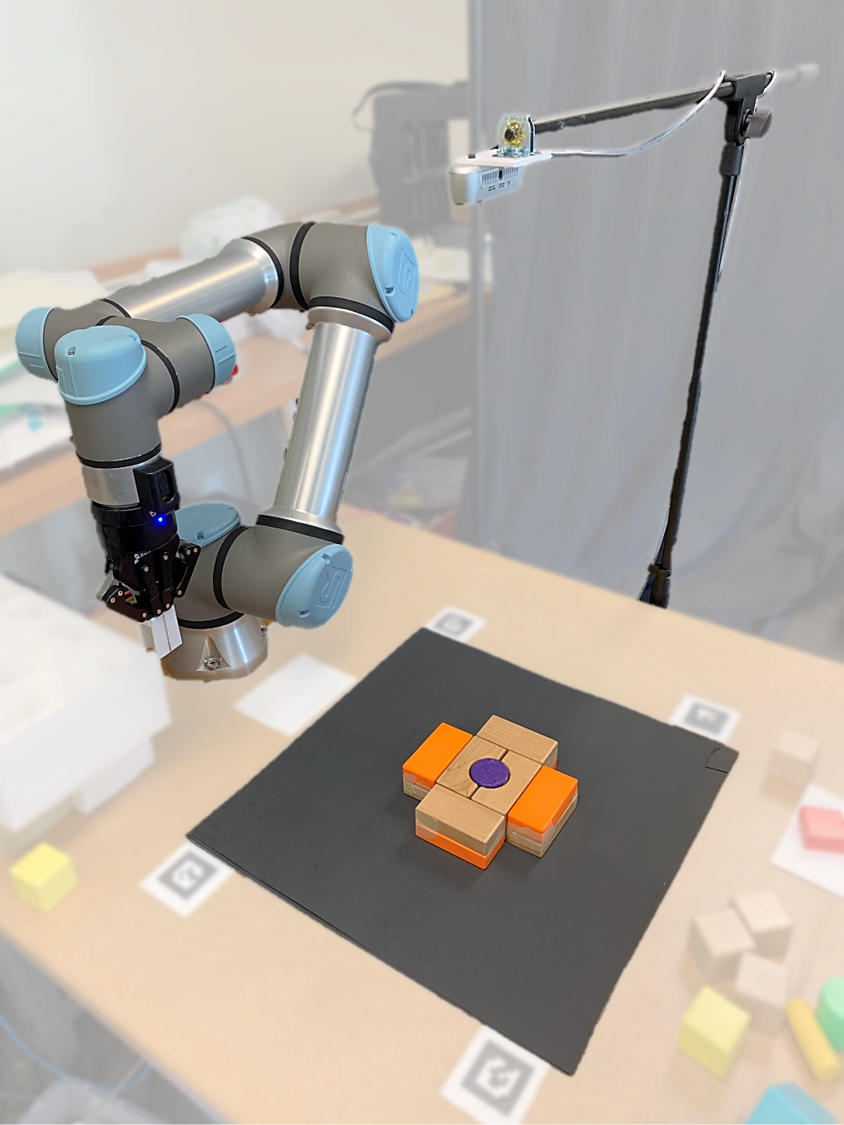}\label{fig:vft-intro-setup}}
    \end{minipage}
    \begin{minipage}{.28\linewidth}
        \subfloat[First push]{\includegraphics[width = \linewidth, trim = 0 0 0 0, clip]{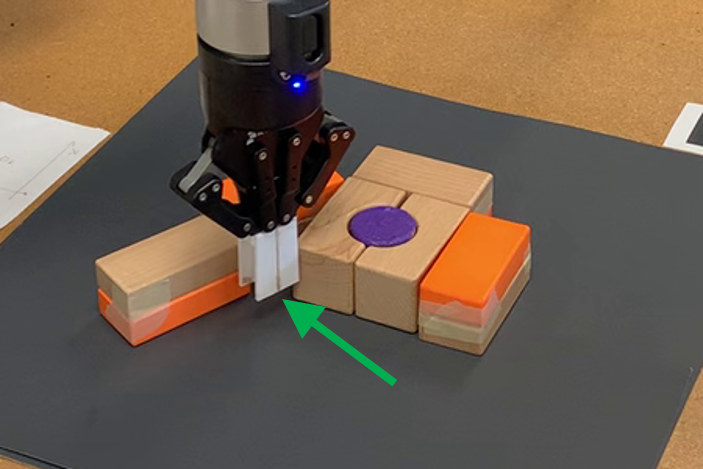}\label{fig:vft-intro-push}}
        
        \subfloat[Second push]{\includegraphics[width = \linewidth, trim = 0 0 0 0, clip]{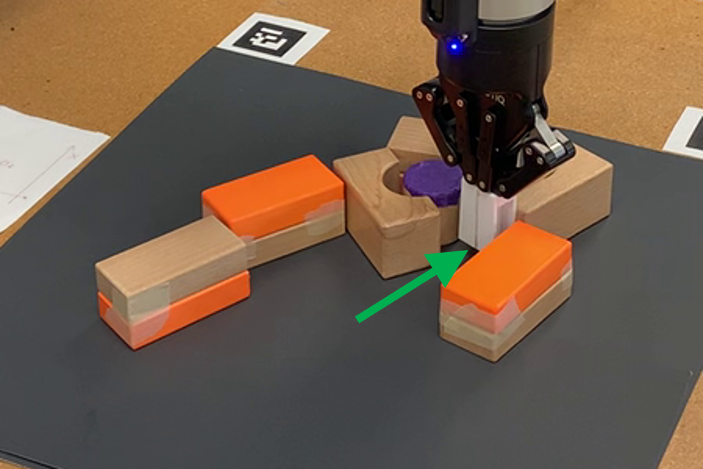}} 
    \end{minipage} 
    \begin{minipage}{.28\linewidth}
        \subfloat[Third push]{\includegraphics[width = \linewidth, trim = 0 0 0 0, clip]{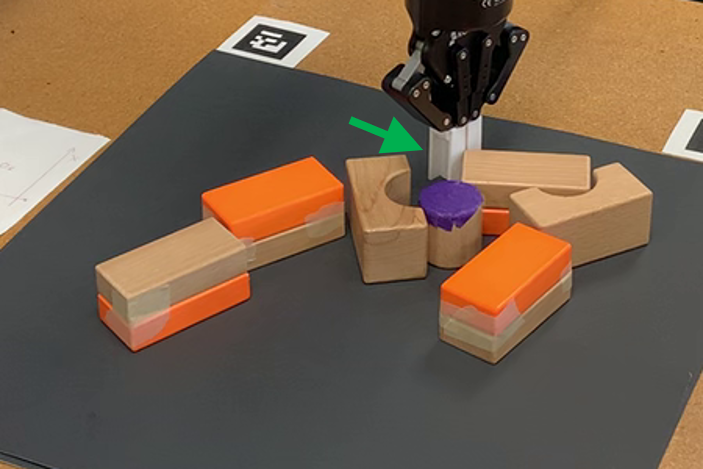}}
        
        \subfloat[Grasp]{\includegraphics[width = \linewidth, trim = 0 0 0 0, clip]{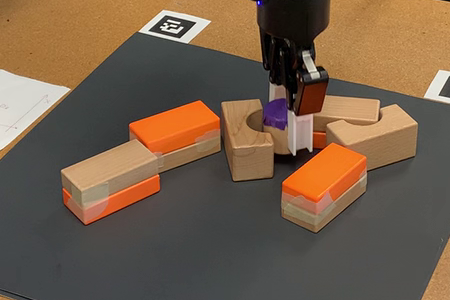}\label{fig:vft-intro-grasp}}
    \end{minipage} 

    \caption{\label{fig:vft-intro}
    (a) The hardware setup for object retrieval in a clutter includes 
    a Universal Robots UR-5e manipulator 
    with a Robotiq 2F-85 two-finger gripper, 
    and an Intel RealSense D435 RGB-D camera. 
    The objects are placed in a square workspace. 
    (b), (c), (d) Three sequential push actions (green arrows) create space to access the target (purple) object. 
    The push directions are toward top-left, top-right, 
    and bottom-right, respectively. 
    (e) The target object is successfully grasped and retrieved. 
    }
    % \vspace{-3mm}
\end{figure}

In this work, we propose to employ {\it visual foresight trees} (VFT) to address the computational and modeling challenges related to the object retrieval problem. 
A key building block of VFT is a Convolutional Neural Network (CNN) extending DIPN~\cite{huang2021dipn}, capable of predicting multi-step push outcomes involving multiple objects.
A second CNN evaluates the graspability of the target object in predicted future images. 
A Monte Carlo Tree Search utilizes the two CNNs to obtain the shortest sequence of pushing actions that lead to an arrangement where the target can be grasped. 

To our knowledge, the proposed technique is the first model-based learning solution to the object retrieval problem.
Extensive experiments on a real robot with physical objects, as exemplified in~\autoref{fig:vft-intro}, demonstrate that the proposed approach succeeds in retrieving target objects with manipulation sequences that are shorter than model-free reinforcement learning techniques and a limited-horizon planning technique.

\section{Problem Formulation}

\subsection{Problem Statement}

The Object Retrieval from Clutter (\orc) challenge asks a robot manipulator to retrieve a target object from a set 
of objects densely packed together. The objects may have different shapes, sizes, 
and colors. 

Objects other than the target object are unknown a prior.  
Focusing on a mostly planar setup, the following assumptions are made:
\begin{enumerate}
\item The hardware setup (\autoref{fig:vft-intro-setup}) contains a manipulator, a 
planar workspace with a uniform background color, and a camera on top of the 
workspace. 
\item The objects are rigid and are amenable to the gripper's prehensile 
and non-prehensile capabilities, limited to straight-line planar push actions and top-down grasp actions.
%with fixed distances
%
\item The objects are confined to the workspace without overlapping. As a result, 
the objects are visible to the camera. 
\item The target object, to be retrieved, is visually distinguishable from the others. 
\end{enumerate}
Under these assumptions, the \emph{objective} is to retrieve only the target 
object, while minimizing the number of pushing/grasping actions that are used. Each grasp or push is considered as one atomic action. While a mostly planar setup is assumed in our experiments, the proposed data-driven solution is general and can be applied to arbitrary object shapes and arrangements. In the experiments, we mainly work with woodblocks; we also evaluate the proposed approach on novel objects such as soapboxes, which are challenging as their widths are 
close to the maximum distances between the gripper's fingers.

\begin{figure*}[ht!]
    \centering
    \includegraphics[width = \linewidth]{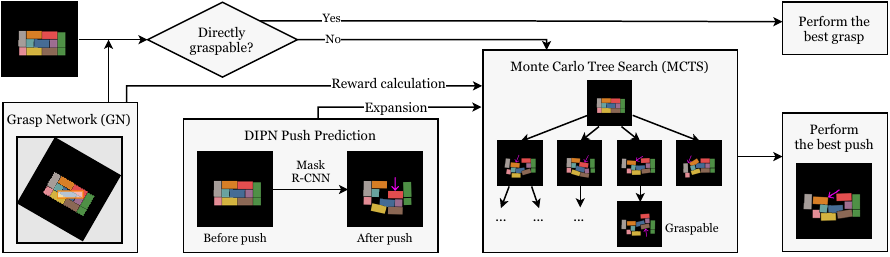}\label{fig:pipeline}
    \caption{\label{fig:vft-pipeline}
        Overview of the proposed technique for object retrieval from clutter with nonprehensile rearrangement. The problem is iteratively solved by observing the environment at each time step, taking the current state as input, and returning the best action. It is repeated until the object is retrieved.
    }
\end{figure*}

\subsection{Manipulation Motion Primitives}
Similar to studies closely related to the \orc challenge, e.g.,~\cite{zeng2018learning, huang2021dipn, xu2021efficient}, we employ a set of pre-defined and parameterized 
pushing/grasping manipulation primitives. The decision-making problem then entails 
the search for the optimal order and parameters of these primitives.
A grasp action $a^\text{grasp} = (x, y, \theta)$ is defined as 
a top-down overhead grasp motion at image pixel location $(x, y)$, 
with the end-effector rotated along with the world $z$-axis by $\theta$ degrees. 
In our implementation, a grasp center $(x, y)$ can be any pixel in a 
down-sampled $224 \times 224$ image of the planar scene, while rotation angle 
$\theta$ can be one of $16$ values evenly distributed between $0$ and $2\pi$. 
To perform a complete grasp action, the manipulator moves the open gripper 
above the specified location, then moves the gripper downwards until a contact 
with the target object is detected, closes the fingers, and transfers the 
grasped object outside of the workspace. 
%An object that is successfully grasped is considered as retrieved. 

When objects are densely packed, the target object is generally not directly 
graspable due to collisions between the gripper and surrounding objects. 
When this happens, non-prehensile push actions can be used to create 
opportunities for grasping. For a push action $a^\text{push} = (x_0, y_0, x_1, y_1)$, 
the gripper performs a quasi-static horizontal motion. Here, $(x_0, y_0)$ and 
$(x_1, y_1)$ are the start and end location of the gripper center, 
respectively. The gripper's orientation is fixed along the motion direction during 
a push maneuver.

\section{Methodology}

\subsection{Overview of the Proposed Approach}
When objects are tightly packed, the robot needs to carefully select an appropriate 
sequence of pushes that create a sufficient volume of empty space around the target 
object before attempting to grasp it. 
In this work, we are interested in challenging scenarios where multiple push 
actions may be necessary to de-clutter the surroundings of the target, and where the 
location, direction, and duration of each push action should be carefully optimized 
to minimize the total number of actions. 
Collisions among multiple objects often occur while pushing a single object, further
complicating the matter. 
To address the challenge, we propose a solution that uses a neural network 
to forecast the outcome of a sequence of push actions in the future, and estimates 
the probability of succeeding in grasping the target object in the resulting scene. 
The optimal push sequence is selected based on the forecasts.

A high-level description of the proposed solution pipeline is depicted in~\autoref{fig:vft-pipeline}. 
At the start of a planning iteration, an RGB-D image of the scene is taken, and the objects are detected and classified as {\it unknown clutter} or {\it target object}.
With the target object located, a second network called Grasp Network (\gn) predicts 
the probability of grasping the target. \gn is a Deep Q-Network (\dqn) \cite{mnih2015human}
adopted from prior works~\cite{zeng2018learning, huang2021dipn} for \orc. It 
takes the image input, and outputs the estimated grasp success probability for 
each grasp action. The target object is considered directly graspable if the maximum estimated grasp success probability is larger than a threshold. The robot executes the corresponding optimal grasp action; otherwise, push actions must be performed to create space for grasping. 

When push actions are needed, the next action is selected using Monte-Carlo 
Tree Search (\mcts). In our implementation, which we call the Visual Foresight 
Tree (\vft), each search state corresponds 
to an image observation of the workspace. Given a push action and a state, \vft 
uses the Deep Interaction Prediction Network (\dipn)~\cite{huang2021dipn} as the 
state transition function. Here, \dipn is a network that predicts the motions 
of multiple objects and generates a synthetic image corresponding to the scene 
after the imagined push. \vft uses \gn to obtain a reward value for each search node and detect 
whether the search terminates. Both \dipn and \gn are trained offline on different objects.

\subsection{Visual Foresight Trees}

This section discusses the three main components of \vft: \gn, \dipn, and Monte-Carlo Tree Search (\mcts).

\subsection{Grasp Network}

The Grasp Network (\gn), adapted from \cite{huang2021dipn}, takes the image $s_t$ as input, and outputs a pixel-wise reward prediction $R(s_t) = [R(s_t, a^1),\dots,R(s_t,a^n)]$ for grasps $a^1,\dots,a^n$. 
The output is a 2D map with the same size as the input image, and where each point contains the predicted reward of performing a grasp at the corresponding input pixel.
Table $R(s_t)$ is a single-channel image with the same size as input image $s_t$ ($224\times224$ in our experiments), and a value $R(s_t, a^i)$ represents the %\sout{predicted success probability} 
expected reward of the grasp at the corresponding action. 
To train GN, we set the reward to be $1$ for grasps where the robot successfully picks up only the target object, and $0$ otherwise. 
GN is the reward estimator for states in VFT (\autoref{subsec:vft}).

A grasp action $a^\text{grasp} = (x, y, \theta)$ specifies the grasp location and the end-effector angle. %To make the grasp reward  more consistent, 
\gn is trained while keeping the orientation 
of the end-effector fixed relative to the support surface, while randomly varying the poses of the objects. Therefore, \gn assumes that the grasps are aligned to the principal axis of the input image. 
To compute %\sout{$Q$-values} 
reward $R$ for grasps with $\theta \neq 0$, 
the input image is rotated by $\theta$ before passing it to \gn. 
As a result, for each input image, \gn generates $16$ different grasp $R$ reward tables. 

The training process of the \gn used in this work is based on previous works~\cite{zeng2018learning, huang2021dipn} but differs in terms of objectives, which requires a significant modification, explained in the following. 
%\sout{but different from that of previous works.}
The objective in previous works is to grasp all the objects; the goal of \orc 
is to retrieve a specific target among a large number of obstacles.
We noticed from our experiments that if \gn is trained to grasp all the objects, then a greedy policy will be learned, and it will always select the most accessible object to grasp. In contrast, all other objects that can also be directly grasped are ignored because they have low predicted rewards. This causes the problem that \gn cannot correctly predict the grasp success rate of a specific target object.
One straightforward adaptation to this new objective is only to give reward when the grasp center is inside the target object, which is the approach that was followed in~\cite{xu2021efficient}. 
However, we found that we can achieve a higher sample efficiency by providing a reward for successfully grasping any object.
The proposed training approach is similar in spirit to Hindsight Experience Replay (HER)~\cite{andrychowicz2017hindsight}. 
To balance between exploration and exploitation, grasp actions are randomly sampled from %from the probability density function 
$P(s, a^\text{grasp}) \propto b R(s, a^\text{grasp})^{b-1}$ where $b$ is set to $3/2$ in the experiments. 

After training, \gn can be used for selecting grasping actions in new scenes.
Since the network returns %\sout{Q-values} 
reward $R$ for all possible grasps, and not only for
the target object, the first post-processing step consists in selecting a 
small set of grasps that overlap with the target object. This is achieved by computing 
the overlap between the surface of the target object and the projected footprint 
of the robotic hand, and keeping only grasps that maximize the overlap. 
Then, grasps with the highest predicted values obtained from the trained 
network are ranked, and the best choice without incurring collisions 
is selected for execution. 

\subsection{Push Prediction Network}
\dipn~\cite{huang2021dipn} is a network that takes an RGB-D image, 2D masks of objects, center positions of objects, and a vector of the starting and endpoints of a push action. It outputs predicted translations and rotations for each passed object. 
%The center positions of objects can be predicted by using the transformations. 
The predicted poses of objects are then used to create a synthetic image.  
Effectively, \dipn imagines what happens to the clutter if the robot executes 
a certain push.

The de-cluttering tasks considered in~\cite{huang2021dipn} required only 
single-step predictions. The \orc challenge requires highly accurate predictions 
for multiple consecutive pushes in the future. To adapt \dipn for \orc, we fine-tuned
its architecture, replacing ResNet-18 with ResNet-10~\cite{he2016deep} 
while increasing the output feature dimension from $256$ to $512$ to predict 
motions of more objects simultaneously and efficiently. The number of decoder MLP layers is 
also increased to six, with sizes $[768, 256, 64, 16, 3, 3]$. Other augmentations 
are reported in section of experiments. Finally, we trained the 
network with $200,000$ random push actions applied on various objects.
This number is higher than the $1,500$ actions used in~\cite{huang2021dipn}
as we aim for the accuracy needed for long-horizon visual foresight. 
Given a sequence of candidate push actions, the fine-tuned DIPN predicts complex interactions, e.g.,~\autoref{fig:vft-predictions}.

\begin{figure}[ht!]
    \centering
    \includegraphics[width = \linewidth]{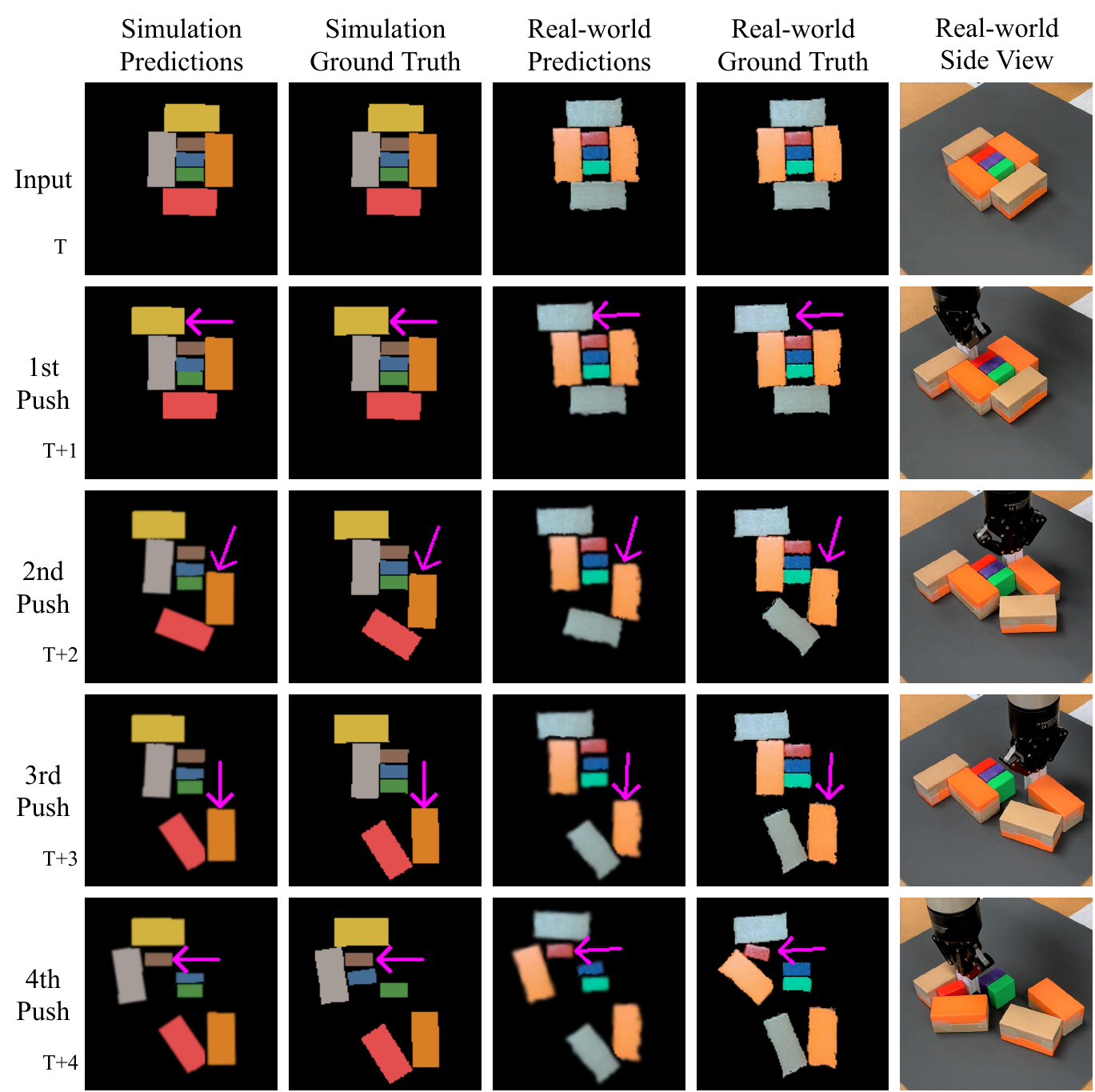}
    \caption{\label{fig:vft-predictions}
        Example of $4$ consecutive pushes showing that 
        \dipn can accurately predict push outcomes over a long horizon. 
        We use purple arrows to illustrate push actions. 
        The first and second columns are the predictions 
        and ground truth (objects' positions after executing the pushes) 
        in simulation. 
        The third and fourth columns show results on a real system. 
        The last column is the side view of the push result.
        Each row represents the push outcome with the previous row 
        as the input observation. 
    }
\end{figure}

\subsection{Visual Foresight Tree Search (\vft)}\label{subsec:vft}
We introduce \dipn for predicting single-step push outcome and \gn for 
generating/rating grasps as building blocks for a multi-step procedure capable 
of long-horizon planning. A natural choice is Monte-Carlo Tree Search 
(\mcts)~\cite{mcts2012}, which balances scalability and optimality.
In essence, \vft fuses \mcts and \dipn to generate an optimal multi-step 
push prediction, as graded by \gn. 
A search node in \vft corresponds to an input scene or one imagined by \dipn. 
\mcts prioritizes the most promising states when expanding the search tree; 
in \vft, such states are the ones leading to a successful target retrieval 
in the least number of pushes. 
In a basic search iteration, \mcts has four essential steps: 
selection, expansion, simulation, and back-propagation. 
First, the {\em selection} stage samples a search node and a push action based on a selection function. 
Then, the {\em expansion} stage creates a child node of the selected node. 
After that, the reward value of the new child node is determined by 
a {\em simulation} from the node to an end state. 
Finally, the {\em back-propagation} stage updates the estimated Q-values of 
the parent nodes. 

For describing \mcts with visual foresight, let $N(n)$ be the number of visits 
to a node $n$ and $Q(n) = \{r_1, \dots, r_{N(n)}\}$ as the estimated Q-values of each visit.
We use $N_{max}$ to denote the number of iterations the MCTS performed; 
we may also use an alternative computational budget to stop the search~\cite{mcts2012}.
The high-level workflow of our algorithm is depicted in~\autoref{alg:vft}, and illustrated in~\autoref{fig:vft-pipeline}.
We will describe one iteration (\autoref{alg:vft-iteration}-\autoref{alg:vft-bac-end}) 
of \mcts in \vft along with the pseudo-code in the remaining of this section.

\SetCommentSty{mycommfont}
\begin{algorithm}
    \begin{small}
    \DontPrintSemicolon
    \SetKwFunction{FMain}{VFT}
    \SetKwFunction{FMCTS}{MCTS}
    \Fn{\FMain{$s_t$}}{
    
        \While{\normalfont there is a target object in workspace}{
            % $R(o_t) \gets GN(o_t)$ \;
            % \If{$\max (R(o_t)) > R_g^*$}{$a \gets \argmax_{a^{grasp}}(R(o_t, a^{grasp}))$}
            $R(s_t) \gets \gn(s_t)$\;
            \If{\normalfont $\max_{a^{\text{grasp}}} R(s_t, a^{\text{grasp}}) > R_{g}^*$}{Execute $\argmax_{a^{\text{grasp}}}R(s_t, a^{\text{grasp}})$\tcp*[f]{Grasp}}
            \lElse{Execute \FMCTS{$s_t$}\tcp*[f]{Push}}
        }
    }
    % \;
    \vspace*{3pt}
    \SetKwProg{Pn}{Function}{:}{}
    \Pn{\FMCTS{$s_t$}}{
        Create root node $n_0$ with state $s_t$ \;
        $N(\cdot) \gets 0$, $Q(\cdot) \gets \varnothing$\tcp*{Default $N$, $Q$ for a search node}
        % $i \gets 0$ \;
        \For{$i \gets 1,  2, \dots,  N_{max}$}{
            $n_c \gets n_0$ \;\label{alg:vft-iteration}
            \Comment{\bf Selection and Expansion}
            \While{$n_c$ \normalfont is not expandable}{\label{alg:vft-selection}
                $n_c \gets \pi_{\text{tree}}(n_c)$\label{alg:vft-tree-policy}
                \tcp*{Use (\autoref{equation:vft-uct}) to find a child node}
            } 
            %\Comment{\bf Expansion}
            \vspace{1mm}
            $a^{\text{push}} \gets$ sample from untried push actions in $n_c$ \label{alg:vft-expansion1}\;
            $n_c \gets \dipn(n_c, a^{\text{push}})$\label{alg:vft-expansion2}\tcp*{Generate node by push prediction}
            % $s \gets $ state of $n_c$ \;
            % $R \gets 0$ \;
            % $d \gets 1$ \;
            \Comment{\bf Simulation}
            $r \gets 0$, $d \gets 1$, $s \gets n_c.\text{state}$\label{alg:vft-sim-start}\tcp*{$s$ is the state of $n_c$}
            % , $s \gets \text{state of } n_c$ \; 
            \While{$s$ \normalfont is not a terminal state}{
                % $a^{\text{push}} \gets \pi_{\text{rollout}}(n_c)$
                $a^{\text{push}} \gets$ randomly select a push action in $s$\label{alg:vft-simulation-roll} \;
                $s \gets \dipn(s, a^{\text{push}})$ \label{alg:vft-simulation-pred}\tcp*{Simulate to next state}
                $R(s) \gets \gn(s)$\;
                $r \gets \max\{r, \gamma^{d} \max_{a^{\text{grasp}}} R(s, a^{\text{grasp}})\}$\; \label{alg:vft-sim-gn}
                $d \gets d + 1$
            } \label{alg:vft-sim-end}
            \Comment{\bf Back-propagation}
            \While{$n_c$ \normalfont is not root\label{alg:vft-bac-start}}{
                $N(n_c) \gets N(n_c) + 1$ \;
                $R(n_c\text{.state}) \gets \gn(n_c\text{.state})$\;
                $r \gets \max\{r, \max_{a^{\text{grasp}}} R(n_c\text{.state}, a^{\text{grasp}})\}$ \; \label{alg:vft-bac-gn}
                $Q(n_c) \gets Q(n_c) \cup \{r\}$\tcp*{Record the reward}
                $r \gets r \cdot \gamma$\; 
                $n_c \gets \text{parent of } n_c$
            } \label{alg:vft-bac-end}
            % $i \gets i + 1$
        }
        % $n_{\text{best}} \gets \argmax\limits_{n_i \in \text{children of } n_0}(\uct(n_i, n_0))$ \;
        $n_{\text{best}} \gets \argmax_{n_i \in \text{children of } n_0}(\uct(n_i, n_0))$ \label{alg:vft-best-action} \;
        \Return push action $ a^{\text{push}}$ that leads to $n_{\text{best}}$ from the root
    }
    \caption{\label{alg:vft}
    Visual Foresight Tree Search}
    \end{small}
\end{algorithm}

\textbf{Selection.} 
The first step of \mcts is to select an {\em expandable} search node
(\autoref{alg:vft-selection}-\autoref{alg:vft-tree-policy}) using a tree policy 
$\pi_\text{tree}$. 
Here, {\em expandable} means the node has some push actions that 
are not tried via selection-expansion; more details of the push action space 
will be discussed later in the expansion part. 
To balance between exploration and exploitation, 
when the current node $n_c$ is already fully expanded, 
$\pi_\text{tree}$ uses Upper Confidence Bounds for Trees (\uct)~\cite{mcts2012} 
to rank its child node $n_i$. We customize \uct as
\begin{equation}\label{equation:vft-uct}
    \uct(n_i, n_c) = \frac{Q^{m}(n_i)}{\min\{N(n_i), m\}} + 
    C \sqrt{\frac{\ln{N(n_c)}}{N(n_i)}}.  
\end{equation}

Here, $C$ is an exploration weight. 
In the first term of (\autoref{equation:vft-uct}), unlike  typical \uct that favours 
the child node that maximizes $Q(n_i)$, 
we keep only the most promising rollouts of $n_i$ and denote by $Q^m(n_i)$ the average returns of the top $m$ rollouts of $n_i$. 
In our implementation, $m = 3$ and $C = 2$. 
We also use (\autoref{equation:vft-uct}) with parameters $m = 1$ and $C = 0$ 
to find the best node, and thus the best push action to execute, after the search is completed, as shown in~\autoref{alg:vft-best-action}.

\textbf{Expansion.} 
Given a selected node $n$, we use \dipn to generate a child node 
by randomly choosing an untried push action $a^\text{push}$
(\autoref{alg:vft-expansion1}-\autoref{alg:vft-expansion2}). 
The action $a^\text{push}$ is uniformly sampled at random from the 
selected node's action space, 
which contains two types of push actions:
\begin{enumerate}
    \item 
    For each object, we apply principal component analysis to 
    compute its feature axis. For example, for a rectangle object, the feature axis will be parallel to its long side. Four push actions are then sampled with directions perpendicular or parallel to the feature axis, pushing the object from the outside to its center.
    \item 
    To build a more complete action space, eight additional actions are evenly 
    distributed on each object's contour, with push direction also 
    towards the object's center.  
\end{enumerate}

\textbf{Simulation.}
After we generated a new node via expansion, 
in~\autoref{alg:vft-sim-start}-\autoref{alg:vft-sim-end}, 
we estimate the node's Q-value by 
% using a rollout policy $\pi_{\text{rollout}}$ to 
uniformly randomly select push actions at random (\autoref{alg:vft-simulation-roll}) 
and use \dipn to predict future states (\autoref{alg:vft-simulation-pred}) 
until one of the following two termination criteria is met:
\begin{enumerate}
    \item The total number of push actions used to reach a simulated state 
    is larger than a constant $D^*$.
    \item 
    The maximum predicted reward value of a simulated state 
    % as predicted by the Grasp Network,
    exceeds a threshold $R_{gp}^*$.
\end{enumerate}
In~\autoref{alg:vft-sim-gn}, when calculating $r$, a discount factor $\gamma$ 
is used to penalize a long sequence of action. 
Here, we use $\max\gn$ to reference the maximum value in a grasp reward table.  
In our implementation, \gn is only called once for each unique state 
and the output is saved by a hashmap.

\textbf{Back-propagation.} 
After simulation, the terminal grasp reward is back-propagated 
(\autoref{alg:vft-bac-start}-\autoref{alg:vft-bac-end})
through its parent nodes to update their $N(n)$ and $Q(n)$.
Denote by $r_0$ the max grasp reward of a newly expanded 
node $n_0$, and $n_1, n_2, \dots, n_k$ as the sequence of $n_0$'s parents in the ascending order up to node $n_k$. With $Q(n_0) = \{r_0\}$, the Q-value of $n_k$ in this iteration 
is then $\max_{0 \leq j < k} \gamma^{k - j}\max Q(n_j)$, 
which corresponds to the max reward of states 
along the path~\cite{song2020multi}.
Here, $\gamma$ is a discount factor to penalize a long sequence of actions. 
As a result, for each parent $n_k$, $N(n_k)$ increases by $1$, and 
$\max_{0 \leq j < k} \gamma^{k - j}\max Q(n_j)$ 
is added to $Q(n_k)$. 

\section{Experimental Evaluation}
We performed an extensive evaluation of the proposed method, \vft, in simulation 
and on the real hardware system illustrated in~\autoref{fig:vft-intro}. \vft is compared with multiple state-of-the-art
approaches~\cite{zeng2018learning, xu2021efficient} and DIPN in~\autoref{chap:dipn}, with 
necessary modifications for solving \orc, i.e., minimizing the number of actions
in retrieving a target. 
The results convincingly demonstrate \vft to be robust and more efficient 
than the compared approaches. 

Both training and inference are performed on a machine with an Nvidia GeForce 
RTX 2080 Ti graphics card, an Intel i7-9700K CPU, and 32GB of memory.

\begin{figure}[ht!]
    \centering
    \includegraphics[width = \linewidth]{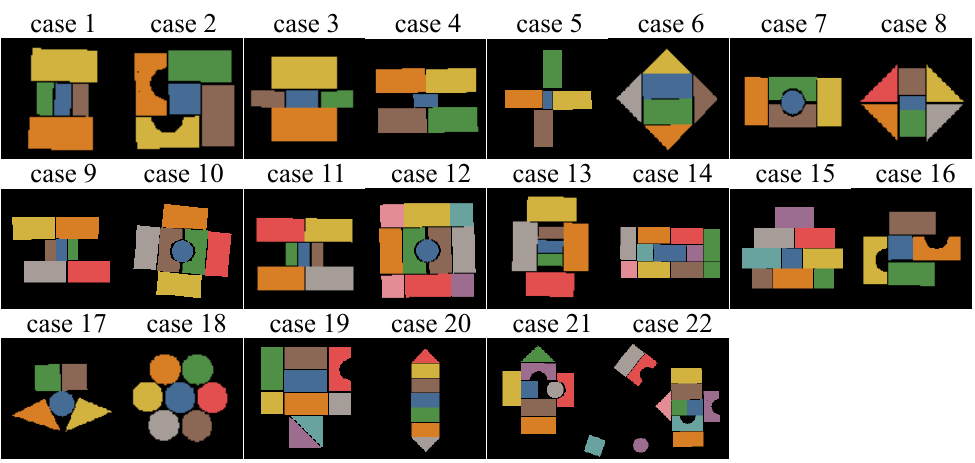}
    \caption{\label{fig:vft-testcases}
        22 Test cases used in both simulation and real world experiments. 
        The target objects are blue. 
        Images are zoomed in for better visualization.
    }
\end{figure}

\subsection{Experiment Setup}
The complete test case set includes 
\begin{enumerate}
    \item the full set of $14$ test cases from \cite{xu2021efficient} 
    \item $18$ hand-designed and more challenging test cases where the objects are tightly packed. 
\end{enumerate}
All test cases are constructed using wood blocks with different shapes, colors, 
and sizes. 
We set the workspace's dimensions to $44.8 \text{cm} \times 44.8 \text{cm}$. 
The size of the images is $224 \times 224$. Push 
actions have a minimum $5$cm {\em effective push distance}, 
defined as the end-effector's moving distance after object contact. 
Multiple planned push actions may be concatenated if they are in the 
same direction and each action's end location is the same as the next 
action's start location. 
In all scenes, the target object is roughly at the center of the scene.

The hyperparameters for \vft are set as follows. The number of iterations 
$N_{max}= 150$. The discount factor $\gamma=0.8$. The maximum depth $D^*$ of the tree is capped at $4$. The terminal threshold of grasp reward $R_{gp}^* = 1.0$. Threshold $R_g^*$ that decides to grasp or to push is $0.8$ in the simulation experiments and $0.7$ in the real hardware experiments. 
Such thresholds can potentially be fully optimized for a production system; it is not carried out in this work as reasonably good values are easily obtained while it is prohibitively time-consuming to carry out a full-scale optimization.

\subsection{Network Training Process}
\vft contains two deep neural networks: \gn and \dipn. 
Both are trained in simulation with the same objects as used in real experiments to capture the physical properties and dynamics of the environment. 
No prior knowledge is given to the networks except the dimensions of the gripper fingers.

\gn is trained on-policy with \num{20000} grasp actions. Similar to 
\cite{zeng2018learning, huang2021dipn, xu2021efficient}, randomly-shaped objects are uniformly dropped onto the workspace to construct the training scenarios. 
A successful grasp is decided by checking the distance between grippers, which should be greater than 0. 
A Huber loss on the pixel where the robot performed the grasp action is used. 
All other pixels do not contribute to the loss during back-propagation.
Image-based pre-training~\cite{huang2021dipn, yen2020learning} was employed to initialize the training parameters.
We then train the \gn by stochastic gradient descent with the momentum of $0.9$, weight decay of $10^{-4}$, and batch size of $12$.
The learning rate is set to $5 \times 10^{-5}$ and by half every $2000$ iteration.
%We also perform an extra round of \num{20000} iterations supervised learning to ensure convergence of the learned policy. % supervised learning is unclear here.

\dipn \cite{huang2021dipn} is trained in a supervised manner with \num{200000} 
random push actions from simulation. In the push data set, $20\%$ of the scenes contain randomly placed objects, and $80\%$ contain densely packed objects. %\sout{A Huber loss of $2$ is used.} %instead of $1$  (as in \cite{huang2020dipn}) is used. 
The push distance for DIPN is fixed to 7.4 cm (effective touch distance is 5 cm). 
In the DIPN (\autoref{chap:dipn}), the distance was 5 cm and 10 cm without considering the effective range.

We note that a total of $2000$ actions ($500$ grasps and $1500$ pushes) 
are sufficient for the networks to achieve fairly accurate results
(see, \autoref{chap:dipn}). Because training samples are readily 
available from simulation, it is not necessary to skimp on training data.
We thus opted to train with more data to evaluate the full potential of \vft. 

A Smooth L1 Loss with beta equals to $2$ is used instead of $1$ in~\autoref{chap:dipn}.
We train the \dipn by stochastic gradient descent with the momentum of $0.9$, weight decay of $10^{-4}$, and using cosine annealing schedule~\cite{loshchilov2016sgdr} with learning rates of learning rate of $10^{-3}$ for $76$ epochs, and the batch size is $128$.

\subsection{Compared Methods and Evaluation Metrics}

\textbf{Goal-Conditioned VPG (\gcvpg).} Goal-conditioned VPG (\gcvpg) is a modified 
version of Visual Pushing Grasping (VPG)~\cite{zeng2018learning}, which uses two 
DQNs \cite{mnih2015human} for pushing and grasping predictions. VPG by itself does not 
focus on specific objects; it was conditioned \cite{xu2021efficient} to focus on the 
target object to serve as a comparison point, yielding \gcvpg. 

\textbf{Goal-Oriented Push-Grasping.} In 
\cite{xu2021efficient}, many modifications are applied to VPG to render the resulting 
network more suitable for solving \orc, including adopting a three-stage training 
strategy and an efficient labeling method \cite{NIPS2017_453fadbd}. For convenience, 
we refer to this method as \gopg (the authors of \cite{xu2021efficient} did not provide a 
short name for the method).

\textbf{\dipn.}
As an ablation baseline for evaluating the utility of employing deep tree search, we replace 
\mcts from \vft with a search tree of depth one. In this baseline, \dipn is  used to evaluate 
all candidate push actions. The push action whose predicted next state has the 
highest grasp reward for the target object is then chosen. This is similar to 
how \dipn is used in \autoref{chap:dipn}; we thus refer to it simply as \dipn.

In our evaluation, the main metric is the total number of push and grasp 
actions used to retrieve the target object. 
For a complete comparison to \cite{zeng2018learning, xu2021efficient}, 
we also list \vft's grasp success rate, which is the ratio of successful grasps 
in the total number of grasps during testing. The completion rate, i.e., the chance of eventually grasping the target object, is also reported. Similar to
\autoref{chap:dipn}, when \dipn is used, a $100\%$ completion rate often reached.

We only collected evaluation data on \dipn and \vft. For the other two baselines, 
\gcvpg and \gopg, results are directly quoted from \cite{xu2021efficient} (at 
the time of our submission, we could not obtain the trained model or the information
necessary for the reproduction of \gcvpg and \gopg). While our hardware setup is identical to that of~\cite{xu2021efficient}, and the poses of objects are also identical,
we note that there are some small
differences between the evaluation setups:
\begin{enumerate}
    \item We use PyBullet~\cite{coumans2019} for simulation, 
    while \cite{xu2021efficient} uses CoppeliaSim; 
    the physics engine is the same (Bullet). 
    \item \cite{xu2021efficient} uses an RD2 gripper in simulation and 
    a Robotiq 2F-85 gripper for real experiment; 
    all of our experiments use 2F-85. 
    \item \cite{xu2021efficient} has a $13$cm push distance, 
    while we only use a $5$cm effective distance 
    (the distance where fingers touch the objects)
    \item \cite{xu2021efficient} uses extra top-sliding pushes which 
    expand the push action set. 
\end{enumerate}
We believe these relatively minor differences in experimental setup do not provide our algorithm an unfair advantage. 

\subsection{Simulation Studies}
\autoref{fig:vft-baseline-hist} and~\autoref{tab:vft-10table} show the evaluation results 
of all algorithms on the $10$ simulation test cases from \cite{xu2021efficient}. Each experiment is repeated $30$ times, and the average number of actions until task completion in each experiment is reported.
%
%These instances are relatively simple because they generally require only one push 
%action to free up the target object. 
Our proposed method, \vft, which uses an average of $2.00$ actions, significantly 
outperforms the compared methods. Specifically, \vft uses one push action and one 
grasp action to solve the majority of cases, except for one instance with a 
half-cylinder shaped object, which is not included during the training of the 
networks. Interestingly, when only one push is necessary, \vft, with its main 
advantage as multi-step prediction, still outperforms \dipn due to its extra 
simulation steps. 
The algorithms with push prediction perform better than \gcvpg and \gopg in all metrics.

\begin{figure}[ht!]
    \centering
    \includegraphics[width = .99\linewidth]{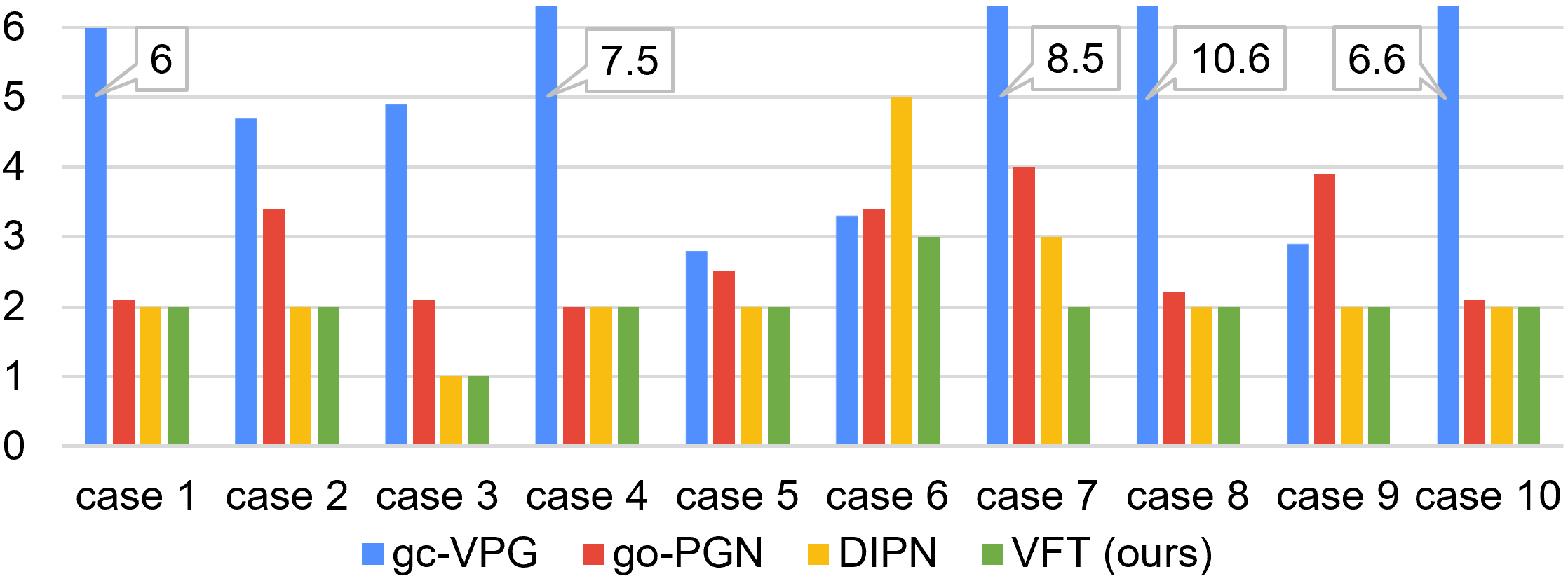}
    \caption{\label{fig:vft-baseline-hist}
        Simulation results per test case for the $10$ problems from 
        \cite{xu2021efficient}.
        The horizontal axis shows the average number of actions used to solve 
        a problem instance: the lower, the better. 
    } 
\end{figure}

\begin{table}[ht!]
    \centering
    \begin{tabular}{c|c|c|c}
        & Completion & Grasp Success & Number of Actions \\ \hline
        \gcvpg \cite{xu2021efficient} & $89.3\%$ & $41.7\%$ & $5.78$ \\ \hline
        \gopg \cite{xu2021efficient} & $99.0\%$ & $90.2\%$ & $2.77$ \\ \hline
        \dipn \cite{huang2021dipn} (\autoref{chap:dipn})  & $100\%$ & $100\%$ & $2.30$ \\ \hline
        \vft (\autoref{chap:vft}) & $100\%$& $100\%$ & $\mathbf{2.00}$ \\ \hline
    \end{tabular}
    \caption{Simulation results for the $10$ test cases from \cite{xu2021efficient}.}
    \label{tab:vft-10table}
\end{table}

To probe the limit of \vft's capability, we evaluated the methods on harder cases 
demanding multiple pushes. The test set includes $18$ manually designed instances 
and $4$ cases from \cite{xu2021efficient} (see~\autoref{fig:vft-testcases}). 
As shown in~\autoref{fig:vft-bar-sim-22} and~\autoref{tab:vft-22table-sim}, \vft uses 
fewer actions than \dipn as \vft looks further into the future. Though we could not 
evaluate the performance of \gcvpg and \gopg on these settings for direct comparison because we could not obtain the information necessary for the reproduction of these systems,
notably, the average number of actions ($2.45$) used by \vft on harder instances is 
even smaller than the number of actions ($2.77$) \gopg used on the $10$ simpler cases. 

\begin{figure}[ht!]
    \centering
    \includegraphics[width = .97\linewidth]{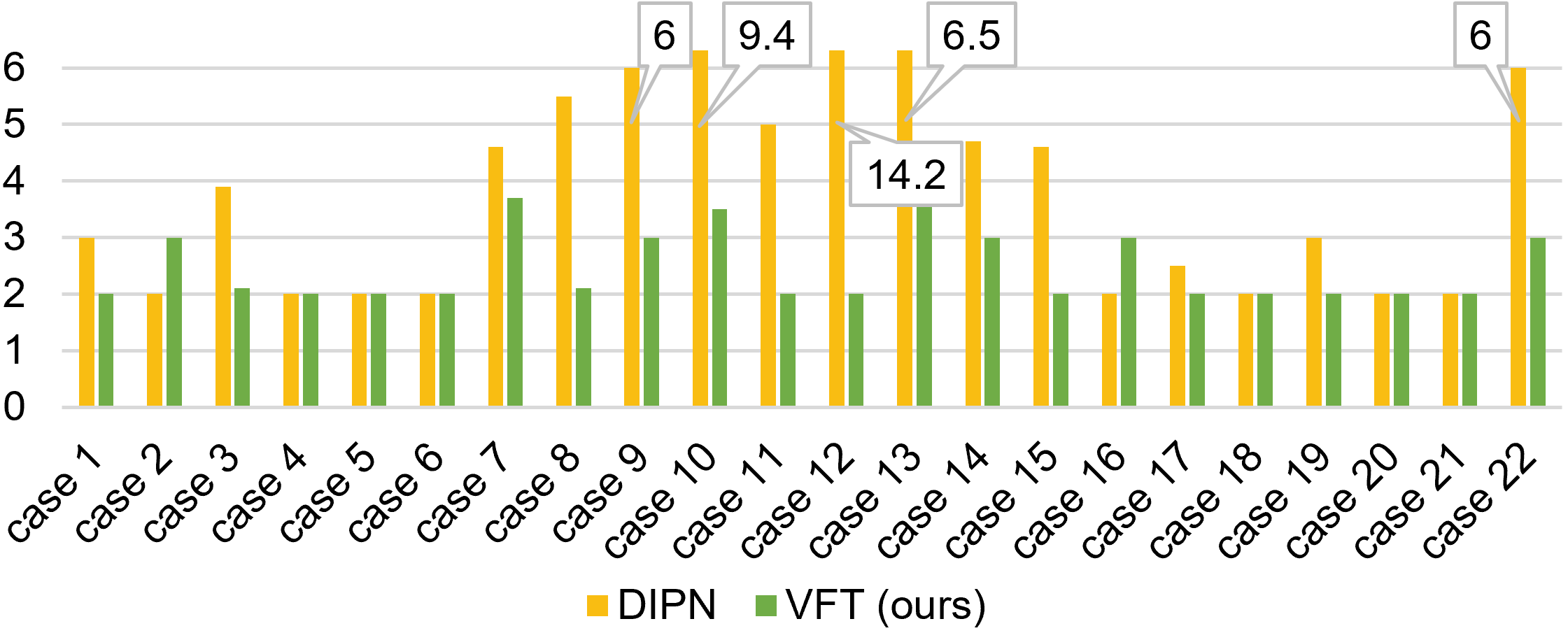}
    \caption{\label{fig:vft-bar-sim-22}
        Simulation result per test case for the $22$ harder problems 
        (\autoref{fig:vft-testcases}).
        The horizontal axis shows the average number of actions used to solve 
        a problem instance: the lower, the better.
    }
\end{figure}

\begin{table}[ht!]
    \centering
    \begin{tabular}{c|c|c|c}
        & Completion & Grasp Success & Num. of Actions \\ \hline
        \dipn \cite{huang2021dipn} (\autoref{chap:dipn}) & $100\%$ & $98.3\%$ & $4.31$ \\ \hline
        \vft (\autoref{chap:vft}) & $100\%$ & $98.8\%$ & $\mathbf{2.45}$ \\ \hline
        
    \end{tabular}
    \caption{Simulation result for the $22$ test cases in 
    ~\autoref{fig:vft-testcases}.}
    \label{tab:vft-22table-sim}
\end{table}

\subsection{Evaluation on a Real System}

We repeated the $22$ hard test cases on a real robot system (\autoref{fig:vft-intro-setup}). 
Both \vft and \dipn are evaluated. We also bring the experiment result from
\cite{xu2021efficient} on its $4$ real test cases for comparison. All cases are 
repeated at least 5 times to get the mean metrics. The result, shown in~\autoref{fig:vft-bar-real-22}, \autoref{tab:vft-22table-real}, and~\autoref{tab:vft-4table} 
closely matches the results from simulation. We observe a slightly lower grasp 
success rate due to the more noisy depth image on the real system. The real 
workspace's surface friction is also different from simulation. However, \vft and 
\dipn can still generate accurate foresight. 

\begin{figure}[ht!]
    \centering
    \includegraphics[width = .97\linewidth]{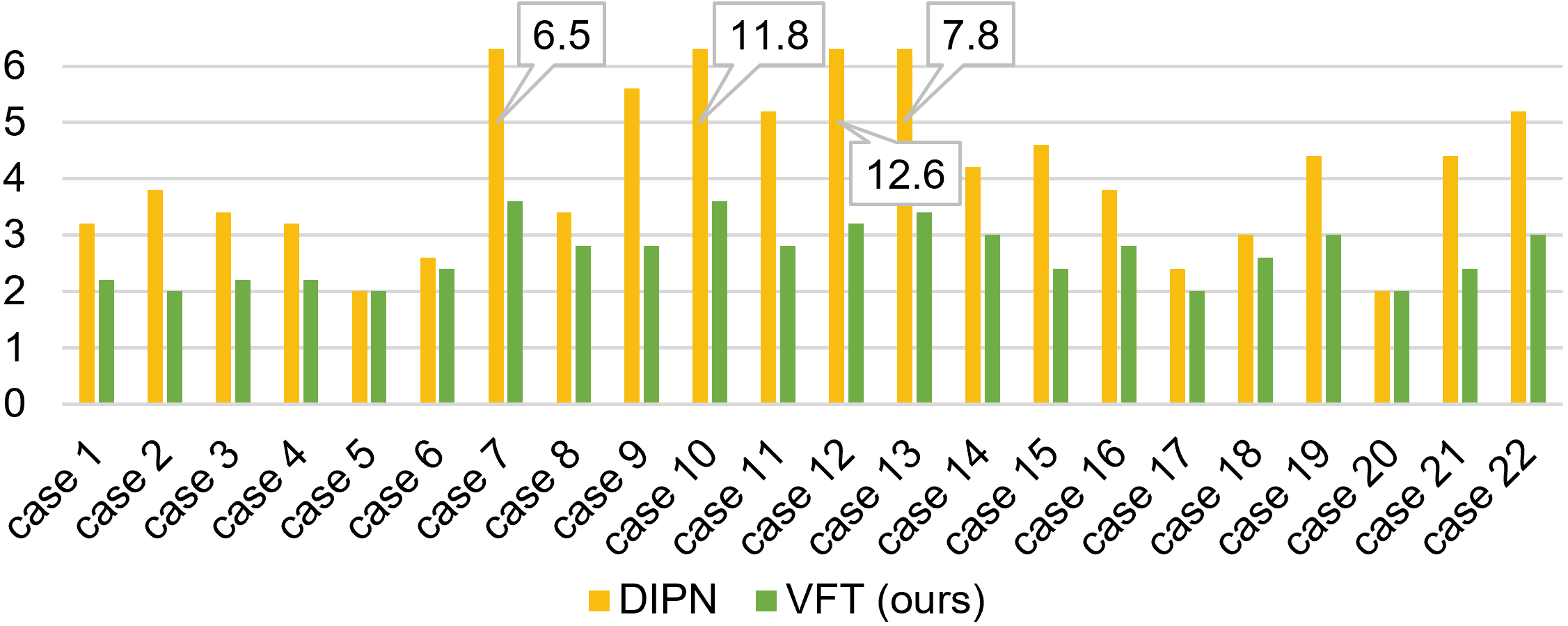}
    \caption{\label{fig:vft-bar-real-22}
        Real experiment results per test case for the $22$ harder problems 
        (\autoref{fig:vft-testcases}).
        The horizontal axis shows the average number of actions used to solve 
        a problem instance: the lower, the better.
    }
\end{figure}

\begin{table}[ht!]
    \centering
    \begin{tabular}{c|c|c|c}
        & Completion & Grasp Success & Num. of Actions \\ \hline
        \dipn \cite{huang2021dipn} (\autoref{chap:dipn}) & $100\%$ & $97.0\%$ & $4.78$ \\ \hline
        \vft (\autoref{chap:vft}) & $100\%$ & $98.5\%$ & $\mathbf{2.65}$ \\ \hline
    \end{tabular}
    \caption{Real experiment results for the $22$ Test cases in~\autoref{fig:vft-testcases}.}
    \label{tab:vft-22table-real}
\end{table}

\begin{table}[ht!]
    \centering
    \begin{tabular}{c|c|c|c}
        & Completion & Grasp Success & Num. of Actions \\ \hline
        \gopg \cite{xu2021efficient} & $95.0\%$ & $86.6\%$ & $4.62$ \\ \hline
        \dipn \cite{huang2021dipn} (\autoref{chap:dipn}) & $100\%$ & $100\%$ & $4.00$ \\ \hline
        \vft (\autoref{chap:vft}) & $100\%$ & $100\%$ & $\mathbf{2.60}$ \\ \hline
        
    \end{tabular}
    \caption{Real experiment results for cases $19$ to $22$ in~\autoref{fig:vft-testcases}.}
    \label{tab:vft-4table}
\end{table}

We also explored our system on everyday objects (\autoref{fig:vft-car-result}),  
where we want to retrieve a small robotic vehicle surrounded by soapboxes. 
%\sout{Without seeing neither the soap boxes nor the small vehicles,} 
Although the soapboxes and the small vehicles are unseen types of objects during training, the robot is 
able to strategically push the soapboxes away in two moves only and retrieve the vehicle. 

\begin{figure}[ht!]
    \centering
    \includegraphics[width=0.24\linewidth]{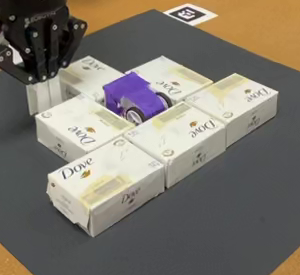}
    \hfill
    \includegraphics[width=0.24\linewidth]{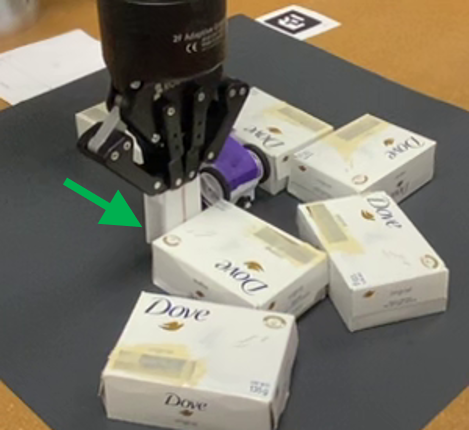}
    \hfill
    % \includegraphics[width=0.19\linewidth]{figures/vft/car-2.png}
    % \hfill
    \includegraphics[width=0.24\linewidth]{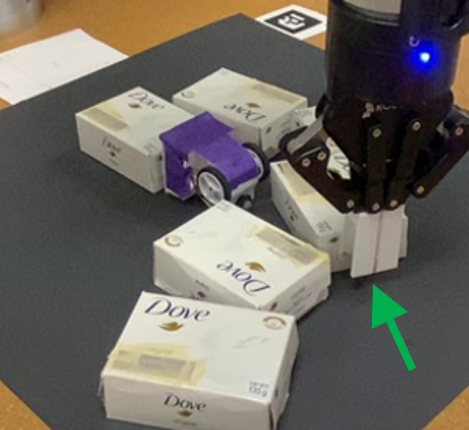}
    \hfill
    \includegraphics[width=0.24\linewidth]{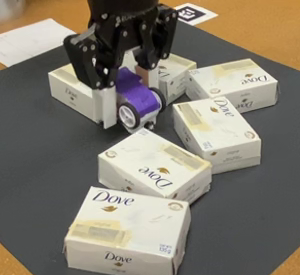}
    \caption{\label{fig:vft-car-result} 
        Test scenario with soap boxes and masked (in purple) 3D printed vehicle. 
        Two push actions and one grasp action.
    }
\end{figure}

We report that the running time to decide one push action is around 
$3$ minutes on average when the number of \mcts iterations is set to be $150$.
A single push prediction of \dipn took $30$ milliseconds.
While using the simulator as the transition function in \mcts under a similar criterion would take $8$ minutes on average to decide one push action.
In this chapter, our primary focus is action optimization. 

\section{Summary}
In conclusion, through an organic fusion of Deep Interaction Prediction Network (\dipn) and MCTS, the proposed Visual Foresight Trees (\vft) can make a high-quality multi-horizon prediction for optimized object retrieval from dense clutter. The effectiveness of \vft is convincingly demonstrated with extensive evaluation.
As to the limitations of \vft, the time required is relatively long because of the large \mcts tree that needs to be computed. This can be improved with multi-threading because the rollouts have sufficient independence. 
Currently, only a single thread is used to complete the \mcts.
It would also be interesting to develop a learned heuristic or value function to guide or truncate the rollout phase, potentially reducing inference time.
This technique would be similar in spirit to the MuZero algorithm~\cite{schrittwieser2020mastering}, which has been shown to be efficient by combining Monte Carlo tree search and learning by self-playing in an end-to-end manner. The learned rollout policy could lead to better performance, which will be covered in~\autoref{chap:more}. One issue related to end-to-end training is data efficiency, which is why this type of technique has been limited to games. Improving the data efficiency of end-to-end techniques is crucial to the deployment of these techniques on robotic tasks.
    
\chapter{Interleaving Monte Carlo Tree Search and Self-Supervised Learning for Object Retrieval in Clutter}\label{chap:more}
\thispagestyle{myheadings}

\def\ours{\textsc{MORE}\xspace}
\newcommand{\ppn}{\textsc{PPN}\xspace}

\section{Introduction}
%System 1 vs System 2
Kahneman \cite{kahneman2011thinking} proposed a thought-provoking hypothesis of human intelligence: in solving real-world problems, humans engage fast or ``System 1'' (S1) type of thinking for making split-second decisions, e.g., speech, driving, and so on. For other decision-making processes, e.g., playing chess, a slow or ``System 2'' (S2) approach is taken, where the brain would perform a search over some structured domain for the best actions to take. 
After repeatedly using S2 thinking to solve a given problem, patterns can be distilled over time and burned into S1 to accelerate the overall process. In playing chess, for example, good chess players can instinctively identify good candidate moves. First-time or beginner drivers rely heavily on S2 and gradually converge to S1 as they gain more experience. 
This S2$\to$S1 thinking has gained significant attention and has been explored in many directions in machine learning, including attempts at building machines with consciousness \cite{bengio2017consciousness}.
\begin{figure}[ht!]
\centering
\vspace{1mm}
\begin{overpic}[width=0.9\linewidth,tics=5]{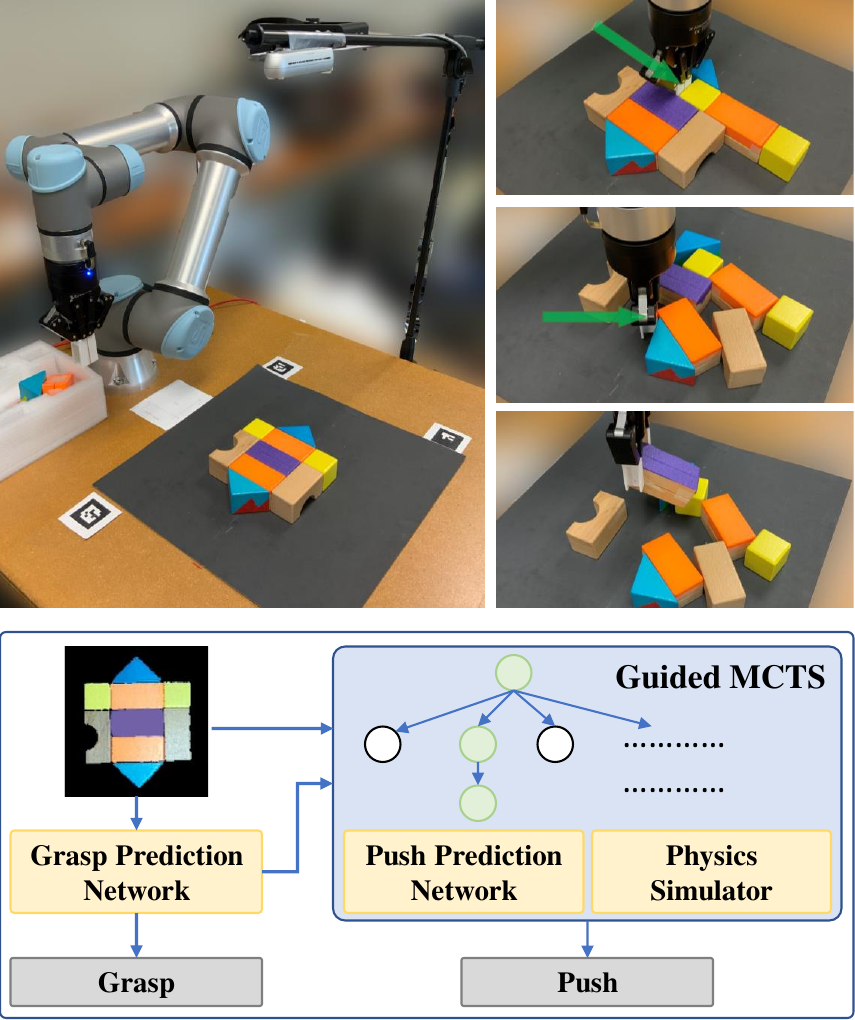}
\put(1,42){{\small (a) Hardware setup}}
\put(49.5,82){{\small (b) First push}}
\put(49.5,62){{\small (c) Second push}}
\put(49.5,42){{{\hypertarget{mylink}{\small (d) Grasp}}}}
\put(78.5,2.5){{\small (e)}}
\end{overpic}
%\vspace{0mm}
    \caption{\label{fig:more-intro}
    (a) The hardware setup for object-retrieval-from-clutter includes 
    a Universal Robots UR5e manipulator 
    with a Robotiq 2F-85 two-finger gripper, 
    and an Intel RealSense D455 RGB-D camera. 
    The objects are placed in a square workspace and the target object is masked in \textcolor{RoyalPurple}{purple}. 
    (b)(c) Two push actions (shown with green arrows) are used to enable the grasping of  the target (\textcolor{RoyalPurple}{purple}) object. 
    (d) The target object is successfully grasped and retrieved. 
    (e) The overview of our overall system.
    }
\end{figure}
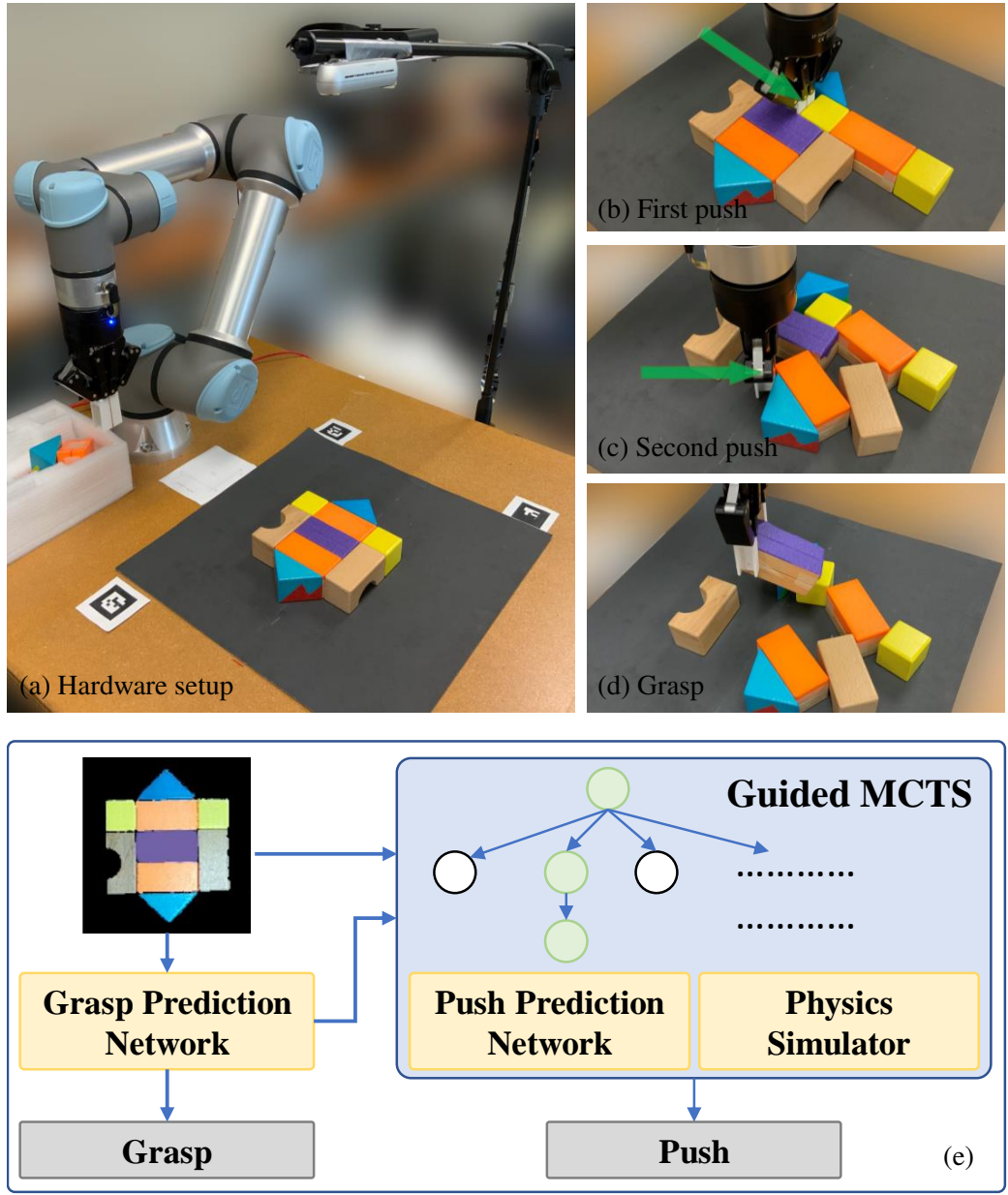
Perhaps the most prominent line of work in reinforcement learning \cite{sutton2018reinforcement} that closely aligns with this paradigm is the application of Monte Carlo Tree Search (MCTS) for carrying out self-supervised learning in games \cite{silver2018general,schrittwieser2020mastering}, where an ``understanding'' of a game emerges from a lifelong self-play and is gradually distilled so that it significantly reduces the search effort. Gradually, the overall system learns enough useful information that allows it to play perfect games with much less time and computing resources. 

Inspired by \cite{silver2018general,schrittwieser2020mastering} that show a search-and-learn approach for realizing S2$\to$S1 applies well to game-like settings with relatively well-defined rules, we set out 
to find out whether we could build a similar framework that enables 
real robots to interact with real-world physics and uncertainties to perform physical tasks, somewhat akin to \cite{DBLP:journals/corr/abs-1912-07024}. Specifically, we focus on the task of retrieving an object enclosed in clutter using non-prehensile actions, such as pushing and poking, followed by prehensile two-finger grasping. The goal is to obtain a computationally efficient system and produce high-quality solutions (i.e., using the minimum number of actions).  
%
%The contribution of our attempt, if successful, is that it would prove the feasibility of applying this generic approach in building real robotic systems capable of self-supervise in achieving optimal efficiency, thus opening the doors for a broad range of robotics applications. 

As pointed out by Valpola \cite{boney2019regularizing}, due to the difficulty in exploring the landscape of the state space of real-world problems, in addition to uncertainty, naive applications of the S2$\to$S1 paradigm often lead to undesirable behavior. 
Non-trivial design as well as engineering efforts are needed to build such S2$\to$S1 systems.
In the object-retrieval-from-clutter setting, the challenge lies in the difficulty of predicting the outcome of push actions, with the tip of the gripper, when many objects are involved.
This is due to discontinuities inherent in object interactions; for example, while a certain pushing action might move a given object, a slightly different push direction could miss that same object entirely. 

The main contribution of this work is proving the feasibility of applying the S2$\to$S1 philosophy to build a self-supervised robotic object retrieval system capable of continuously improving its computational efficiency, through \emph{cloning} the behavior of the time-consuming initial MCTS phase. 
Through the careful design and integration of two Deep Neural Networks (DNNs) with MCTS, our proposed self-supervised method, named \textbf{M}onte Carlo tree search and learning for \textbf{O}bject \textbf{RE}trieval (\ours), enables a DNN to learn from the manipulation strategies discovered by MCTS. Then, learned DNNs are fed back to the MCTS process to guide the search. 
\ours significantly reduces MCTS computation load and achieves identical or better outcomes, i.e., retrieving the object using very few strategic push actions. 
In other words, our method ``closes the loop''. This contrasts with \cite{DBLP:journals/corr/abs-1912-07024}, which only learns to replace the rollout function of MCTS. 

\section{Problem Formulation}

The Object Retrieval from Clutter task consists in using a robot manipulator to retrieve a hard-to-reach target object (\autoref{fig:more-intro}). 
Objects are rigid bodies with various shapes, sizes, and colors; the target object is assigned a unique color. % (\textcolor{RoyalBlue}{blue} in simulation, \textcolor{RoyalPurple}{purple} in robot experiments). 
%The blue and purple are uniquely assigned to the target object.
Similar to \autoref{chap:dipn}, a top-down fixed camera is installed to observe the workspace. %, where objects are rigid and non-occluded (no overlaps), with the scene largely a planar one. 
The camera takes an RGB-D image of the workspace (e.g., the top-left image of \autoref{fig:more-intro}), which serves as the only input to our system. % (the model of an object is known).
%
%where each cell is a square with side length $0.002$m.

\emph{Pushing} and \emph{grasping} actions are allowed, the execution of each is considered as one \emph{atomic action}.
A grasp action is defined as a top-down overhead grasp motion $a^g=(x, y, \theta)$, corresponding to the gripper's target location and orientation, based on a coordinate system defined over the input image.
A push action is defined as a quasi-static planar motion $a^p=(x_0, y_0, x_1, y_1)$ where $(x_0, y_0)$ and $(x_1, y_1)$ are the start and the end locations of the gripper tip.
The horizontal push distance is fixed and it is $10$cm in our experiments.
Each primitive action is transformed to the real-world coordinates for execution, but all the planning and reasoning are in image coordinates.
The robotic arm keeps pushing objects until the target object can be grasped or until the target object is pushed outside of the workspace, in which case the task is considered a failure. The problem is to find a policy that maximizes the frequency of successfully grasping the target object, while also minimizing the number of pre-grasp pushing actions.

\section{Methodology} 
The \ours framework consists of three components: a Grasp Network (\gn), a Monte Carlo Tree Search (MCTS) routine, and a Push Prediction Network (\ppn). 
\gn is a neural network that predicts the success probabilities of grasp actions. It is trained online similarly to \autoref{chap:vft}. The success probabilities can be interpreted as immediate rewards. MCTS uses a physics engine as a transition function to simulate long sequences of consecutive push actions that end with a terminal grasp action. Each branch in MCTS is composed of push actions as internal nodes, and a grasp action as a leaf. Grasp actions are evaluated with \gn, and the returned rewards are back-propagated to evaluate their corresponding branches. The branch with the highest discounted reward, or {\it Q-value}, is selected for execution by the robot. 

While highly effective in finding near-optimal paths, MCTS suffers from a high computation time that makes it impractical. 
To solve this, \ours employs a second neural network, \ppn, to prioritize the action selection in the rollout policy.
The robot starts by relying entirely on MCTS (S2 type of thinking) to solve various instances of the object-retrieval problem. Instead of throwing away the computation performed by MCTS for solving the various instances, we use the computed Q-values as ground-truth to train \ppn. Note that this computation data is free, since it is generated by the simulations performed by MCTS as a byproduct of solving the actual problem. \ppn is a neural network that learns to imitate MCTS and clone its behavior, while avoiding heavy computation and physics simulations by MCTS. As \ppn becomes more accurate in predicting the outcome of MCTS, the robot starts relying on both MCTS and \ppn for action selection. In a nutshell, \ppn is used for orienting the search in MCTS toward more promising push actions that rearranges the scene and renders the target object graspable. After a long experience, \ppn's accuracy in predicting the Q-values of push actions matches that of MCTS, and the robot switches entirely to \ppn to make decisions in a few milliseconds (S1 type of thinking).

\subsection{Monte-Carlo Tree Search}\label{sec:more-mcts} 
Monte-Carlo Tree Search (\mcts) \cite{coulom2006efficient} is used in \ours for both decision-making and training \ppn.
A typical \mcts routine has four steps: selection, expansion, simulation, and back-propagation.
In our case, the goal of the search is to find the shortest action sequence; we can stop the search as soon as the best solution is found without exploring the rest.
The search stops in two cases:
\begin{enumerate}
    \item the number of iterations $n$ exceeds a pre-set budget $N_{max}$, or
    \item the expanded node with state that the target object can be grasped, and all nodes in parent level are expanded.
\end{enumerate}
A node is considered as a leaf if $\max_{i,j,\theta}R_{gm}(s_t)[i,j, \theta] > R_{g*}$ where $R_{gm}(s_t)$ is obtained from \gn and $R_{g*}$ is a pre-defined high probability. 
%
%from Grasp Network is used to determine a leaf node (terminal).
The maximum depth of the tree is limited to $d$, where $d$ is set to $4$ in our experiments. 

In the selection phase, we find an expandable node starting from the root according to the search policy
\begin{equation}\label{eq:more-pi}
    \pi_n (s) = \argmax\limits_{a^p}(Q(s, a^p) + C \sqrt{\frac{\ln{N(s)}}{N(s, a^p)}}),
\end{equation}
where $N(s)$ is the number of visits to node (state) $s$ and $N(s, a^p)$ is the number of times push action $a^p$ has been selected at node (state) $s$.
The Q-value is calculated as
\begin{equation}
    Q(s, a^p) = \frac{\sum_{i=1}^{m}{r}_i(s, a^p)}{{\min\{N(n_i), m\}}},
\end{equation}
where $r(s, a^p)$ is the returned long-term reward and $m$ is a pre-set maximum.
Only the best $m$ terms ${r}_i(s, a^p)$ are used to compute the Q-value in the equation above. 
$m$ is set to $10$ when expanding nodes and $1$ when selecting the best solution.
$C$ is the coefficient of the exploration term,  and it is set to $2$ when expanding nodes and $0$ when selecting the best solution.
In the expansion phase, we use a physics simulator to execute the chosen push action $a^p$ at state $s_i$ and predict new state $s_{i+1}$. 
Then, a random policy is used to sample actions to simulate until a grasp is possible or a failure is encountered.
%.
The reward $r$ is predicted by \gn at a terminal state $s_t$. Reward 
$r$ is set to $1$ if $\max_{i,j,\theta}R_{gm}(s_t)[i,j, \theta] > R_{g*}$, and $0$ otherwise.
One additional term $\delta\max(R_{gm}(s_t))$ is added to $r$, to distinguish between good and bad push actions.
We set $\delta$ to be $0.2$.
In the last step, reward $r$ is propagated back to its parent nodes to update their $Q$-values with a discount factor $\gamma=0.5$. 

\begin{figure}[ht!]
    \centering
  \includegraphics[width=0.5\linewidth]{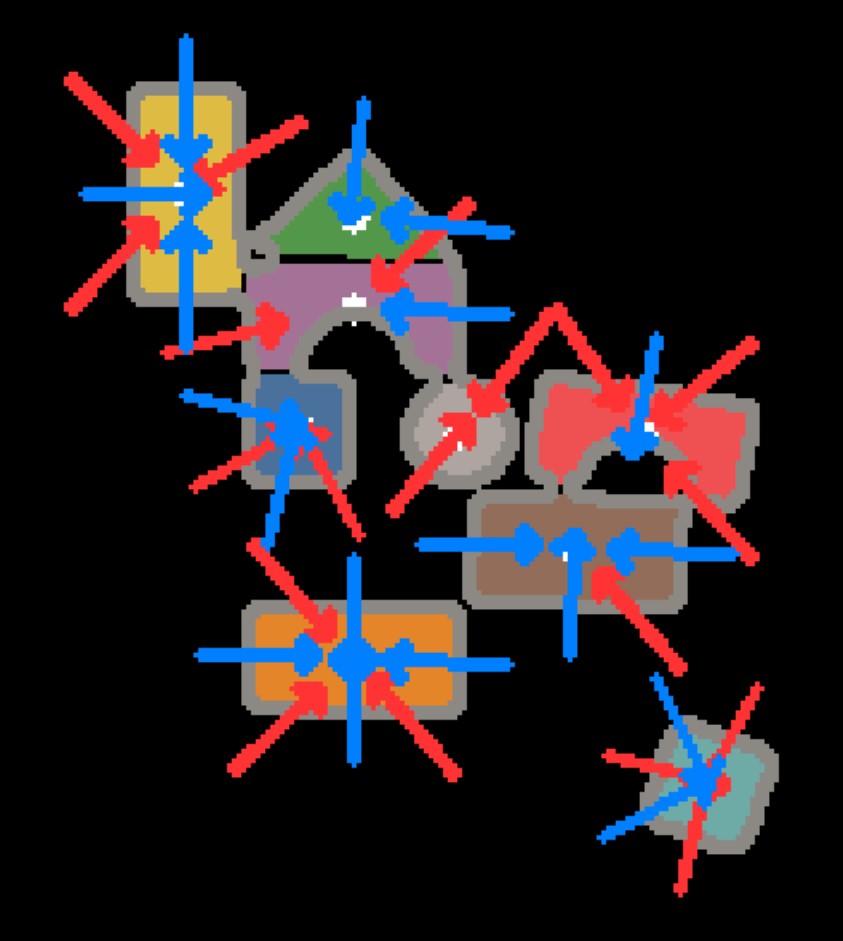}
  \caption{\label{fig:more-sample}
    Sampled push actions.}
\end{figure}

As the push action space is enormous even after discretization, we further sample a subset of actions such that all push actions start around the contour of an object and point to the center of the object (\autoref{fig:more-sample}).
This action sampling method has been discussed
in VFT (\autoref{chap:vft}) and was empirically proven efficient for a similar setup of object retrieval.

In our implementation, $N_{max}$ is set to $300$ when \mcts is used to collect data to train \ppn. 
The second and the third conditions for stopping the search are only activated after at least $50$ roll-outs, so that the number of visits to a state is statistically significant and to reduce the variance of \ppn.

\subsection{Push Prediction Network (\ppn)}\label{sec:more-pushnet}
As previously mentioned, \ppn learns to imitate MCTS.
\ppn is a deep neural network with ResNet-34 FPN~\cite{he2016deep, DBLP:journals/corr/LinDGHHB16} as the backbone, where the P2 level of the FPN connects to the head. 
It takes a two-channel input and outputs a single channel pixel-wise push Q-value map, similar to the reward map produced by \gn.
An example input is shown in~\autoref{fig:more-push-in-out}, where the first channel is a segmented image of all objects and the second channel is a binary image of the target object.
The output is the image on the right of~\autoref{fig:more-push-in-out}, where the arrow shows a push action with the highest Q-value.
\ppn estimates the Q-value (discounted rewards) $Q_p(s_t)$ of executing push actions at the corresponding pixel, where the action is assumed to push $10$cm to the right.
$\max(Q_p(s))$ is limited to the range $[0, \eta]$, where $\eta$ is the maximum reward of a terminal state.

\begin{figure}[ht!]
    \centering
    \includegraphics[width = .97\linewidth]{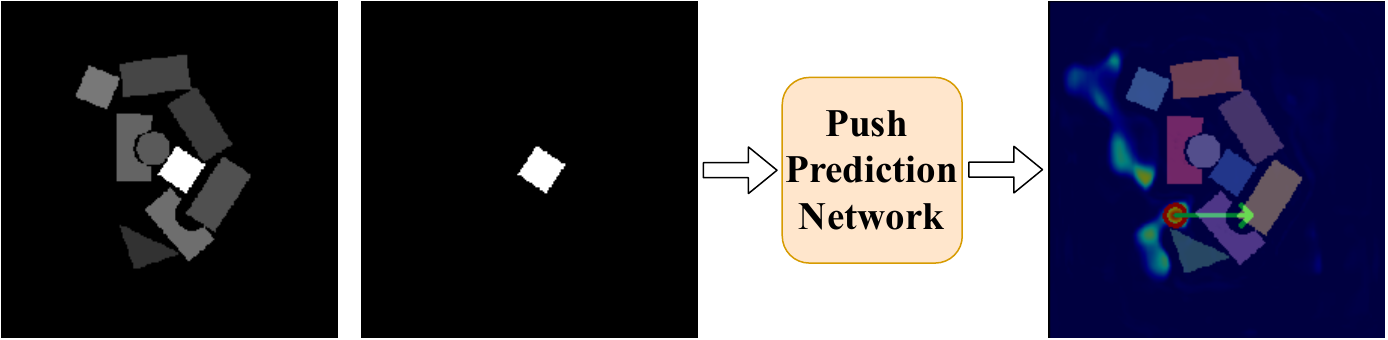}
    \caption{\label{fig:more-push-in-out}
        The left two figures are the input to \ppn. The first is a segmentation of objects; the second is the mask of the target object. The image on the right is the output from the \ppn. We use Jet colormap to represent the reward value, where the value ranges from red (high) to blue (low). The pixel with the highest Q-value is plotted with a circle and attached with an arrow on the right image, representing pushing action starting at the circle and moving to the right with a distance of $10$cm.
    } 
\end{figure}

When \mcts is used to generate training cases, it builds a tree and saves the transitions for each case: the state (image) $s$, the push action $a^p$, the Q-value $Q(s, a^p)$, and the visited number $N(s, a^p)$.
As such, \ppn is trained in a self-supervised manner.
The input image is rotated based on a push action so that the corresponding push action points to the right.
Because a single action is generated by MCTS (i.e., a $\delta$ signal over the entire input), which is not conducive to training \ppn, we ``expand'' the Q-value over a $3 \times 3$ patch centered around MCTS action but set invalid pushes (e.g., if part of the patch is inside an object) to be zero. 
Now, the label is relatively dense compared to a one-hot pixel, so we can use Smooth L1 loss from Pytorch~\cite{NEURIPS2019_9015} with $\beta$ equals to $0.8$ to regress.
Only gradients on the labeled pixels are used.
Loss weighting is also applied: label values from the \mcts are weighted based on $N(s, a^p)$, label values (zero Q-value) from void push actions are weighted with a small number, $0.001$ for collision and $0.0001$ for pushing thin air.
We observe that \ppn has difficulty learning to create clearance around the target object.
Data augmentation is applied here so that for each training case, we also randomly choose the target object for the \mcts to solve; 
so each arrangement becomes many training cases.
It mitigates over-fitting; given similar visual information, it could learn different strategies, as the target object could be anywhere.

The head model is an FCN with four layers, where the first two layers have a kernel size of 3, the last two 1, and the strides of four layers are all 1.
Batch normalization is used at each layer of the head model except the last.
Bilinear interpolation ($\times 2$) is applied interleaved between the last three layers of the head model to scale up the hidden state to the same size as the input image.
The training process has two stages, one to train the network with a batch size of $8$, learning rate starts at $1\mathrm{e}{-4}$, for $50$ epochs.
The learning rate decays with cosine annealing~\cite{loshchilov2016sgdr}, where the maximum number of iterations is set to be the same as the epoch number $50$ and the minimum learning rate is $1\mathrm{e}{-8}$.
The second is a fine-tuning stage; we increase the batch size to $28$ and the learning rate to $1\mathrm{e}{-5}$ with an epoch of $20$.

\subsection{Guided Monte-Carlo Tree Search}
With the trained \gn and \ppn, a guided MCTS is implemented to accelerate the search process, cutting cost from time-consuming expansion and simulation phases.
\gn is again used to determine the terminal state and if so, calculate its estimated reward, as discussed in~\autoref{sec:more-mcts}.
\ppn, trained with data from \mcts, can estimate how much reward can be gained from taking a push action at a certain state.

For this combination of \mcts with \ppn, some additional updates are made (compared to~\autoref{sec:more-mcts}) to incorporate the guidance from \ppn.
The exploration term is removed from the search policy, so $C$ in equation~\autoref{eq:more-pi} is set to 0.
Similar to~\cite{hamrick2019combining}, we use the estimated reward from \ppn as a prior, so the Q-value is calculated as follows
\begin{equation}
    Q_{guide}(s, a^p) = \frac{\max(Q_p(s)) + \sum_{i=1}^{m}{r}_i(s, a^p)}{N(s, a^p)},
\end{equation}
where $m$ is set to $3$ when expanding nodes and $N(s, a^p)$ is initialized to $1$ for all state-action pairs.
Instead of computing an average as standard MCTS, only best $m$ of $Q_p$ are considered, this is due to the number of rollout is small, a good action could be averaged out.
To select the best action as the next step solution, the Q-value is calculated without the denominator
\begin{equation}
    Q_{best}(s, a^p) = \max(Q_p(s)) + \max({r}_i(s, a^p)),
\end{equation}
where only the best explored solution is considered.% with the prior knowledge.

The push action space of the guided MCTS is limited to a subset (like~\autoref{fig:more-sample}) so that the estimated reward from \ppn is more accurate and the branching factor of the tree is of a reasonable size.
To make the selection mimic the training data, we rotate the image for each sampled push action such that the push action in the rotated image is always pointing to the right.
Then, we only use the estimated Q-value at the corresponding pixel (push action) of the output Q-value map.
An example of guided \mcts is given in~\autoref{fig:more-guided-mcts}.
The expansion of the tree is prioritized by \ppn, where the push action with higher Q-value is sampled earlier, and the rollout policy is also prioritized. %The proposed method can find a solution in two steps.
The maximum depth of the tree is limited to $3$ instead of $4$ as used in the earlier version of MCTS for collecting data to train \ppn.

\begin{figure}[ht!]
    \centering
    \includegraphics[width = .97\linewidth]{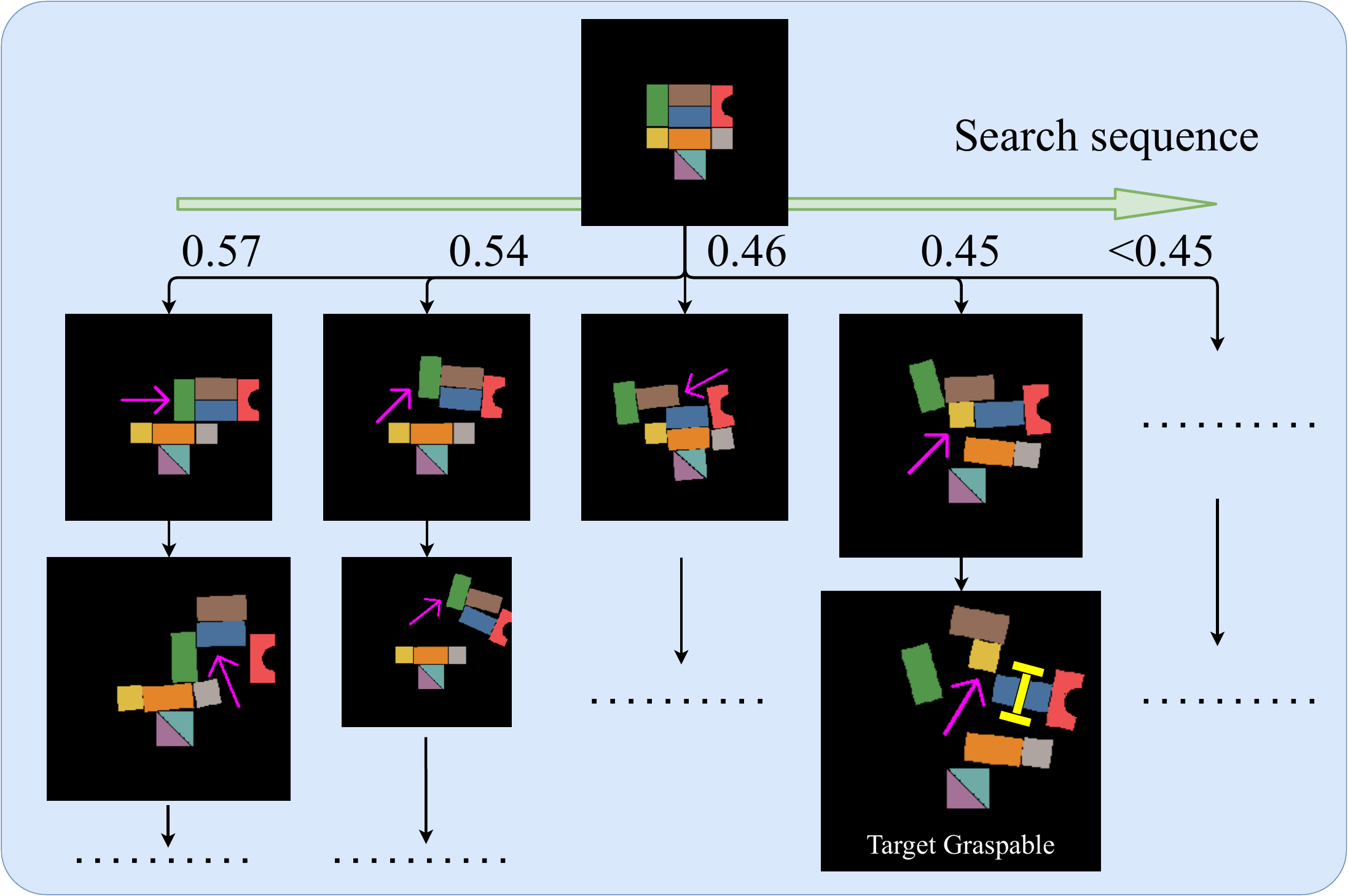}
    \caption{\label{fig:more-guided-mcts}
        An example of the guided MCTS with a budget of 10 iterations. State with larger image have higher estimated Q-values. All expanded nodes are plotted. The numbers in the first levels represent the estimated Q-value returned by \ppn for corresponding push action. These values, together with the reward returned from simulation, guide the tree search.
    } 
\end{figure}

\section{Experimental Evaluation} 

We evaluated the proposed technique both in simulation (PyBullet~\cite{coumans2021}) and on adversarial test cases on a UR5e robot with a Robotiq 2F-85 gripper using real objects.
The robot, workspace, objects, and camera are the same in simulation and real-world experiments, so that we can seamlessly transfer from simulation to the real setup.
The workspace is limited to a square with a side length of $0.448$m; it is discretized as a grid of $224 \times 224$ cells during the image processing step.
The friction of objects and table surface cannot be accurately measured; nevertheless, high-fidelity physical properties do not seem to be needed for this particular application.
The results demonstrate that the proposed method significantly outperforms MCTS in~\autoref{chap:vft} \cite{huang2021visual} in terms of time efficiency while returning plans of equal quality. The plans returned by the proposed technique contain fewer actions and yield higher success rates than those returned by the purely learning-based solution presented in~\cite{xu2021efficient}. 
%time-efficient without loss of action efficiency, compared to a \mcts solution~\cite{huang2021visual}, and action efficient compared to a learning-based solution~\cite{xu2021efficient}.
%
Training and evaluation are completed on a machine with an Intel i7-9700K CPU and an Nvidia GeForce RTX 2080 Ti.

\subsection{Simulation experiments}
\textbf{Tasks.} 
Given an arrangement of heterogeneous and tightly packed objects, a target object is to be retrieved using push and grasp actions from a two-finger gripper.
In simulation, we benchmark on 22 adversarial test cases from~\autoref{chap:vft} (\autoref{fig:vft-testcases}) and 10 from~\cite{zeng2018learning, xu2021efficient}.
Here ``adversarial'' means that at least one push action has to be executed for a grasp action to be feasible (insert gripper without collision).
Random cases, which are too easy from \autoref{chap:dipn} and \autoref{chap:vft} \cite{huang2021dipn, huang2021visual}, are not discussed here.

\textbf{Metrics.} 
We use four metrics:
\begin{enumerate}
    \item the number of actions used to retrieve the target object,
    %As \gn is well-trained and the object is not hard to grasp.
    \item the total time used for retrieving the target object, which includes both planning time and execution time for simulation results,
    \item the completion rate, failures occur when the target object is pushed out of the workspace, and %, as it does not have a hard constraint.
    \item the grasp success rate, which is the number of successful grasps divided by the total number of grasping attempts.
\end{enumerate}
The number of re-arrangement actions that are needed to make the target object graspable and time are the two main metrics.
The completion and grasp success rates are also reported but are not the main focus as they are often close to $100\%$.

\textbf{Baseline Methods.} We compare with three methods:
\begin{enumerate}
    \item A self-supervised reinforcement learning method denoted as \gopg~\cite{xu2021efficient}, which trains a grasp \dqn and a push \dqn then selects an action with the highest Q-value out of the two networks to execute.
    \item \mcts as described in Section~\ref{sec:more-mcts}. This is adapted from~\cite{huang2021visual}, but we use here a simulator to predict the next state instead of the originally used learned model, for fair comparisons. %Aswe use simulator to simulate states in our network-driven \mcts.
    \item \ppn as described in Section~\ref{sec:more-pushnet}. \ppn proposes push actions based on their predicted Q-values and the robot executes those actions until the target object can be grasped according to \gn.
\end{enumerate}

\textbf{Simulation Studies.} We ran our method and the three alternative methods on 22 cases~\autoref{chap:vft} and 10 cases~\cite{zeng2018learning, xu2021efficient}, in simulation first.
\autoref{tab:more-sim22table} and~\autoref{tab:more-sim10table} show the overall performance of the four methods, where \mcts based methods are limited to a budget of $50$ iterations per test case.
%
%\TODO{
In this paper, we denote the tree search methods with different budgets of search iterations as \mcts-10/20/50 and \ours-10/20/50, where the suffix denotes the iterations limit.
%}
%
The 22 cases are generally harder to solve than the 10 cases, where the target object can be retrieved after one push action.
The time metric records the average time (out of 5 trials) for retrieving the object, including planning and execution times.
%
%\TODO{Segmentation is obtained from the simulator for simulation experiments, which is different from the real robot experiments.}
%

For the baseline \gopg, results on 10 cases are directly quoted from the paper (at the time of our submission, we could not obtain the trained model or the information necessary for fully reproducing \gopg).
\ours uses the fewest number of actions to solve the task.
Performance details on 22 cases can be found in~\autoref{fig:more-22-test-num} for the number of actions and \autoref{fig:more-22-test-time} for the running time.
\ppn is fast as it is a one-stage DNNs solution. It learned a policy that creates free spaces around the target object, but it is less consistent and less stable than the tree search solutions. From our observation, \ppn can propose non-prevailing pushing actions. 
%Take an example case 14 in Fig.\ref{fig:22-test-num}. 
%This could be due to the lack of training data close to Case 14.
%
\mcts provides a consistent and good quality solution, but requires a much longer planning time.
\ours, combining the benefits of both, reduces the planning time and delivers high-quality solutions.

\begin{table}[ht!]
    \centering
    \begin{tabular}{c|c|c|c|c}
        & Num. of Actions & Time & Completion & Grasp Success   \\ \hline
        \ours-50 & $\mathbf{2.61}$ & $82$s & $100\%$ & $99.2\%$ \\ \hline
        \mcts-50~\cite{huang2021visual} & $2.69$ & $208$s & $100\%$ & $99.1\%$ \\ \hline
        \ppn & $3.68$ & $8$s & $100\%$ & $97.7\%$ \\ \hline
    \end{tabular}
    \vspace{2mm}    
    \caption{Simulate experiment results for 22 cases~\autoref{chap:vft}. Budgets of \mcts and \ours are limited up to 50 iterations.}
    \label{tab:more-sim22table}
\end{table}

\begin{table}[ht!]
    \centering
    \begin{tabular}{c|c|c|c|c}
        & Num. of Actions & Time & Completion & Grasp Success   \\ \hline
        \ours-50 & $\mathbf{2.10}$ & $16$s & $100\%$ & $100\%$ \\ \hline
        \mcts-50~\cite{huang2021visual} & $2.20$ & $32$s & $100\%$ & $93.4\%$ \\ \hline
        \ppn & $2.70$ & $4$s & $100\%$ & $95.0\%$ \\ \hline
        \gopg~\cite{xu2021efficient} & $2.77$ & $-$ & $99.0\%$ & $90.0\%$ \\ \hline
    \end{tabular}
    \vspace{2mm}    
\caption{Simulate experiment results for 10 cases~\cite{xu2021efficient}. Budgets of \mcts and \ours are limited up to 50 iterations.}
    \label{tab:more-sim10table}
\end{table}

\begin{figure}[ht!]
\vspace{1mm}
    \centering
    \includegraphics[width = .97\linewidth]{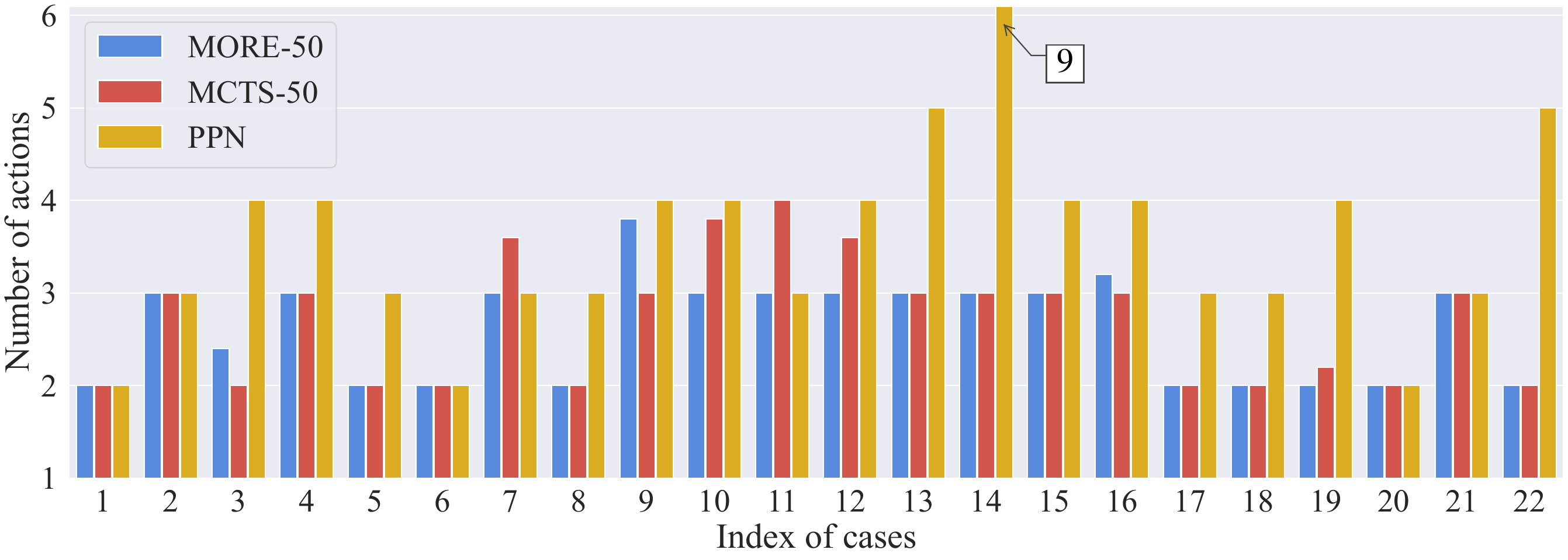}
    \caption{\label{fig:more-22-test-num}
        The average number (out of 5 trials) of action used to solve one case for 22 cases.
    } 
\end{figure}

\begin{figure}[ht!]
\vspace{1mm}
    \centering
    \includegraphics[width = .97\linewidth]{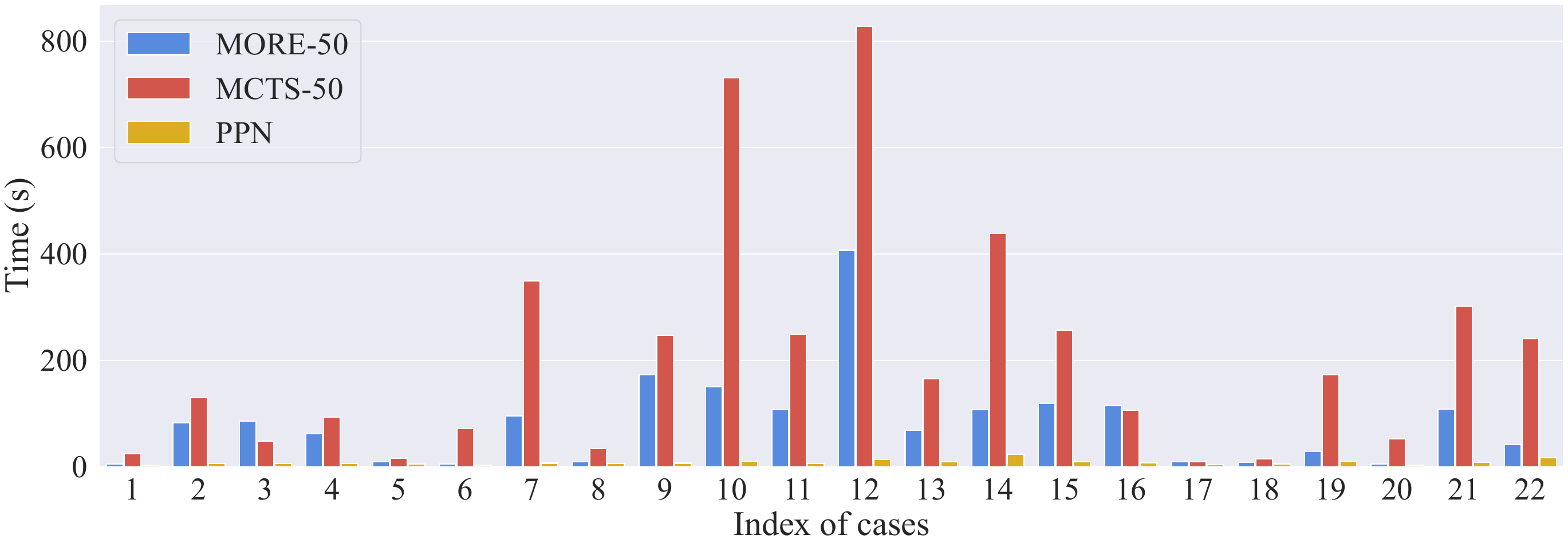}
    \caption{\label{fig:more-22-test-time}
        The average time (of 5 trials) used to solve one case for 22 cases.
        %\TODO{use log on y-axis may be more proper, but I think in this linear scale, the reader can feel the improvements more intuitively.}
    } 
\end{figure}

\textbf{Ablation Studies}
Although the data generated by MCTS for training \ppn is free because it is collected fully automatically in simulation, we set to explore data efficiency in training, which can be important for building larger models in practice.
For this purpose, we collected $243$ training cases ($65384$ transitions in 30 hours with PyBullet) with \mcts as described in~\autoref{sec:more-mcts}.
Training on \ppn on all data took approximately 22 hours.
As shown in~\autoref{fig:more-bar-num-time}, we tested \mcts and \ours with different budgets. Also, \ours is trained on different numbers of training data.
Clearly, the problem can be solved by all tested methods with fewer actions when the search iteration limits are increased.
But the time for solving the problem also increases as a consequence.
%, as for some cases, the stop condition of the \mcts is hard to reach.
%The number of training cases influences the performance of \ours, but not by much. The budget is the main factor. %to the number of action.
The proposed \ours technique can retrieve target objects with only $2.8$ executed actions and using only $10$ iterations of MCTS that last $36$ seconds on average. This is close to the best that MCTS without \ppn can achieve, $2.69$ actions, after $50$ iterations that last $208$ seconds. When we limit the number of iterations of MCTS (without \ours) to $10$, the number of executed actions increases to $3.19$, and the search time remains relatively high ($127$ seconds). This clearly demonstrates the superior performance of the proposed approach in terms of both time and action efficiency. 

\begin{figure}[ht!]
    \centering
    \includegraphics[width = \linewidth]{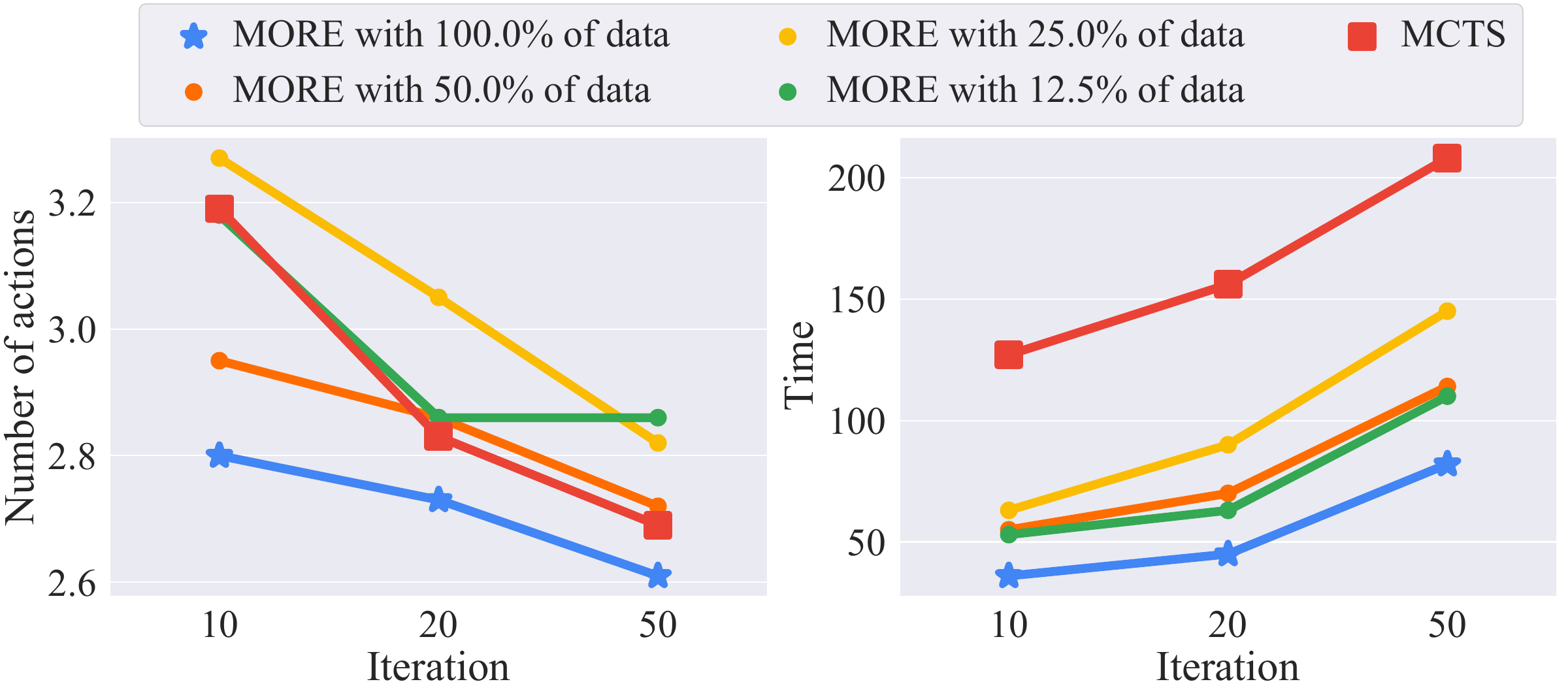}
    \caption{\label{fig:more-bar-num-time}
        Different amounts of training data are used to train \ppn, which are evaluated on \ours with different budgets (iteration).
        This is the evaluation of the 22 cases.
    } 
\end{figure}

\subsection{Robot Experiments}
We evaluated the four methods on six real test cases (four from~\cite{xu2021efficient} and two from~\autoref{chap:vft}).
These six test cases are representative in that they contain more objects and often require at least two push actions to solve.
For these real experiments, the results are shown in~\autoref{tab:more-real6table} and~\autoref{fig:more-real-num-time}. 
The budget of \mcts and \ours is limited to $10$ iterations. We note that the results for \gopg are taken from~\cite{xu2021efficient}. The execution time of \ppn is not listed in~\autoref{tab:more-real6table} as it is a near-constant small value as we had in the simulation experiments.
From the result, we observe only negligible performance degradation in comparison to simulation, which may be due to differences in friction, slight differences in the dimensions of the objects between simulation and real world, statistical error, or a combination of these.
Overall, the sim-to-real transfer was very successful and showed that \ours can learn in simulation and directly apply the learned skill to real-world tasks.
We assume models of objects are known, such that simple pose estimation can be used to locate objects in the real world and placed in simulation for planning. We could also use sophisticated tracking systems~\cite{wen2020se, wen2021bundletrack, mitash2020scene} for general purpose.

\begin{figure}[ht!]
    \centering
    \includegraphics[width = 0.15 \linewidth, trim = {0, 0, 0, 0}, clip]{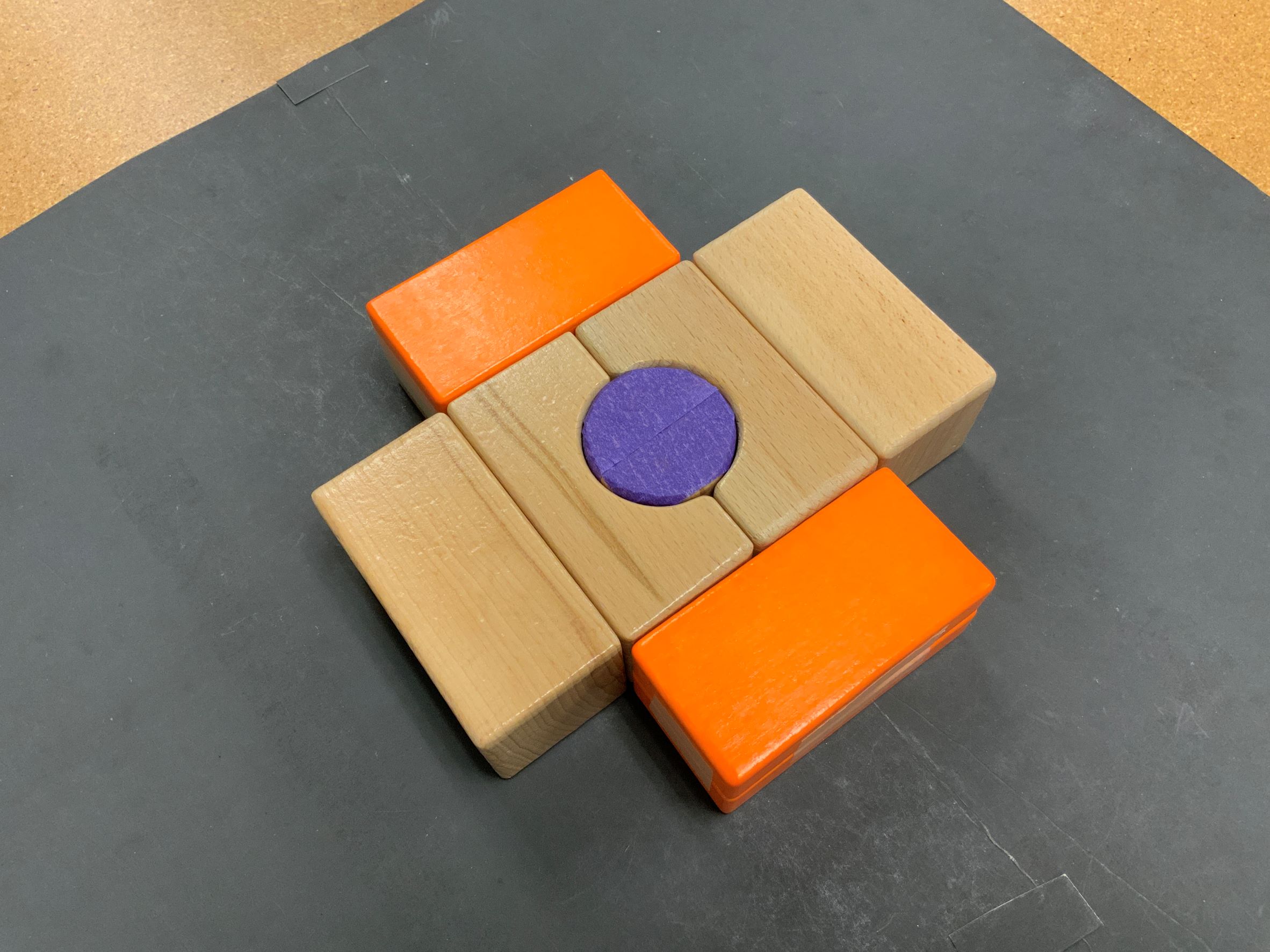} 
    \includegraphics[width = 0.15 \linewidth, trim = {0, 0, 0, 0}, clip]{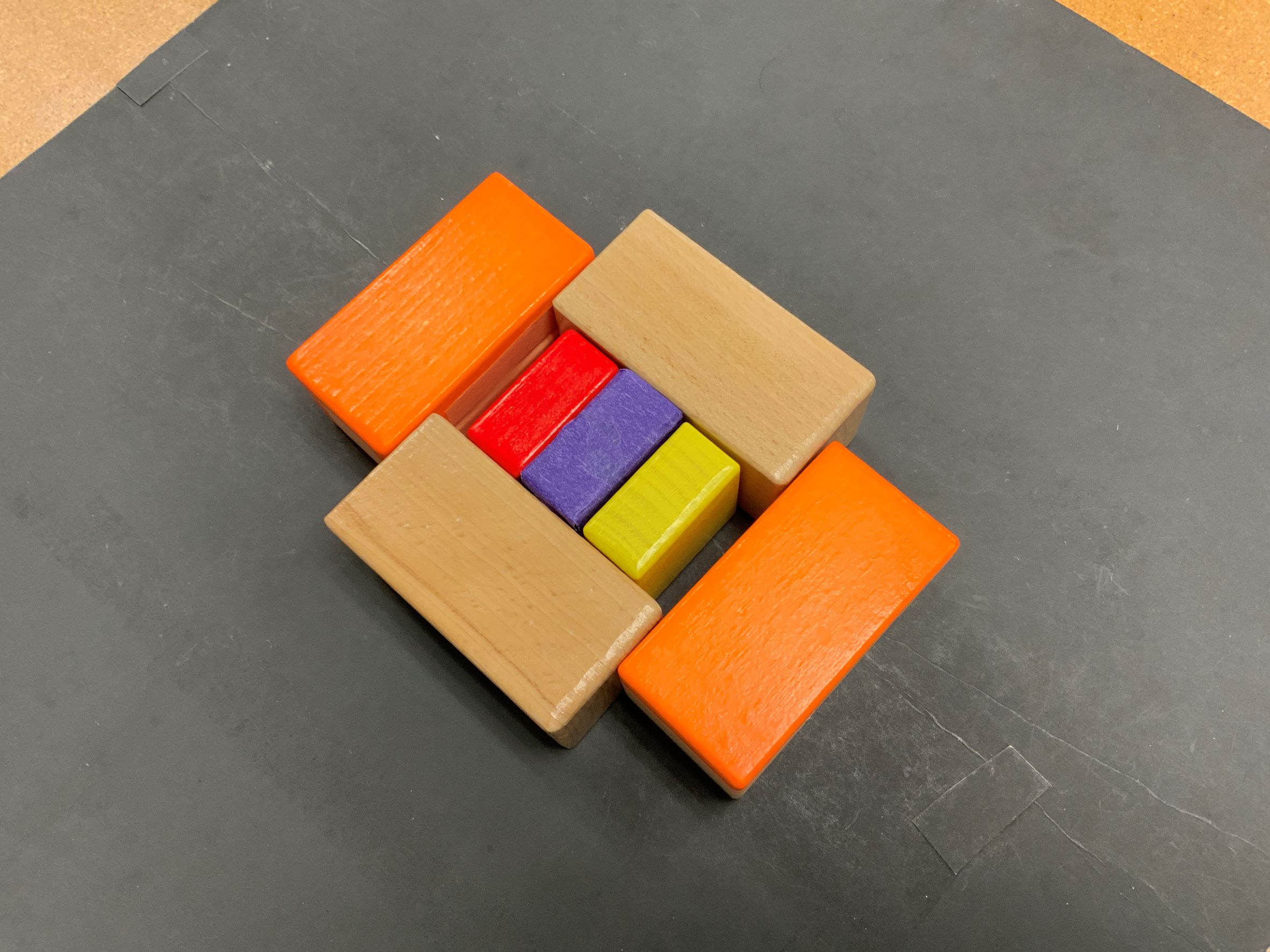} 
    \includegraphics[width = 0.15 \linewidth, trim = {0, 0, 0, 0}, clip]{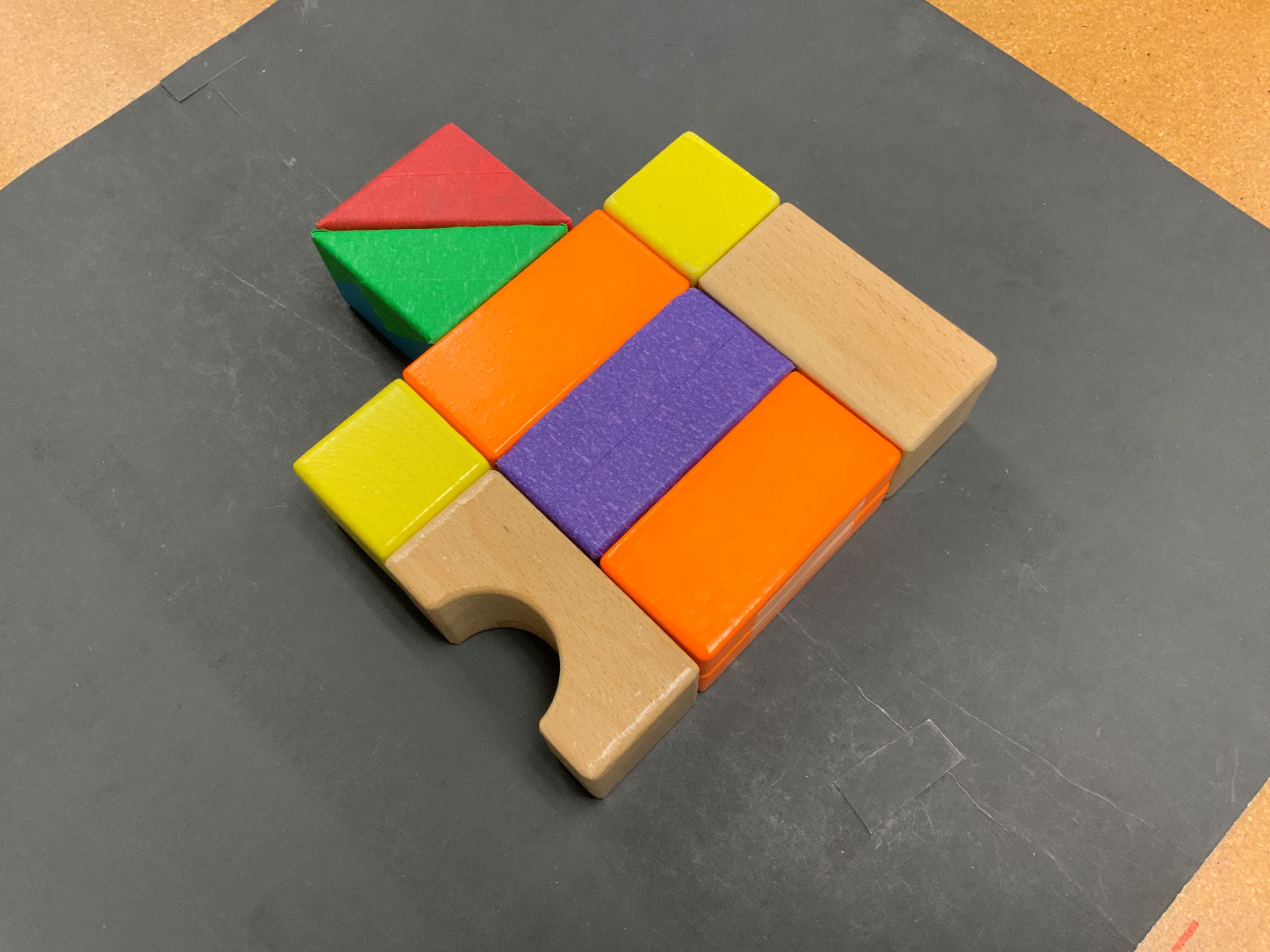} 
    \includegraphics[width = 0.15 \linewidth, trim = {0, 0, 0, 0}, clip]{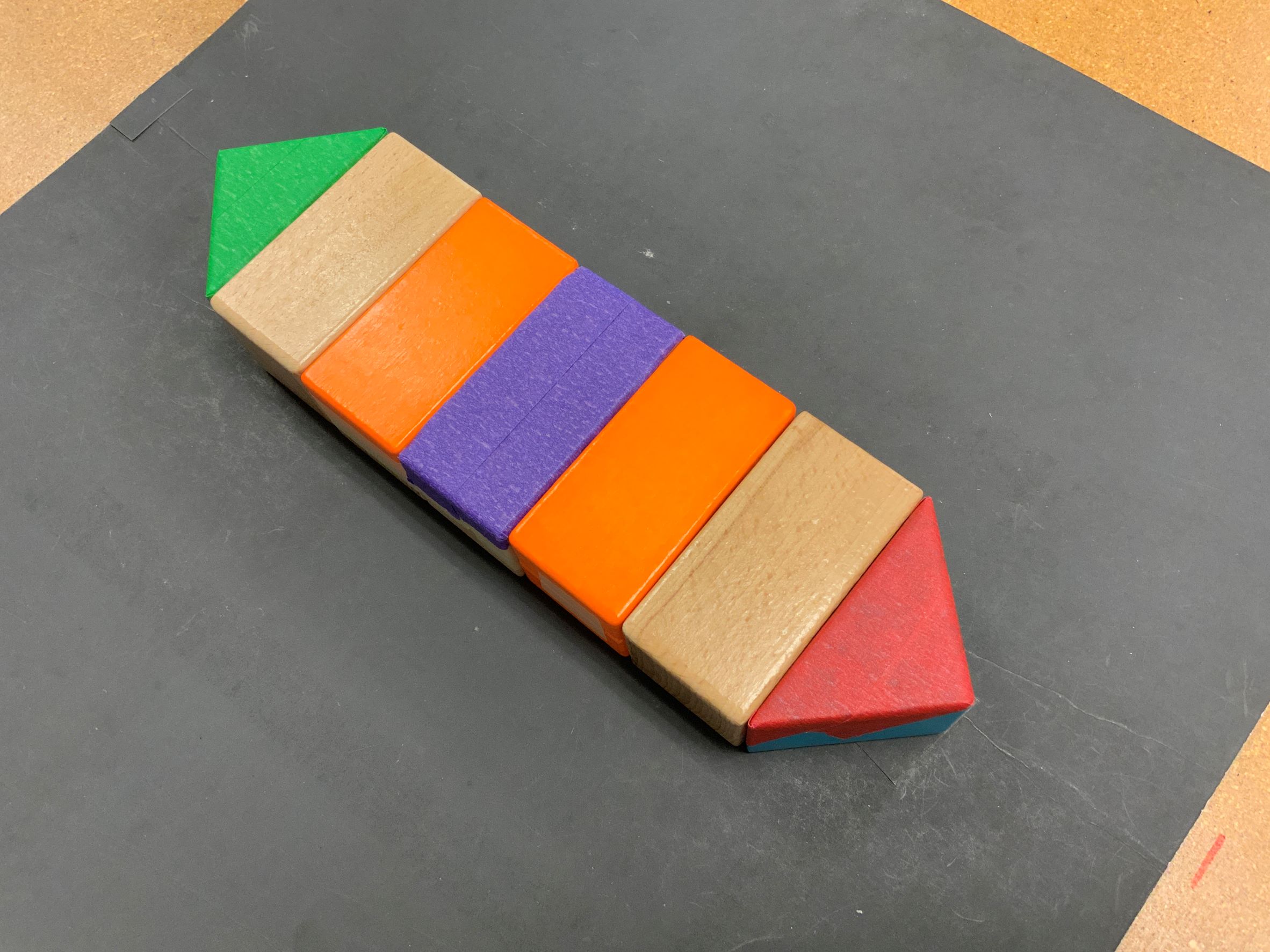} 
    \includegraphics[width = 0.15 \linewidth, trim = {0, 0, 0, 0}, clip]{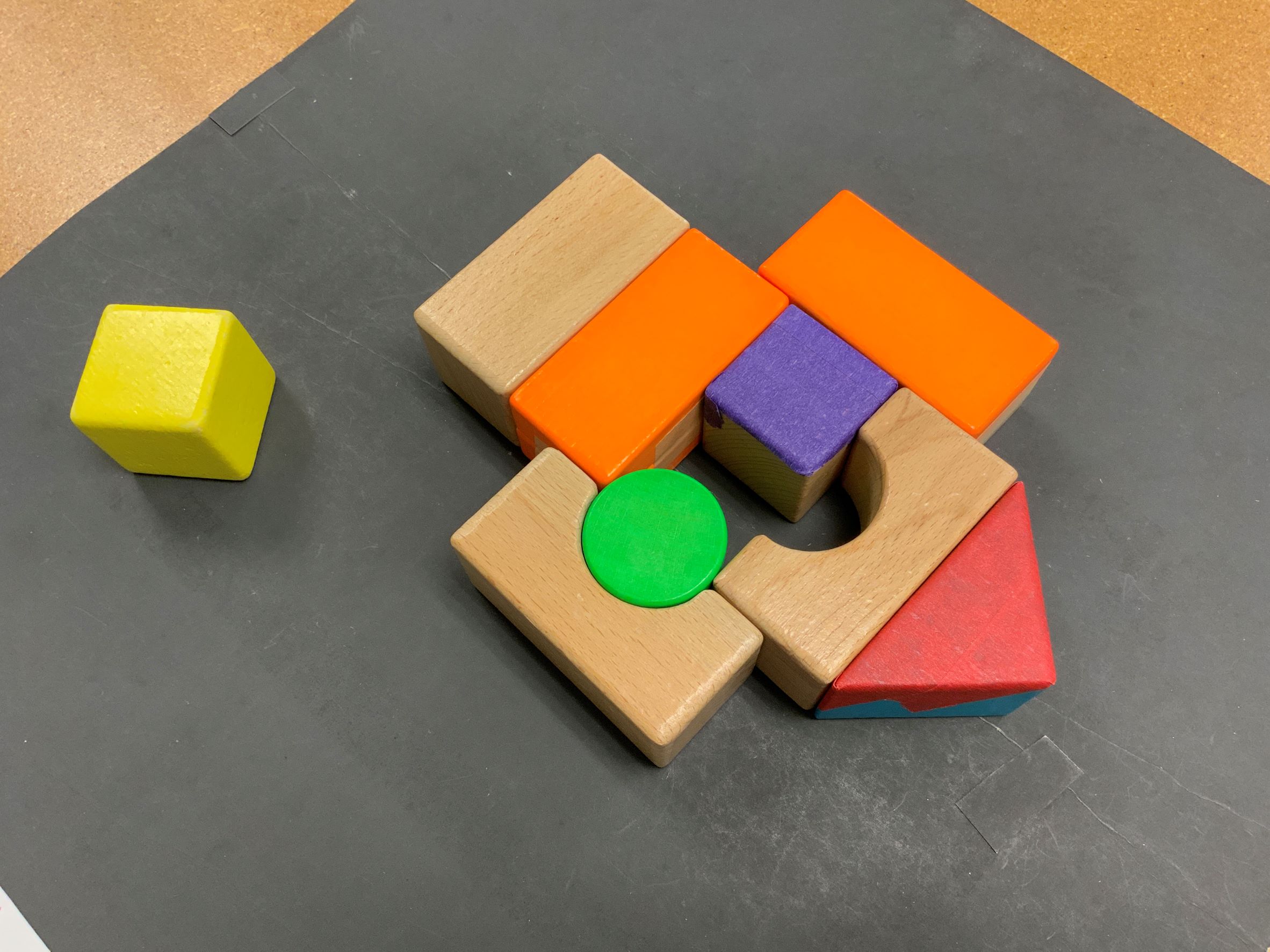} 
    \includegraphics[width = 0.15 \linewidth, trim = {0, 0, 0, 0}, clip]{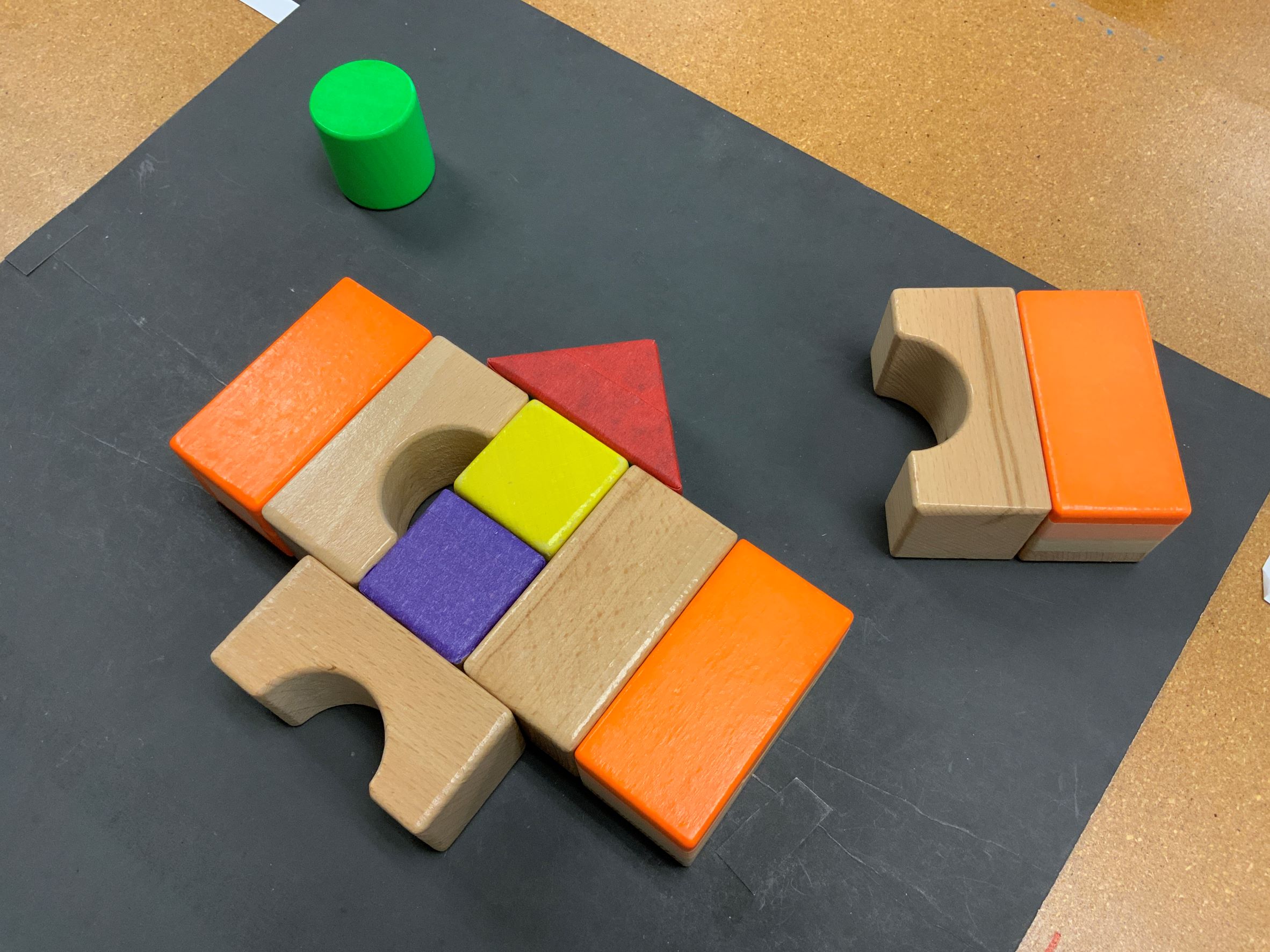}
    \vspace{1mm}
    \caption{\label{fig:more-real-cases}
    Manually generated cases similar to~\cite{xu2021efficient} and~\autoref{chap:vft}. The target object is masked in \textcolor{RoyalPurple}{purple}. These cases are used also in simulation experiments as shown in~\autoref{fig:vft-testcases}.
    }
\end{figure}

\begin{table}[ht!]
    \centering
    \begin{tabular}{c|c|c|c|c}
        & Num. of Actions & Time & Completion & Grasp Success   \\ \hline
        \ours-10 & $\mathbf{2.83}$ & $36s$ & $100\%$ & $100\%$ \\ \hline
        \mcts-10~\cite{huang2021visual} & $3.67$ & $190s$ & $100\%$ & $95.8\%$ \\ \hline
        \ppn & $3.72$ & $3s$ & $94.5\%$ & $95.8\%$ \\ \hline
        \gopg~\cite{xu2021efficient} & $4.62$ & $-$ & $95.0\%$ & $86.6\%$ \\ \hline
    \end{tabular}
\vspace{2mm}    
    \caption{Real experiment results for six cases as shown in~\autoref{fig:more-real-cases}. The budget of \mcts and \ours is limited to 10 iterations.
    For \gopg, only the first four cases apply, and results are from~\cite{xu2021efficient}. 
    Only planning time is recorded (robot execution was intentionally slowed down for safety).
    The computation time for \ppn to solve a task is 3 seconds on average (estimated).
    }
    \label{tab:more-real6table}
\end{table}

\begin{figure}[ht!]
    \centering
    \includegraphics[width = \linewidth]{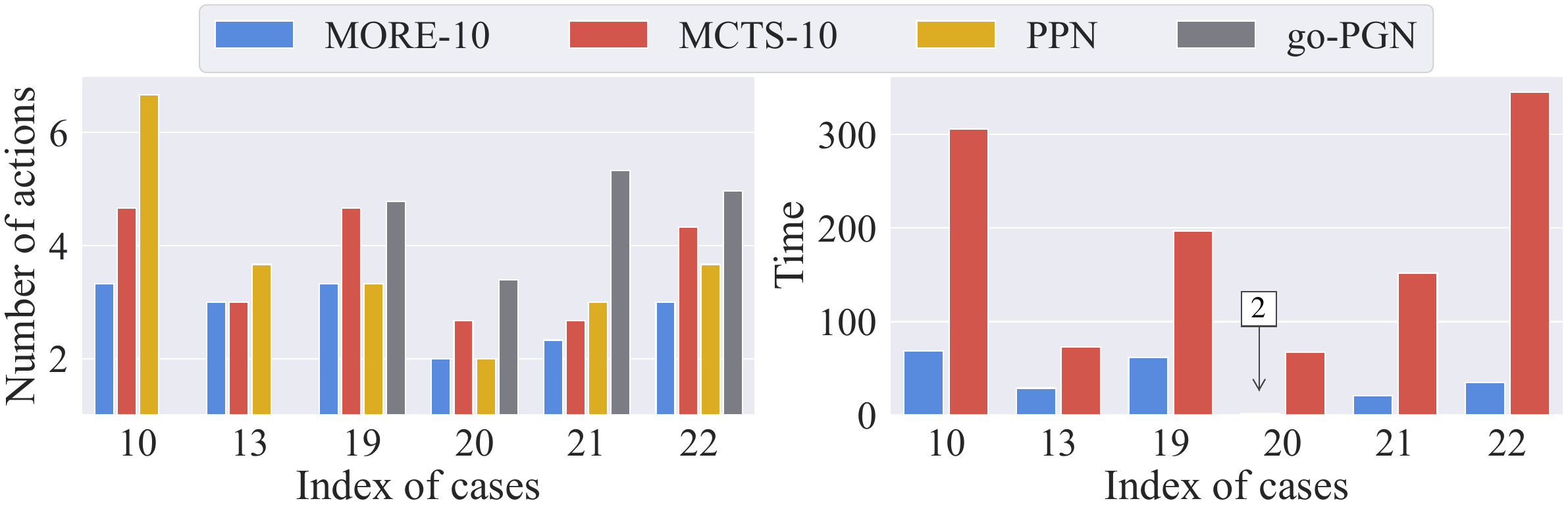}
    \caption{\label{fig:more-real-num-time}
        The number of action and time used on solving six cases.
        The budget is up to $10$ iterations for \mcts and \ours.
    } 
\end{figure}

\section{Summary}
The main limitation of this work is that we need to know the models of the objects to do the planning. One possible solution is instead of using an explicit simulator, we can use a learned model~\autoref{chap:dipn} to simulate the push results. Generalization to novel objects could then be possible.
We can further utilize the Push Prediction Network to estimate the simulation (rollout) result instead of using a physics engine. However, this can introduce additional uncertainties that typically result from using DNNs, which can cause unexpected behaviors such as pushing objects out of the workspace. 
Building on the know-how gained from developing \ours, we are exploring other real-world robotic manipulation tasks that would benefit from the S2$\to$S1 search-and-learn philosophy. 
%\TODO{
We point out that \ours can be further sped up by implementing a parallel version of \mcts, as we only utilized a single CPU thread in our implementation and \ppn (on GPU) is not being used most of the time. 
%}
    
\chapter{Parallel Monte Carlo Tree Search with Batched Rigid-body Simulations for Speeding up Long-Horizon Episodic Robot Planning}\label{chap:pmbs}
\thispagestyle{myheadings}

\section{Introduction}
The past decade has witnessed dramatic leaps in robot motion planning for solving problems that involve
sophisticated interaction between the robot and its environment, with milestones 
including teaching quadrupeds to perform impressive tricks \cite{RoboImitationPeng20,hwangbo2019learning} 
and navigate challenging terrains \cite{KumarA-RSS-21}, enabling high-DOF robot hands to 
solve the Rubik's cube \cite{andrychowicz2020learning}, and so on. While some of the
success can be attributed to the rapid advancement in deep learning \cite{krizhevsky2012imagenet} and deep reinforcement learning \cite{mnih2013playing},
another undeniable factor is the availability of fast, high-fidelity 
physics engines, including PyBullet \cite{coumans2021} and MoJuCo \cite{todorov2012mujoco}.
These physics engines allow the simulation of the physics of complex rigid-body systems, 
sometimes faster than real-time, which enables the collection of large amounts of 
realistic system behavior data without even touching the actual robot hardware. 
Nevertheless, most physics simulators are CPU-based, which can only simulate a limited 
number of robots simultaneously; this has led to some studies seeking parallelism by 
using a massive amount of computing resources. For example, the OpenAI hand study 
\cite{andrychowicz2020learning} used a total of $6,144$ CPU cores to train their model 
for over $40$ hours, which is costly and time-consuming.

As physics simulation starts to become a bottleneck in solving robotic tasks, GPU-based physics engines have recently begun to emerge, including Isaac Gym 
\cite{makoviychuk2021isaac} and Brax \cite{brax2021github}, to address the issue by enabling large-scale rigid body simulation. Early results are 
fairly promising; for example, the training of the OpenAI hand using Isaac Gym
can be done on a single GPU in one hour, translating to a combined resource-time saving of several orders of magnitude. Similar success has also been realized in applying reinforcement 
learning on quadrupeds, robotic arms, and so on \cite{makoviychuk2021isaac}. 
\begin{figure}[ht!]
\vspace{1.5mm}
    \centering
    \includegraphics[width = 0.9\linewidth]{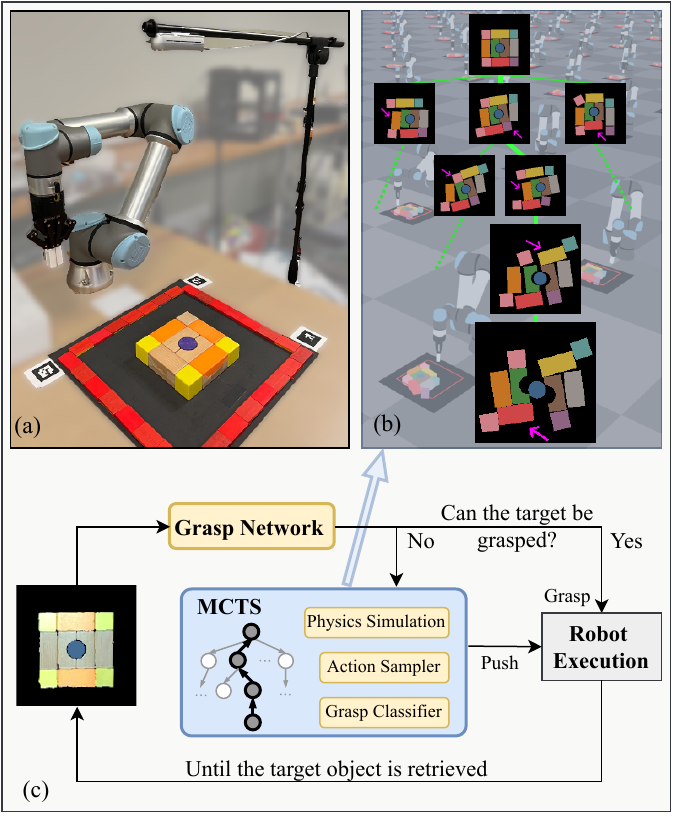}
    \caption{(a) The hardware setup includes a Universal Robots UR-5e with a Robotiq 2F-85 two-finger gripper and an Intel RealSense D455 RGB-D camera. (b) Planning and simulation carried out in physics simulator where thousands of virtual robots operate in parallel. (c) Overview of our system; the small blue cylinder at the center is the target object to be retrieved.
    }\label{fig:pmbs-overview}
\end{figure}

In this work, we exploit the power of large-scale rigid body simulation for optimally 
solving long-horizon episodic robotic planning tasks, such as multi-step object retrieval 
from clutter, leveraging the strength of another powerful tool that has attracted a 
great deal of attention -- Monte Carlo tree search (MCTS) \cite{coulom2006efficient}. 
MCTS demonstrates clear advantages in solving long-horizon optimization problems 
without the need for significant domain knowledge \cite{silver2017mastering}, and was already employed for solving challenging manipulation tasks \autoref{chap:vft} \cite{huang2021visual} and \autoref{chap:more} \cite{huang2022interleaving}. However, even with significant guidance using domain knowledge \autoref{chap:more}, MCTS incurs fairly long planning times due to its need of carrying out numerous rounds of sequential \emph{selection-expansion-simulation-backpropagation} cycles. 
The long planning time, sometimes several minutes per decision step, limits the
applicability of the methods from \autoref{chap:vft} and \autoref{chap:more} toward real-time decision making. 

Through combining MCTS and large-scale rigid-body simulation with Isaac Gym
\cite{makoviychuk2021isaac}, and carefully introducing parallelism into the mix, we
have developed a new line of parallel MCTS algorithms for efficiently solving 
long-horizon episodic robotic planning tasks. 
The development of the large-scale rigid-body simulation enabled parallel MCTS is 
the key contribution of this research, which is highly non-trivial. This is 
because MCTS has an inherently serial characteristic; as will be explained in more 
detail, the \emph{selection} phase of an MCTS iteration depends on the 
completion of the previous selection-expansion-simulation-backpropagation iteration. 
Fusing MCTS and Isaac Gym for solving long-horizon manipulation planning tasks also 
brings significant technical integration challenges because many computational 
bottlenecks must be addressed for the parallel MCTS implementation to be efficient.

We call our algorithm \textbf{P}arallel \textbf{M}onte Carlo tree search with 
\textbf{B}atched \textbf{S}imulation (PMBS). 
As its name suggests, PMBS realizes parallel MCTS computation through batched rigid-body simulation enabled by Isaac Gym.
Efficiently combining MCTS and Isaac Gym, PMBS achieves over $30\times$ speedups in planning efficiency for solving the task of object retrieval 
in clutter, while still achieving better solution quality, as compared to an optimized serial MCTS implementation, using identical computing hardware.
%For the same solution value, PMBS achieves $30\times$ speedup over serial MCTS.
PMBS drops the single-step decision making time to a few seconds on average, which is close to 
being able to solve the task in real-time. 
We further demonstrate that PMBS can be directly applied to real robot hardware 
with negligible sim-to-real differences.

\section{Problem Formulation}

In this paper, we task a robot equipped with a camera and a two-finger gripper to grasp a desired object from a densely packed clutter, as a concrete instance of 
long-horizon episodic robot planning problems. 
The workspace is a confined planar surface.
Two types of primitive actions are allowed: pushing and grasping.
All objects are rigid; the target object has a different color to facilitate its detection.
The only observation available to the robot is an RGB-D image that is taken by a top-down fixed camera, as shown in~\autoref{fig:pmbs-overview}. 
Every time the robot executes a push or a grasp action, a new image is taken.
A similar problem has been previously defined in \autoref{chap:dipn} \cite{huang2021dipn}, \autoref{chap:vft} \cite{huang2021visual}, and \autoref{chap:more} \cite{huang2022interleaving}.
Compared to~\cite{huang2021visual, huang2022interleaving}, the problem addressed in the present work is significantly more challenging to solve because the workspace is confined to a substantially smaller area, while keeping the number and sizes of objects the same. Consequently, the free space between the objects is reduced, and the robot needs to find a larger number of shorter surgical push actions in order to free the target object and grasp it. In fact, we found from our experiments (\autoref{sec:pmbs-experiments}) that the original setup considered in~\cite{huang2021visual, huang2022interleaving} can be solved using a brute-force parallel search in a GPU-based physics simulator, without a Monte Carlo tree search.

\section{Methodology} 
MCTS builds a search tree, balancing exploration and exploitation, by iteratively performing \emph{selection-expansion-simulation-backpropagation} operations. In the \emph{selection} phase, MCTS selects a best node to grow the tree. A popular node selection criterion is based on the \emph{upper confidence bound} (UCB) \cite{auer2002finite,kocsis2006bandit}, 
\begin{equation}\label{eq:pmbs-ucb1}
    \argmax\limits_{n' \in \text{children of } n} \frac{Q(n')}{N(n')} + c \sqrt{\frac{2\ln{N(n)}}{N(n')}}, 
\end{equation}
where $Q(n)$ is the sum of rewards collected starting from the state corresponding to node $n$, $N(n)$ is the number of times $n$ was selected so far. The selection process continues until it finds a node that corresponds to a terminal state or a node that has never-explored children.
We note that a node $n$ is always associated with a state $s$ and an observation $o$; sometimes a node $n$ and the corresponding state $s$ are used interchangeably. After a node $n$ is selected, if it is not a terminal node, it is \emph{expanded}, and its new child, say $n'$, is added to the search tree. Subsequently, a \emph{simulation} will be carried out at $n'$. This \emph{selection-expansion-simulation} process is repeated until a terminal state (or a stopping condition) is reached, which yields a reward. %If $n$ or $n'$ is a terminal node, an immediate reward is obtained without further simulation.
The obtained terminal reward is \emph{propagated back} from $n'$ all the way to the root node, while updating the sum of reward ($Q(n)$) and incrementing the number of visits ($N(n)$) for all the nodes along the path.

Effectively employing MCTS to tackle long-horizon episodic robot planning requires a highly non-trivial adaptation of MCTS. In this section, we first describe the necessary preparation for integrating MCTS and physics simulation for object retrieval, then describe augmentations to the architecture for GPU-based processing, and finally outline our key ideas and design choices in our parallelization effort.

\subsection{Serial MCTS for Object Retrieval from Clutter}\label{subsec:pmbs-smcts}
To use MCTS for the object retrieval task and solve real instances, we integrate it in a process that alternates between search in simulation and execution on the real system. 
Our MCTS process takes in a scene that is segmented into objects, and uses physics simulation to reason about the proper push actions to facilitate the final retrieval of the target object. %
An overview of the MCTS process is provided in~\autoref{fig:pmbs-overview}.
In other words, we first replicate in the simulator the real perceived scene at the beginning of each episode, perform computation and simulation, and then execute with the real robot the action that results from the simulation to guide the resolution of the retrieval task on the real objects. 

We now describe the details of our basic serial MCTS adaptation. For the selection step, the standard UCB formula is used. For the expansion step, for a selected node $n$ that has not been expanded, we sample many potential push directions by examining the contour of the objects. These sampled pushes become the candidate actions under $n$ for expansion. 
After a sampled push action is chosen, the action is executed in the physics simulation and a new node is added to the tree. The MCTS simulation step is then carried out with additional consecutive random pushes to obtain a reward for the newly added node. Note that for each simulation step, we must decide whether the resulting state is a terminal state; this is done using a \emph{grasp classifier}, to be explained later. 

An important design decision we make here, to render MCTS computation more tractable, is to limit the depth of the tree.
We limit the depth of the overall tree to be no more than some $d_T$.
The simulation can be carried out for at least $d_s$ steps.
This means that the maximum depth reached by MCTS does not exceed $d_T + d_s$.
If expansion happens at depth $d_T$, we allow the state to be simulated further until $d_T + d_s$.
Given our goal of finding the least number of pushes for retrieving the target object, $d_T$ and $d_s$ can potentially be dynamically updated when an identified terminal node has a depth $d$ smaller than $d_T$. In this case, we set $d_T = d$ and $d_s = 0$. We terminate an MCTS process if: (1) the elapsed time exceeds a preset budget $T_{\text{max}}$, (2) the tree is fully explored, or (3) the target can be grasped in an explored node and all nodes at its parent's level have been explored.

After each full MCTS run, we execute the best action it returns on the actual scene (simulated or real), and then use a \emph{grasp network} (GN) from \autoref{chap:more} to tell us whether the target object is retrievable. If it is, GN further tells us how to grasp it; the task is then completed. Details of GN, for replication purposes, can be found in the online supplementary material. 

\subsection{Adaptions for GPU}\label{subsec:pmbs-gpua}
Besides simulation, which can be sped up using GPU-based physics engines, there are three additional bottlenecks in the process to parallelize MCTS for object retrieval. One of these is the parallelization of MCTS itself and the other two are specific to the object retrieval problem: action sampling and grasp feasibility prediction. We leave the first bottleneck for \autoref{subsec:pmbs-pmbs} and address the latter two here. 

\begin{figure}[ht!]
    \centering
  \includegraphics[width=0.5\linewidth]{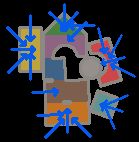}
  \caption{\label{fig:pmbs-action} 
    Sampled push actions.}
\end{figure}

\textbf{Speeding up Action Sampling}. 
Because the number of push action choices is uncountably infinite, action sampling is necessary. 
We modified the action sampler from \autoref{chap:vft} and \autoref{chap:more} with slight changes and a more efficient implementation.
As shown in~\ref{fig:pmbs-action}, for a given $o_t$, actions $a^p_t$ are sampled around the clutter.
$N_a$ actions are evenly sampled around the contour of each object, from edge to center.
Actions that cannot be executed due to collisions are discarded.
Further speedups are obtained by pre-computing the sampled actions for each object and only performing collision checking between the robot's start pose of push and objects at runtime.

\begin{figure}[ht!]
    \centering
  \includegraphics[width=0.8\linewidth]{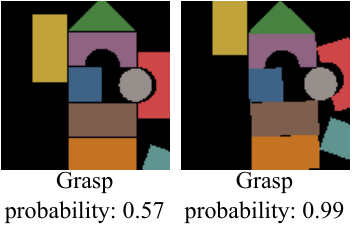}
  \caption{\label{fig:pmbs-grasp} 
    Examples of using the \emph{grasp classifier} to produce probabilities to grasp the object at center (\textcolor{RoyalBlue}{blue} in this case). Here we used an RGB image for illustration purpose (input should be a depth image).}
\end{figure}

\textbf{Grasp Evaluation}.
Previous learning-based methods for object retrieval use a \emph{grasp network} (GN) to evaluate the feasibility of grasping the target object (\autoref{chap:vft} and \autoref{chap:more}), which becomes a time-consuming bottleneck when parallelized directly.
GN is relatively slow because it evaluates a large number of possible grasp poses. However, knowing the best grasp pose is unnecessary if we only want to know how 'graspable' a state is. 
Given this observation, we develop a simplified \emph{grasp classifier} (GC) that only returns a grasp probability (\autoref{fig:pmbs-grasp}). 
GC is an EfficientNet-b0~\cite{tan2019efficientnet} that takes a depth image as input and outputs a probability  between 0 and 1.
Given a depth image and a target object, we can query GC whether the target object is graspable by comparing its output to a preset threshold $R_c^{*}$. Details about GC's implementation and training in simulation can be found in the online supplementary material. Note that GN is still used after each full MCTS run for potentially grasping the target object, as described in~\autoref{subsec:pmbs-smcts}.

\subsection{Parallel MCTS with Batched Simulation}\label{subsec:pmbs-pmbs}
Given the availability of powerful GPU-based physics simulators including Isaac Gym 
\cite{makoviychuk2021isaac} and Brax \cite{brax2021github}, which enables the simulation of a large number of systems independently and simultaneously, a natural route for speeding up long-horizon episodic robot planning tasks is to introduce parallelism into the MCTS pipeline outlined in \autoref{subsec:pmbs-smcts}, to perform many simultaneous simulations. 
However, it is challenging to introduce parallelism into MCTS because optimal node selection depends on the reward of all previous rounds.
To enable parallelism in MCTS for object retrieval from clutter and harness GPU-based simulation, we introduce the following modifications to the MCTS procedure outlined in \autoref{subsec:pmbs-smcts}.
We assume the number of parallel environments in the simulator is fixed to some number $N_e$. Each parallel environment contains an identical virtual robot and objects.

\textbf{Selection with Virtual Loss.}
By observing the operations of MCTS, it is not difficult to see that the parallelism of MCTS requires modifying the UCB formula. Otherwise, the same node in a search tree will be selected for expansion in multiple parallel environments, leading to redundancy and poor performance. 
To address this issue, we adopt the idea of \emph{virtual loss}~\cite{chaslot2008parallel}, which has shown to give good results in multiple application domains~\cite{Liu2020Watch, yang2021practical, silver2018general}. 
Virtual loss is used to adjust the calculation of UCB values for the nodes that have been selected but not yet expanded~\cite{chaslot2008parallel, mcts2012},
\begin{equation}\label{eq:pmbs-ucb2}
    \argmax\limits_{n' \in \text{children of } n} \frac{Q(n')}{N(n') + \hat{N}(n')} + c \sqrt{\frac{2\ln{(N(n) + \hat{N}(n))}}{N(n') +\hat{N}(n')}},
\end{equation}
where the $\hat{N}(n)$ is the number of selected but not yet expanded nodes under node $n$.
$\hat{N}(n)$ will be reset to zero once the selection phase of parallel MCTS is completed.
Basically, \autoref{eq:pmbs-ucb2} penalizes selecting nodes that have already been selected in some other parallel environment but for which expansion and simulation have not yet been completed. 
With Eq.~\eqref{eq:pmbs-ucb2}, it is still possible for a node $n$ to be selected multiple times, which may lead to redundant simulations. 
To avoid this and ensure that no redundant simulations are carried out, we mark all selected actions of node $n$ and share this information across all the parallel environments. 

A collection of state-action pairs is returned from the selection phase. 
The same state could be selected many times, but all state-action pairs in the selected collection are unique.
For example, in \autoref{fig:pmbs-parallel-mcts}, upper left, $(s_{t+2}^1, a^1), \ldots, (s_{t+1}^4, a^4)$ are four such state-action pairs.
Batch-mode expansion/simulation on this collection is then performed in parallel using GPU.

\begin{figure}[ht!]
\vspace{2mm}    
    \centering
    \includegraphics[width = 0.99\linewidth]{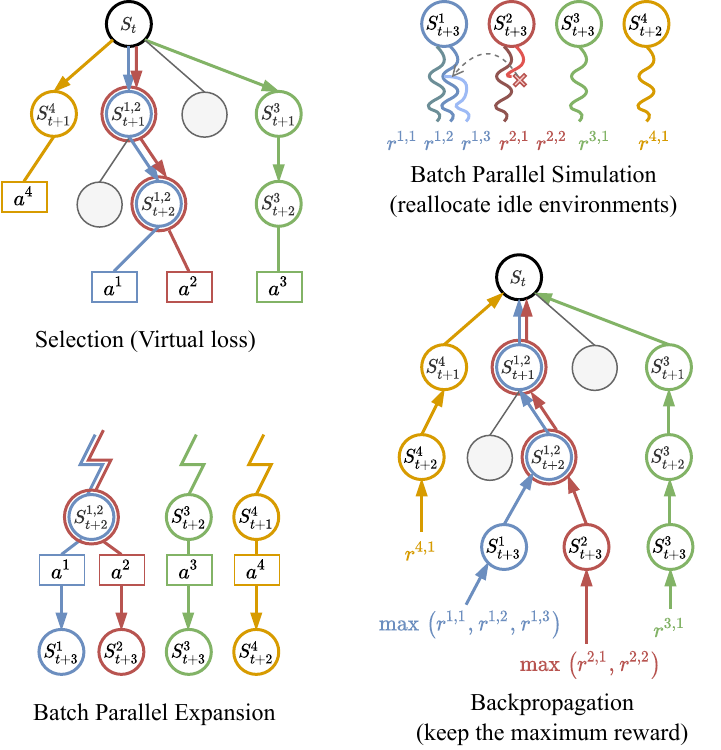}
    \caption{Steps in PMBS, our parallel MCTS with batched operation.\label{fig:pmbs-parallel-mcts}
    }
\end{figure}

\textbf{Batch Expansion.}
After a batch of state-action pairs ($\{(s, a)\}$) has been selected, the expansion step is carried out for all of these pairs simultaneously. For this purpose, environments in the simulator (Isaac Gym) will be loaded with the appropriately selected states, after which expansion (transition) is carried out in parallel. A batch expansion creates a set of new nodes added to the tree, each of which is different. 
In \autoref{fig:pmbs-parallel-mcts}, lower left, $s_{t+3}^1, \ldots, s_{t+2}^4$ are the result of expanding $(s_{t+2}^1, a^1), \ldots, (s_{t+1}^4, a^4)$, respectively.

\begin{algorithm}[ht]
    \begin{small}
    \DontPrintSemicolon
    \SetKwFunction{FMain}{Main}
    \SetKwFunction{FMCTS}{Parallel-MCTS}
    \SetKwFunction{FSelect}{Selection}
    \SetKwFunction{FExpand}{Expansion}
    \SetKwFunction{FSimulation}{Simulation}
    \SetKwFunction{FBackpropagation}{Backpropagation}
    \Fn{\FMain{$s_t, o_t$}}{
        \While{\normalfont there is a target object in workspace}{
            % $P(o_t) \gets \gn(o_t)$ \;
            % $P_m(o_t)=P(o_t)\cap M(o_t)$ \tcp*[f]{Only grasp the target object}\;
            % \If{\normalfont $\max_{a^{g}} P_m(o_t) > P^*$}{Execute $\argmax_{a^{g}}P_m(o_t)$ \tcp*[f]{Grasp}}
            \If{\normalfont the target object can be grasped (query GN)}{Execute grasp of the target object}
            \lElse{Execute \FMCTS{$s_t$}\tcp*[f]{Push}}
        }
    }
    \vspace{1mm}
    \Fn{\FMCTS{$s$}}{
        Create root node $n_0$ with state $s$ \;
        $\text{es\_level} \gets 1$ \tcp*[f]{Early stop level}\;
        $\text{graspable\_nodes} \gets \varnothing$ \;
        \While {\normalfont (within time budget) and (depths of all graspable\_nodes are greater than es\_level)}{
            % $[(n^1,a^1), \dots, (n^{N_\text{e}},a^{N_\text{e}})] \gets \FSelect(n_0)$ \tcp*[f]{node and action pairs} \; 
            $[(n^1,a^1), \dots, (n^{N_\text{e}},a^{N_\text{e}})] \gets \FSelect(n_0)$  \; 
            
            Reset all $\hat{N}(n)$ to 0 \;
            
            % $[{n'}^1, \dots, {n'}^{N_\text{e}}] \gets \FExpand([(n^1,a^1), \dots, (n^{N_\text{e}},a^{N_\text{e}})])$ \tcp*[f]{n' is the child of n}\;
            $[{n'}^1, \dots, {n'}^{N_\text{e}}] \gets \FExpand([(n^1,a^1), \dots, (n^{N_\text{e}},a^{N_\text{e}})])$ \;
            
            \For{\normalfont$n'$ in $[{n'}^1, \dots, {n'}^{N_\text{e}}]$}{
                \If{\normalfont $GC(n'(o)) > R_c^{*}$}{
                    $\text{graspable\_nodes} \gets \text{graspable\_nodes} \cup \{n'\}$ 
                }
            }
            
            \If{\normalfont all nodes at $\text{es\_level} - 1$ are fully expanded or terminal}{
                $\text{es\_level} \gets \text{es\_level} + 1$
            }
            
            $[r^1, \dots, r^{N_\text{e}}] \gets \FSimulation([n_1', \dots, n_{N_\text{e}}'])$\;
            
            \FBackpropagation{$[({n'}^1, r^1), \dots, ({n'}^{N_\text{e}},r^{N_\text{e}})]$}
            
            \Return the $a^p$ that leads to best child node of root, ranked by~\autoref{eq:pmbs-ucb2}
        }
    }
    
    \caption{\label{alg:pmbs-pmcts} Parallel MCTS with Batched Simulation}
    \end{small}
\end{algorithm}

\textbf{Batch Simulation.}
The batch simulation step of our parallel MCTS implementation is similar to the batch expansion step, but with additional steps inserted before and after. Before a push simulation, random actions must first be selected, using the action sampling method outlined in \autoref{subsec:pmbs-gpua}. After each push simulation, GC, as described in \autoref{subsec:pmbs-gpua} is applied to evaluate the outcome.
As we can see, to best exploit the parallelism from the simulator, action sampling and GC should be carried out as efficiently as possible, so that they do not become significant computational bottlenecks. 

During simulation, we also perform \emph{leaf parallelization} \cite{chaslot2008parallel} when the number of simulation environments is more than the number of states for which MCTS simulations are to be carried out. This is reflected in \autoref{fig:pmbs-parallel-mcts}, upper right, where the first two states each are simulated twice initially. 
If some environments, after a push, are predicted by GC as graspable, then further simulation on these environments will not be carried out, and these environments can be re-purposed. For example, in \autoref{fig:pmbs-parallel-mcts}, upper right, a simulation under the second state terminates early, and the associated environment can be used to perform additional simulation for the first state. 

\textbf{Backpropagation.}
The backpropagation phase is straightforward to execute, as it simply backpropagates the rewards to the root of the tree. 
We note that, for a single state for which multiple simulations are carried out, it is natural to select the maximum reward obtained instead of taking averages (see~\autoref{fig:pmbs-parallel-mcts}, lower right).

The pseudo-code of PMBS is given in \autoref{alg:pmbs-pmcts} with the selection subroutine given in \autoref{alg:pmbs-pmcts-select}. Other subroutines of PMBS are mostly straightforward. 

\begin{algorithm}
    \begin{small}
    \DontPrintSemicolon
    \SetKwFunction{FSelect}{Selection}
    \SetKwFunction{FExpand}{Expansion}
    \SetKwFunction{FRollout}{Rollout}
    \SetKwFunction{FBackpropagation}{Backpropagation}
    \Fn{\FSelect{$n_0$}}{
        $\text{Pairs} \gets \varnothing$ \;
        \While{\normalfont $len(\text{Pairs}) < N_\text{e}$}{
            $n^i \gets \text{traverse tree until a leaf node using~\autoref{eq:pmbs-ucb2}}$ \;
            $a^i \gets \text{one sampled action of } n^i$ \;
            Remove $a^i$ from the sampled actions of $n^i$ \;
            $\text{Pairs} \gets \text{Pairs} \cup \{(n^i, a^i)\}$ \;
            $\hat{N}(n^i) \gets \hat{N}(n^i) + 1$ \;
            increment virtual counts of ancestor nodes of $n^i$
        }
        \Return Pairs
    }
    \caption{\label{alg:pmbs-pmcts-select}
    Selection with Virtual Loss}
    \end{small}
\end{algorithm}

\section{Experimental Evaluation} \label{sec:pmbs-experiments}

We evaluated the proposed system (PMBS) in a physics simulator (Isaac Gym) and on a real robot on adversarial test cases.
In comparisons to baseline and ablation studies, we observe significant improvements using the GPU-based physics simulator together with parallel MCTS, which brings episodic decision making for real robots closer to real-time, i.e., a single complex decision is made in a few seconds.
All experiments were conducted on a desktop with an Nvidia RTX 2080Ti GPU, an Intel i7-9700K CPU, and 32GB of memory.

\subsection{Simulation Studies}
In this work, the simulated environment is built with Isaac Gym~\cite{makoviychuk2021isaac}, consisting of a Universal Robot UR5e with a two-finger gripper Robotiq 2F-85, and an Intel RealSense D455 RGB-D camera overlooking a tabletop workspace as shown in \autoref{fig:pmbs-overview}.
The robot is in position-control mode; push and grasp actions control the end-effector's position, and Inverse Kinematics (IK)~\cite{hawkins2013analytic, feng2021team} is applied to convert these to joint space commands.
The effective workspace is at a size of $0.288\times0.288$m, discretized as a grid of $144\times144$ cells where each cell is one pixel in the image (orthographic projection) taken by the camera.
% 0.288 - 144, 0.448 - 224
%
The workspace, in comparison to previous \autoref{chap:vft} and \autoref{chap:more}, is significantly smaller (only about $45\%$ in terms of area), making the setting much more challenging. 
We intentionally selected the setting to demonstrate the power of PMBS. 

All objects should reside in the workspace. %by considering the center of objects.
20 cases from \autoref{chap:vft} used for evaluation can be found in \autoref{fig:pmbs-cases}, where the red lines denote the boundary to which objects centers must be confined at all times.

\begin{figure}[ht!]
    \centering
    \includegraphics[width=\linewidth]{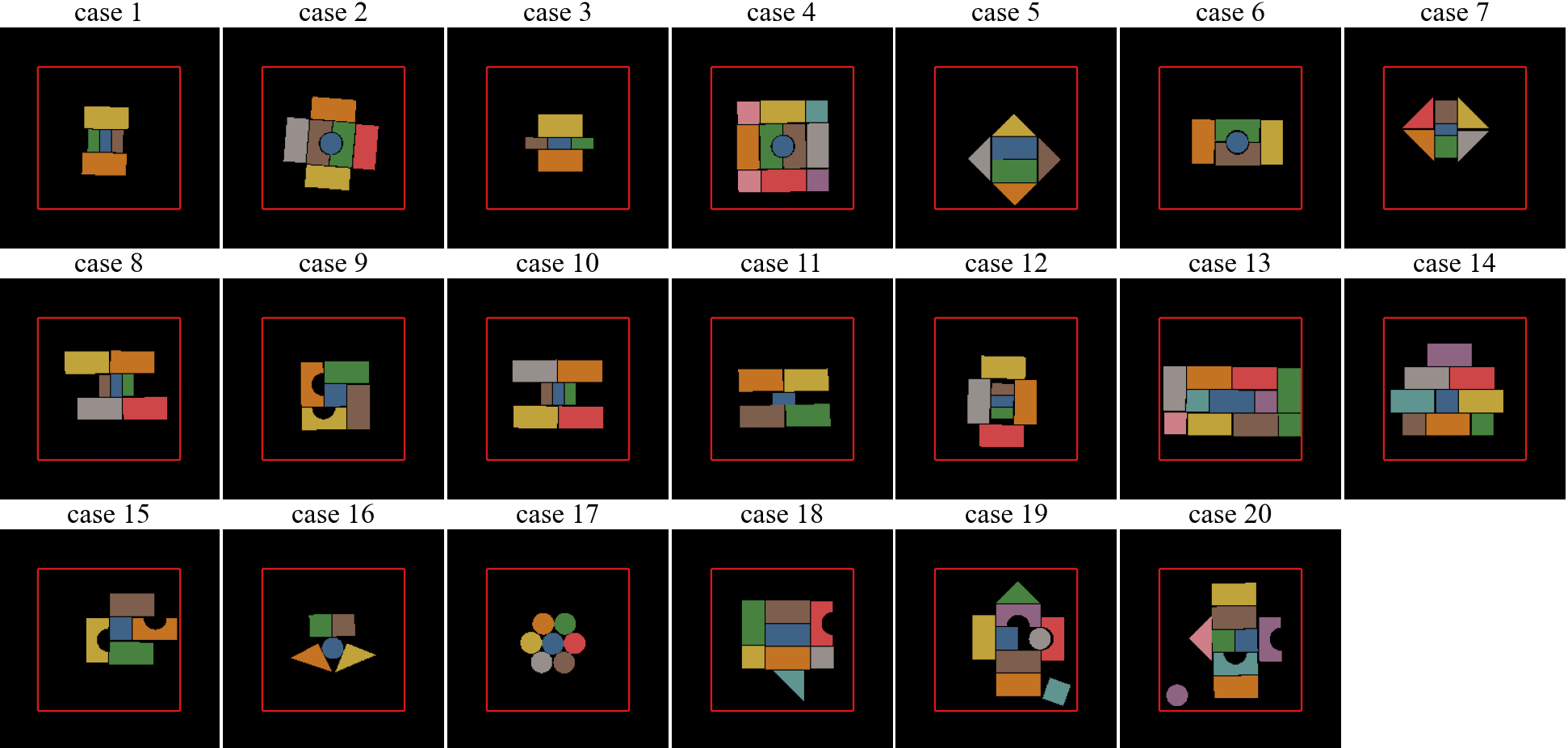}
    \caption{\label{fig:pmbs-cases}
        20 cases from \autoref{chap:vft} used in simulation experiments, where the target object has a \textcolor{RoyalBlue}{blue} mask. No object should exceed the boundary (\textcolor{red}{red} lines).
    } 
\end{figure}

The push distance of a push action is fixed at $5$cm ($10$cm in previous \autoref{chap:vft} and \autoref{chap:more}, the effective push distance is around $3$cm (the distance objects are moved).

\textbf{Metrics.}
Four metrics are used to evaluate our systems: 
\begin{enumerate}
    \item the number of actions used to retrieve the target object
    \item the total planning time used for retrieving the target object (build the tree)
    \item the completion rate in retrieving the target object within $16$ actions
    \item the grasp success rate, which is the number of successful grasps divided by the total number of grasping attempts
\end{enumerate}

\textbf{Baseline.}
We use an optimized serial MCTS implementation as the baseline, where the number of environments used for MCTS is one.
The following hyperparameters are used across all methods in benchmark unless otherwise mentioned.
The discount factor $\gamma=0.8$.
The default maximum tree depth is $d_T=7$, and the default simulation (rollout) depth is $d_s=3$.
The threshold of GC is $R_c^{*} = 0.9$.
The UCB exploration term $c$ in~\autoref{eq:pmbs-ucb1} and~\autoref{eq:pmbs-ucb2} is $0.3$.
The time limit (budget) $T_{\text{max}}$ for one step planning is $60$ seconds.
$1000$ robots (environments) in Isaac Gym are used in our PMBS; it takes around $2.2$ seconds for all robots to complete one push action.

We evaluate the performance of PMBS and the baseline serial MCTS over all 20 cases, running each case five times. For the evaluation, we set a time budget $T_{\text{max}} = 60 \text{ s}$ and denoted the two methods as PMBS-60 and MCTS-60, respectively. 
The summary benchmark for these two methods can be found in the first two rows of \autoref{tab:pmbs-sim20table}; individual results for each case can be found in \autoref{fig:pmbs-20-sim-num} and \autoref{fig:pmbs-20-sim-time}.

We make some observations based on the results. First, PMBS outperforms the serial MCTS version in terms of number of actions and computation time across all cases, which is as expected because PMBS engages many environments to facilitate its search effort. On the other hand, when we view the solution quality and computation time together, the advantage of PMBS over serial MCTS is significant: PMBS-60 uses 35 seconds on average for planning, whereas MCTS-60 uses over 300 seconds. This along translates to an $8.6\times$ speedup. At the same time, PMBS-60 uses $70\%$ fewer actions in solving the tasks. A further data point regarding the speedup at the same solution is given a bit later in \autoref{fig:pmbs-20-sim-ablation}.

A second observation is that, despite the fact that we are dealing with a difficult long-horizon planning problem, PMBS is able to achieve planning that is close to being able to perform reasoning in real-time, as it takes an average of $35/3.91 < 9$ seconds to make a single decision. With further optimization and/or better hardware, we believe that PMBS will achieve real-time decision-making capability for the current set of object retrieval tasks.

\begin{figure}[ht!]
    \centering
    \includegraphics[width = .97\linewidth]{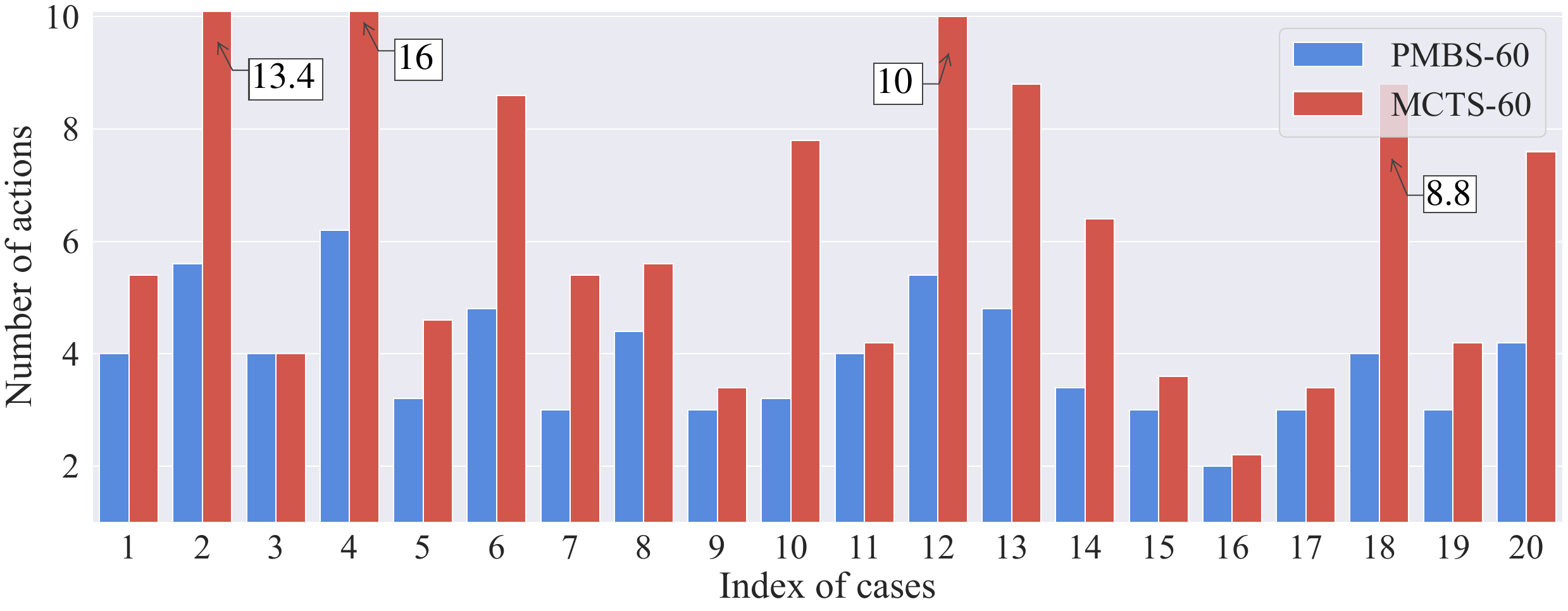}
    \caption{\label{fig:pmbs-20-sim-num}
        The average number (over five independent trials) of actions per case needed for solving the twenty cases, given a time budget of 60 seconds.
    } 
\end{figure}

\begin{figure}[ht!]
    \centering
    \includegraphics[width = .97\linewidth]{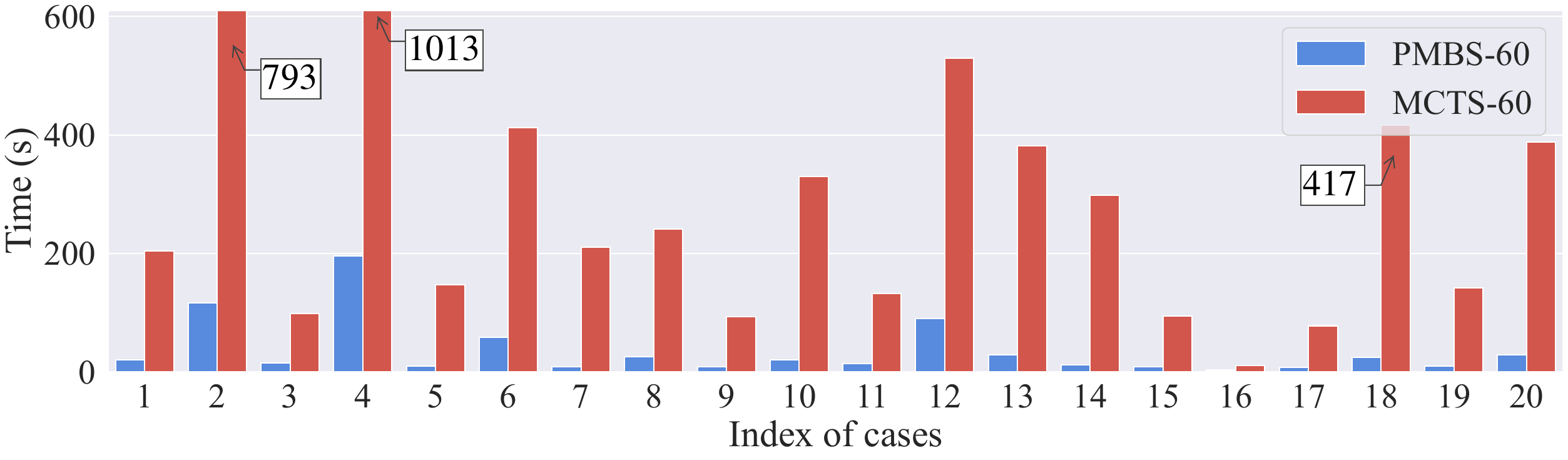}
    \caption{\label{fig:pmbs-20-sim-time}
        The average time (over five independent trials) per case needed for solving the twenty cases, given a time budget of 60 seconds.
    } 
\end{figure}

\begin{table}[ht!]
    \centering
    \scalebox{0.95}{\begin{tabular}{c|c|c|c|c}
        & Num. of Actions & Time & Completion & Grasp Success   \\ \hline
        PMBS-60 & $\mathbf{3.91}$ & $\mathbf{35}$s & $100\%$ & $98.3\%$ \\ \hline
        MCTS-60 & $6.67$ & $301$s & $93.0\%$ & $96.4\%$ \\ \hline
        PMBS-60 ($c = 0$) & $4.03$ & $113$s & $100.0\%$ & $99.2\%$ \\ \hline
        PMBS-60 ($c = \infty$) & $12.71$ & $147$s & $42.0\%$ & $96.7\%$ \\ \hline
    \end{tabular}}
    \caption{Simulation experiment results for 20 cases. Time budgets are limited up to 60 seconds.}
    \label{tab:pmbs-sim20table}
\end{table}

\textbf{Ablation Study.}
The time budget $T_{\text{max}}$ is one of the main factors that influence the solution quality and planning time.
To understand its role, several time budgets are used to evaluate our method, as shown in~\autoref{fig:pmbs-20-sim-ablation}.
Given more time for tree search, serial MCTS and PMBS could improve the solution quality, leading to fewer required actions.
The trends of serial MCTS (number of action) are steep, as it is highly possible that it could not find a solution given a limited time. 
The trend of PMBS (planning time) is more gradual, as the most time-consuming search happens in the first few iterations, which usually uses all the time budget.
While serial MCTS never achieves the same solution quality as PMBS, comparing the first PMBS data point and the last serial MCTS data point, we observe a $855/28 = 30\times$ speedup with PMBS still has some quality advantage. 

On the flip side, we note that the speed-up of $30\times$ seems small considering that we used $1000$ environments. This is due to two factors. First, MCTS is itself a serial process;  
parallelization will incur performance loss. Second, while we have improved many bottlenecks, e.g., on action sampling and grasp classification, the object retrieval task contains many elements that cannot be readily parallelized. 

We also evaluated the impact of the exploration and exploit trade-off on PMBS.
If the $c$ in \autoref{eq:pmbs-ucb2} is set to $0$, i.e., pure exploitation, PMBS-60 uses $4.03$ actions and $113$ seconds (planning time) on average on 20 cases.
The performance is worse than when $c=0.3$, as shown in~\autoref{tab:pmbs-sim20table}.
This is expected as the greedy approach could be stuck in a local optimum.
PMBS-60 is also tested by setting $c$ to be a large number in~\autoref{eq:pmbs-ucb2}, i.e., pure exploration, which uses $12.71$ actions and $147$ seconds (planning time) on averages; the completion rate has a steep drop to $42.0\%$.

\begin{figure}[ht!]
    \centering
    \includegraphics[width = \linewidth]{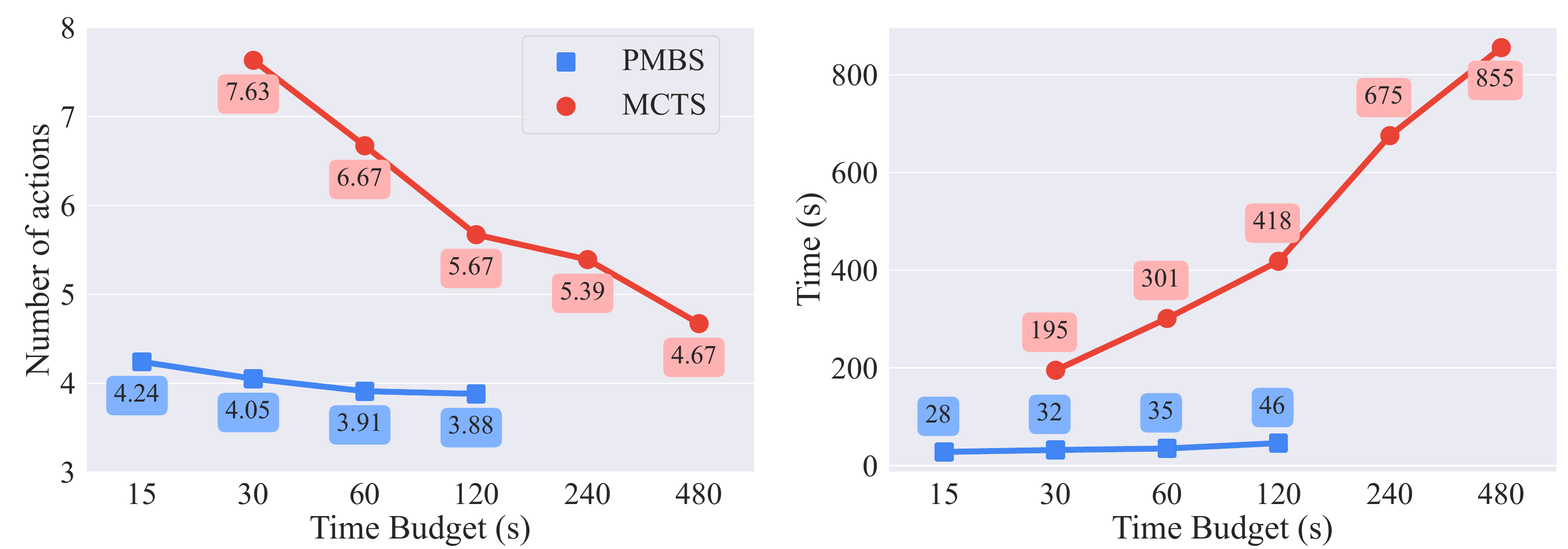}
    \caption{\label{fig:pmbs-20-sim-ablation}
        PMBS and serial MCTS evaluated with different time budgets. The reported values are averages over all 20 cases.
    } 
\end{figure}

\subsection{Real Robot Experiments}
For experiments on the physical UR-5e, the input to PMBS is a single RGB-D image.
A $1280\times720$ RGB-D image is taken, then it is orthogonally projected over the workspace of resolution of $144\times144$ (with cropping). 
Since the same robot and objects are used in both simulation and the real world, we can observe and act on a real robot but plan in a simulator.
For each object, simple pose estimation is performed to transfer the perceived scene from real images to the physics simulator environments.

The pose estimation is done by firstly extracting masks for objects from the image, then a brute-force matching between detected mask and recorded mask is performed for each object. We could achieve it at 0.15 seconds for one image (around 10 objects).
Serial MCTS and PMBS were evaluated the same way as in simulation experiments, except we only run tests on the six most challenging cases.
Individual benchmarks on six cases can be found in~\autoref{fig:pmbs-real-num-time}.
Average statistics are listed in~\autoref{tab:pmbs-real6table}.
We observe minimal sim-to-real performance loss; 
A small gap exists between the real and the simulation experiments, mainly due to pose estimation errors and mismatch of physics properties.

\begin{figure}[ht!]
    \centering
    \includegraphics[width = \linewidth]{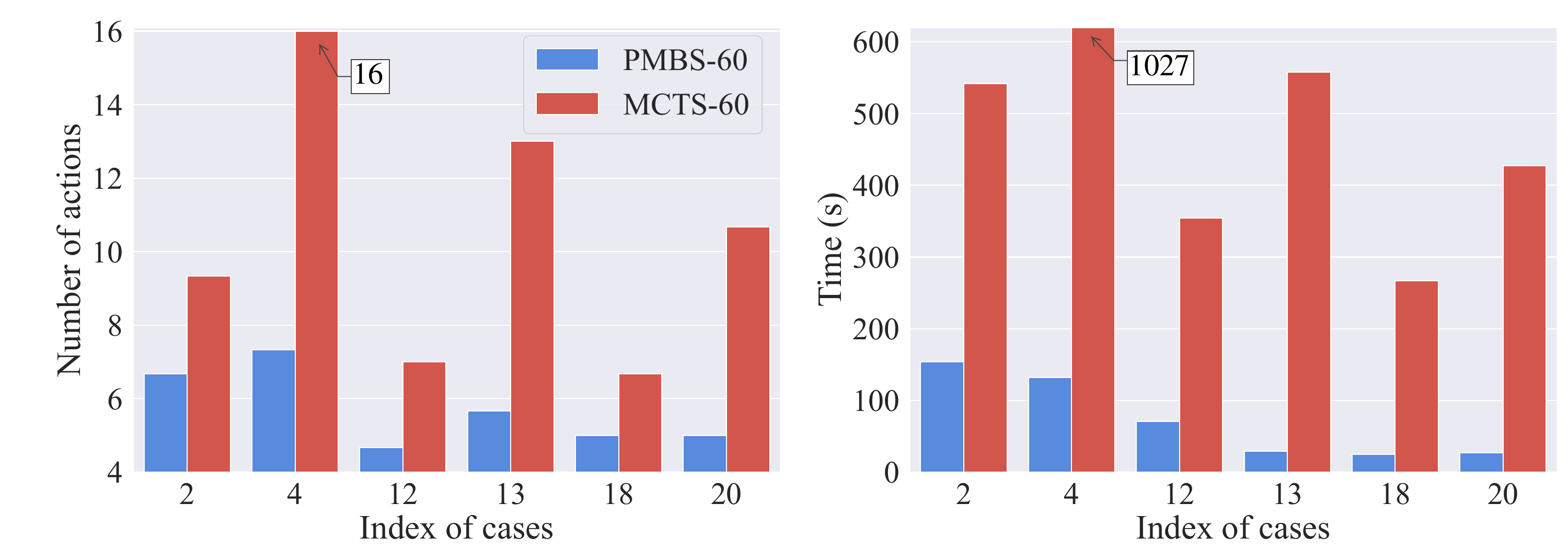}
    \caption{\label{fig:pmbs-real-num-time}
        The number of actions and time used for solving the six most challenging cases on the physical robot.
        The time budget is $60$ seconds.
    } 
\end{figure}

\begin{table}[ht!]
    \centering
    \begin{tabular}{c|c|c|c|c}
        & Num. of Actions & Time & Completion & Grasp Success   \\ \hline
        PMBS-60 & $\mathbf{5.72}$ & $\mathbf{73}$s & $100\%$ & $100\%$ \\ \hline
        MCTS-60 & $10.45$ & $529$s & $83.3\%$ & $87.0\%$ \\ \hline
        PMBS-60 (sim) & $5.03$ & $81$s & $100\%$ & $97.2\%$ \\ \hline
        MCTS-60 (sim) & $10.77$ & $587$s & $76.7\%$ & $96.6\%$ \\ \hline
    \end{tabular}
    \caption{Real robot experiment results on the six most difficult cases. Time budgets are limited to 60 seconds per case.}
    \label{tab:pmbs-real6table}
\end{table}

\textbf{Additional Experimental Details.} Curious readers may find in the online supplementary material additional experimental details including complete, actual execution snapshots of PMBS and MCTS for all 20 cases, as well as the execution snapshots for real robot experiments.

\section{Summary}

In this chapter, we proposed PMBS, a novel parallel Monte Carlo tree search technique with GPU-enabled batched simulations for accelerating long-horizon, episodic robotic planning tasks. Through a series of careful design choices to overcome major parallelization bottlenecks, PMBS achieves an over $30\times$ speedup compared to an optimized serial MCTS implementation while also delivering better solution quality, using identical computing hardware. Real robot experiments show that PMBS directly transfers from simulation to the real physical world to achieve near real-time planning performance in solving complex long-horizon episodic robot planning tasks. 
    
\chapter{Toward Optimal Tabletop Rearrangement with Multiple Manipulation Primitives}\label{chap:remp}
\thispagestyle{myheadings}

\def\prob{\textsc{REMP}\xspace}
\def\hbfs{\textsc{HBFS}\xspace}
\def\pmmr{\textsc{PMMR}\xspace}

\section{Introduction}

Real-world manipulation tasks, e.g., rearranging a messy tabletop or furniture in the house, often require multiple manipulation primitives (e.g., pick-n-place, pushing, toppling, etc.) to accomplish. 
When rearranging small/light objects, e.g., a cell phone on a table or a small chair in a room, it is convenient to do a \emph{pick-n-place}, i.e., to pick up the object, lift it above other objects, move it across the space to above its destination on the table, and then place it. 
On the other hand, for handling large/heavy objects, e.g., a thick book or a heavy couch, \emph{pushing} or \emph{dragging} close to the space's surface is more commonly adopted, executed with added caution. In this case, planning the object's motion trajectory must consider avoiding colliding with other objects more carefully. 
Solving such long-horizon task-and-motion planning tasks efficiently and optimally is highly challenging, as it involves not only an extended horizon but also selecting among multiple types of manipulation primitives at each step, both of which add to the combinatorial explosion of the search space. 
\begin{figure}[ht!]
    \centering
    % \includegraphics[width=\linewidth]{figures/fig1.png}
    % \includesvg[inkscapelatex=false, width = 0.98\linewidth]{figures/fig1.svg}
    % \vspace{0.5mm}
    \begin{overpic}[width=0.99\linewidth]{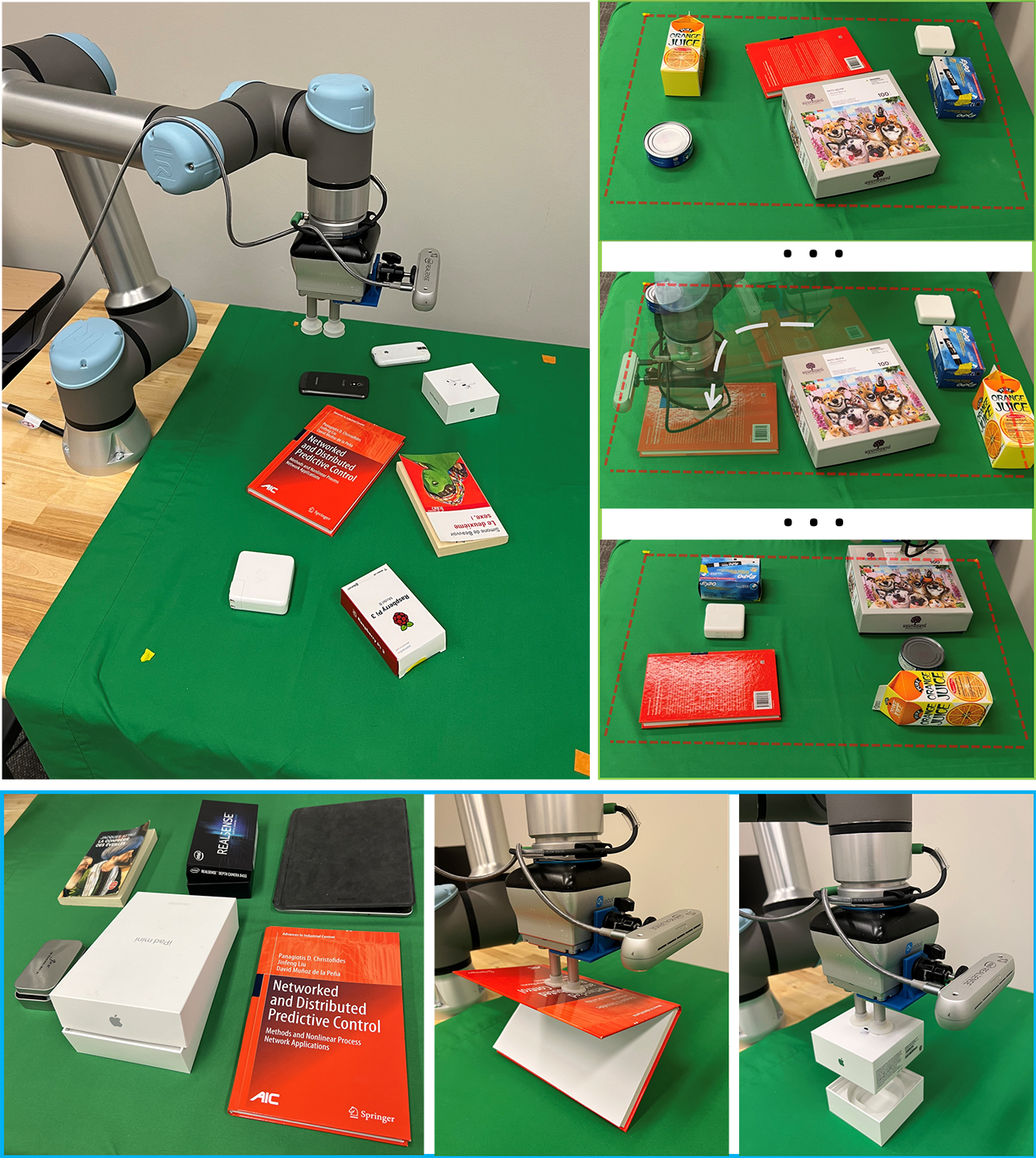}
        \put(1,33.5){\textcolor{white}{{\small (a) Hardware setup}}}
        \put(52,80){\textcolor{white}{{\small (b) Start state}}}
        \put(52,57){\textcolor{white}{\small (c) Push}}
        \put(52,33.5){\textcolor{white}{\small (d) Goal state}}
        \put(0.9,1){\textcolor{white}{\small (e) Objects require push}}
        \put(38,1){\textcolor{white}{{\small (f) Book}}}
        \put(64,1){\textcolor{white}{\small (g) Box}}
    \end{overpic}
    \caption{
    (a) Overview of system setup, a camera is mounted on the end-effector for perception.
    (b)-(d) An example case and an intermediate step in solving it.
    (e) Example objects requiring a \emph{push}. (f) \emph{pick-n-place} may break the book. (g) \emph{pick-n-place} will separate a box, failing to pick it up. 
    \label{fig:remp-system-snapshot}
    }
    % \vspace{-6mm}
\end{figure}
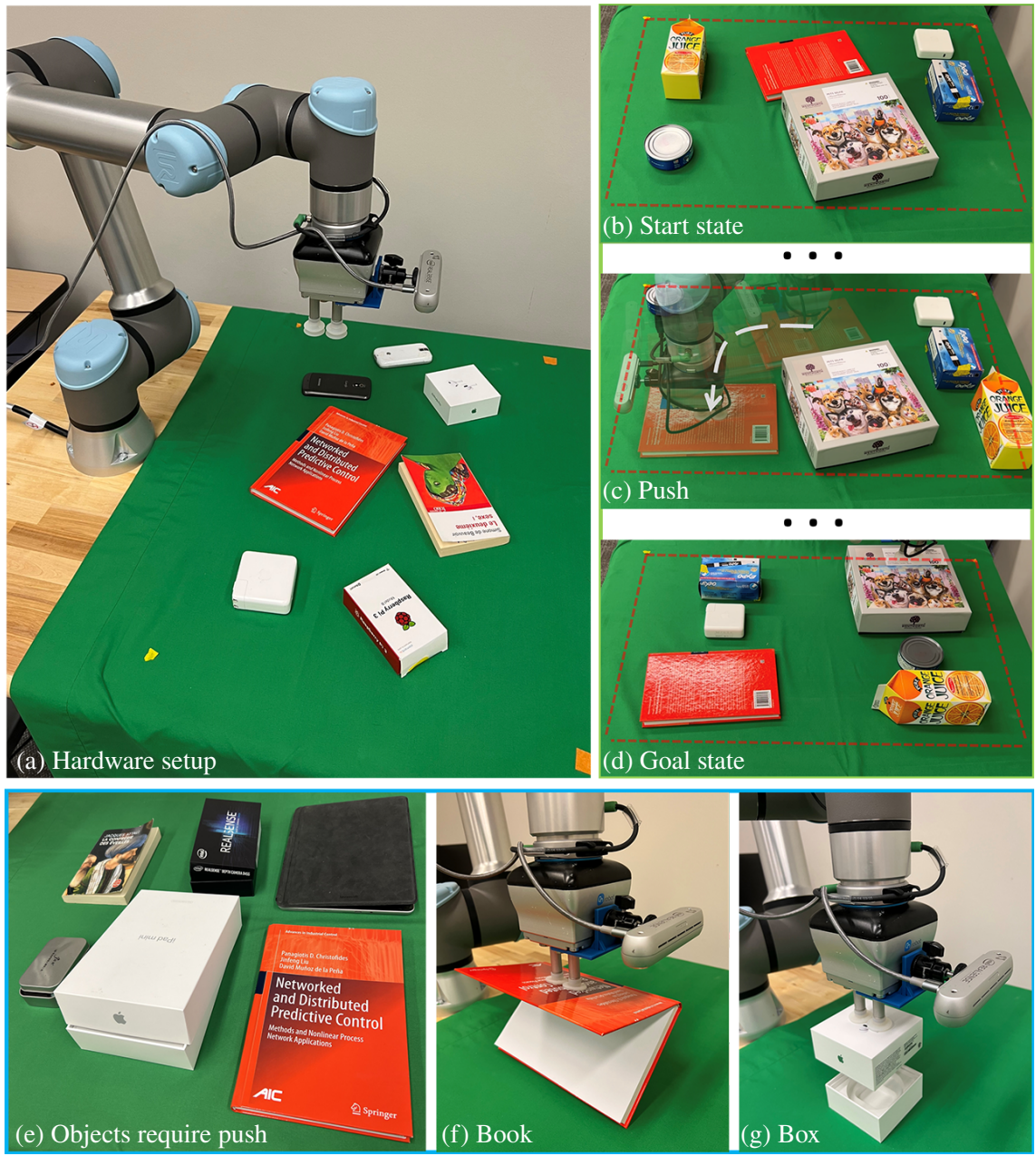

Toward quickly and optimally solving rearrangement tasks using multiple manipulation primitives, we focus on a tabletop setting where both \emph{pick-n-place} and \emph{pushing} are employed to rearrange objects (see \autoref{fig:remp-system-snapshot}). Many objects, such as those shown in \autoref{fig:remp-system-snapshot}(e), cannot be easily picked up and moved around without damaging or disassembling the object. For example, as shown in \autoref{fig:remp-system-snapshot}(f)(g), books and certain boxes cannot be moved around using suction-based pick-n-place manipulation primitive (note that it is also difficult to do pick-n-place using fingered grippers). However, these objects can be effectively rearranged using a \emph{pushing} manipulation primitive in which the suction-based end-effector holds the object on or close to the tabletop and pushes/drags the object around (see \autoref{fig:remp-system-snapshot}(c)). We call the frequently encountered yet largely unaddressed problem \emph{rearrangement with multiple manipulation primitives} (\prob). This study on \prob brings the following contributions: 

\begin{itemize}[leftmargin=4mm]
    \item With the formulation of \prob, we propose a first formal study of solving long-horizon rearrangement tasks utilizing multiple distinctive precision manipulation primitives with the goal of computing an optimized manipulation sequence. 
    Due to its high practical relevance, \prob constitutes an important specialized task and motion planning problem.
    \item We developed two novel algorithms for \prob, the first of which is a fast rule-based solution capable of effectively and quickly solving non-trivial \prob instances. The second, leveraging Monte Carlo tree search (MCTS)~\cite{coulom2006efficient} and parallelism to look further into the planning horizon, delivers a much higher success rate for more challenging tasks, providing higher-quality solutions simultaneously. 
    \item We thoroughly evaluate our methods in simulation and extensive real robot experiments. In particular, our real robot experiments with integrated vision solutions, demonstrate that our algorithms can be readily applied to interact with everyday household objects in real-world scenarios.
\end{itemize}

\section{Problem Formulation}

\subsection{Rearrangement with Multiple Manipulation Primitives}
We now specify the concrete \emph{rearrangement with multiple
manipulation primitives} (\prob) studied in this work. 
Let the workspace be $\mathcal{W}$ a 2D rectangle. 
The robot is provided with a start image (state) $s_s$ and a goal image (state) $s_g$ containing the initial and desired object arrangements.  
The robot must rearrange the objects to match the configurations specified in $s_g$. 
Two manipulation primitives are permitted: \emph{pick-n-place} $\mathcal{A}_{pp}$ and \emph{push} (from top) $\mathcal{A}_{pt}$. 
The robot's objective is to complete the task efficiently in terms of the \emph{execution time}. 
The start and goal states and the objects' transportation should be collision-free, and all objects should remain within the workspace. 
It is assumed that all tasks are feasible, i.e., there is always a viable solution $\mathcal{P} = \{a_1, a_2, ..., a_n\}$ leading from the start state $s_s$ to the goal state $s_g$, where $a \in \{\mathcal{A}_{pp}, \mathcal{A}_{pt}\}$.

A state $s_t$ represents the pose of objects at time $t$.
A pick-n-place action is specified by a pick pose $(x_0, y_0, \theta_0)$ and a place pose $(x_1, y_1, \theta_1)$.
A push action is specified by trajectories $\{(x_0, y_0, \theta_0), ..., (x_n, y_n, \theta_n)\}$, where the robot holds the object against the tabletop at the initial pose, and then pushes it following the waypoints, ending at the final pose. 

One assumption is that the object should be capable of being stably positioned on the workspace, as all primitives are considered to be quasi-static.

\subsection{Monte Carlo Tree Search}

The Monte Carlo tree search (MCTS) algorithm has broad applications. 
It is prevalent in turn-based tasks such as the game of Go~\cite{silver2016mastering}, but its usage extends beyond such contexts. 
MCTS plays a crucial role in solving rearrangement tasks~\cite{labbe2020monte, huang2022parallel, gao2023effectively}.
As an \emph{anytime} tree search algorithm, MCTS is designed to run for a fixed amount of time, each consisting of four stages: selection, expansion, simulation, and backpropagation. 
Fundamentally, MCTS preferentially exploits nodes that yield superior outcomes. 
To strike a balance between exploration and exploitation in the selection phase, an \emph{upper confidence bound} (UCB)~\cite{auer2002finite} formula is utilized (see \autoref{eq:pmbs-ucb2}), where $n$ is the parent node of $n'$, and $Q(n')$ is the total reward $n'$ received after $N(n')$ visits.

\section{Methodology} \label{sec:remp-methods}

This work addresses the primary challenge of synergistic integration of pick-n-place and push actions. 
Because planning a push involves collision-free path planning in $SE(2)$, which is time-consuming, it poses a significant challenge if many push actions are explored. 
MCTS offers a solution capable of elegantly negotiating between the two disparate actions while maintaining optimality, given ample planning time.

\subsection{Action Space Design}
Planning requires searching through candidate manipulation actions, which must first be \emph{sampled}. 
Action space design refers to action sampling, which plays a critical role in dictating the expansion of the tree search because there are an uncountably infinite number of possible manipulation actions. 
In sampling pick-n-place actions, we must ensure the place pose is collision-free. 
The same applies to a push action's final pose (though, in addition, the entire trajectory connecting all waypoints for a push action must be collision-free).
Four criteria are applied to sample the place/final poses for pick-n-place/push actions at the current state $s_t$: 
\begin{enumerate}
  \item \textbf{Random}. A naive approach randomly samples collision-free place/final poses for pick-n-place/push actions.
  \item \textbf{Around Current and Goal}. 
  Random sampling is not always efficient. For object $o_i$, favoring regions around the current and goal poses of $o_i$ in $s_g$ can be helpful.
  \item \textbf{Grid}. Additionally, we adopt a grid-based sampling strategy. By modeling an object as a 2D polygon, we encapsulate it within a rotated bounding box. This allows us to generate a tiled representation in the workspace, denoted as $\mathcal{W}$, resembling a grid structure. This method proves advantageous for covering boundary areas, which can be hard to sample through random methods.
  \item \textbf{Direct to Goal}. If $o_i$ can be directly placed at its goal, this pose will be prioritized over the above three samplings in the search process. 
\end{enumerate}

\begin{figure}[ht!]
    \centering
    \makebox[\linewidth][c]{%
      \fbox{%
        \begin{overpic}[width=0.6\linewidth]{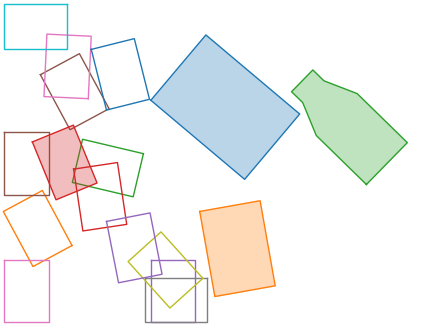}
            \put(52,49){\small 0}
            \put(80,45){\small 1}
            \put(54,16){\small 2}
            \put(13,36){\small 3}
        \end{overpic}%
      }%
    }
    \caption{Consider action sampling for labeled $3$ to be manipulated using push (there are a total of four objects). The absence of sampled actions in the right region is attributed to obstructions posed by objects $0$, $1$, and $2$, preventing the movement of object $3$ to that area.
    \label{fig:remp-sample-actions}
    }
\end{figure}
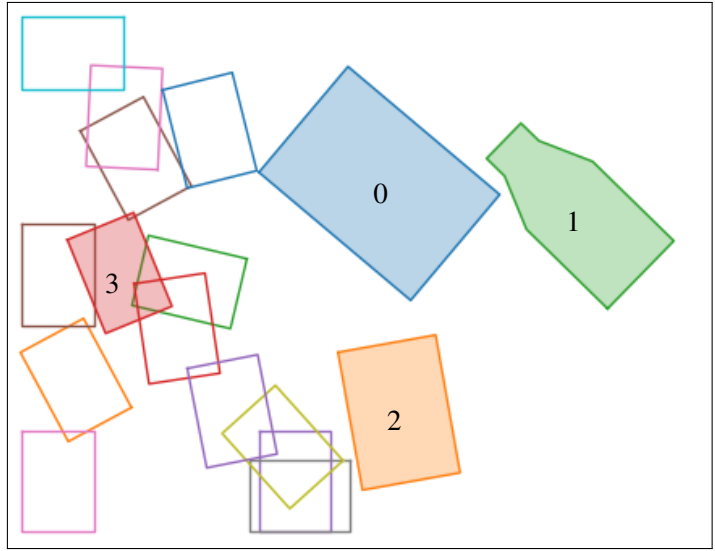

Once the place/final poses have been sampled, a $\mathcal{A}_{pp}$ sample is obtained.
However, $\mathcal{A}_{pt}$ requires an additional step - generating a trajectory from the current pose to the place pose for $o_i$.
We employ RRT-connect~\cite{844730} during the tree search and LazyRRT~\cite{bohlin2001randomized} for robot execution to produce such collision-free trajectories within a set time limit. 
An example of sampling the final pose for a push action is shown in \autoref{fig:remp-sample-actions}.
The criteria for sampling are crucial in solving the problem. Relying solely on standard uniform sampling often proves inefficient, particularly when sampling around boundaries and certain edge cases. While one might consider increasing the sample size to cover these edge cases, this inadvertently leads to many redundant actions that are time-consuming to process.

\subsection{Hierarchical Best-First Search}
The first algorithm we designed for \prob is a (greedy) best-first search algorithm called \emph{hierarchical best-first search} (\hbfs). \hbfs is outlined in \autoref{alg:remp-hbf} and operates according to the following sequence of steps: 
\begin{itemize}
    \item (Lines 3-4) When objects can be directly moved to their goal poses, an action cost is computed for each. The action yielding the smallest cost is then applied.
    \item (Lines 5-9) For each object $o_i$, \hbfs identifies which objects occupy $o_i$'s goal and attempts to displace these obstructing objects in the direction of their respective goals. If no actions are feasible in this direction, a random action is sampled. Again, the action associated with the smallest cost is selected and implemented.
    \item (Line 10) If the above steps do not yield a viable action, an action is randomly selected for execution.
\end{itemize}
The above three phases of \hbfs may best be viewed as a three-level hierarchical search. To boost its performance and solution optimality, \hbfs is implemented by leveraging multi-core capabilities of modern CPUs. This is realized by executing multiple \hbfs in parallel and choosing the best action among the returned solutions.

\begin{algorithm}
    \begin{small}
    \DontPrintSemicolon
    \SetKwFunction{FHBF}{HBFS}
    \Fn{\FHBF{$s_s, s_g$}}{
        $s \gets s_s$, $A \gets \varnothing$ \;
        \lFor{$o_i$ \text{in} $s$}{
            $A \gets A \cup \{\text{move } o_i \text{ to its goal}\}$
        }
        \lIf{$|A| > 0$}{
            \Return the lowest cost action from $A$
        }

        \For{$o_i$ \text{in} $s$}{
            $obs \gets \text{objects occupy the goal pose of } o_i$ \;
            \For{$o_j$ \text{in} $obs$}{
                $A \gets A \cup \{\text{move } o_j \text{ towards its goal, otherwise at random}\}$
            }
        }
        \lIf{$|A| > 0$}{
            \Return the lowest cost action from $A$
        }

        \Return a randomly sampled action
    }
    \caption{Hierarchical Best-First Search (\hbfs)}\label{alg:remp-hbf}
    \end{small}
\end{algorithm}

\subsection{Speeding up MCTS with Parallelism}
Standard MCTS is more straightforward to implement, but it does not fully utilize multi-core processing capabilities of modern hardware. 
We introduce parallelism to the expansion and simulation stages of MCTS, leveraging tree parallelization techniques~\cite{chaslot2008parallel}.
The application of parallelism allows for decoupling the select and expand stages from the simulation stage in an MCTS iteration. 
The decoupling allows multiple MCTS iterations to be carried out simultaneously, limited only by the number of CPU cores. 
The standard UCB formula used in MCTS is updated as \autoref{eq:pmbs-ucb2},
where the idea of virtual loss~\cite{chaslot2008parallel} is applied by adding one extra virtual visit counts $\hat{N}$ that indicates a node has been selected but not yet simulated and backpropagated.
Since the simulation and subsequent backpropagation stages are not yet completed, the tree search algorithm must be notified to update the $Q$ and $N$ of the node. This adjustment minimizes the likelihood of revisiting the node in the next iteration, implementing a conservative approach in anticipation of a potentially poor reward. Once the simulation stage has concluded, $Q$ is updated in backpropagation with returned reward from simulation result, $N$ increments and $\hat{N}$ decrements.

\subsection{Adapting MCTS for \prob}
Given \prob's extremely large search space due to push actions' trajectory planning requirements, modifications are introduced to best apply MCTS to \prob. 

\textbf{Action Space Bias}. The action space used in the simulation stage is a subset of that used in the expansion stage. 
Given that one iteration of the simulation stage constitutes a coarse estimation of action and state, reducing the number of actions leads to a larger number of total iterations.

\textbf{Biased Simulation}. A straightforward implementation of the simulation stage in MCTS is a random policy that indiscriminately selects an action for execution, ultimately obtaining a reward at the terminal state reached. 
In our implementation, we adopt a heuristic to guide the action selection in the simulation stage towards the ultimate goal of a given object. Specifically, with a probability of $\theta_{sim}$ (dynamically changed based on depth of the search), a random action is selected; otherwise, an attempt is made to select an action that will move an object towards its goal pose. 
This introduces a bias in the simulation stage, which offsets the drawback of limited iterations due to the time-consuming nature of motion planning and collision checking.
We note that we did not adopt recent advancements in MCTS for long-horizon planning that injects a data-driven element to partially learn the reward, e.g., a neural network can be trained to evaluate the quality of an action-state pair~\cite{song2020multi, bai2022hierarchical, huang2022interleaving}.

\textbf{Reward Shaping}. We structure the reward to favor the goal state but without introducing undue bias. The reward function plays a critical role as it steers the tree search and is composed of three components.
Firstly, if the task is accomplished, with all objects placed at their goal poses, a reward of $R_g$ is awarded.
Secondly, if an object $o_i$ is located at its goal pose, a reward $r_o$ is given. The cumulative reward from all objects, denoted as $R_o$, is computed as $R_o = \sum_{i} r_{o_i}$. 
Lastly, the reward structure also takes into account the cost associated with the movement of objects.
For the pick-n-place action ($\mathcal{A}_{pp}$), the cost corresponds to the Euclidean distance between the pick-and-place poses, with an additional fixed cost factored in. 
For the push action ($\mathcal{A}_{pt}$), the cost is determined by the Euclidean distance of the path, also supplemented by a fixed cost.
Additionally, a base reward is computed $R_b = R_o(s_0)$ from initial state $s_0$, which is used to normalize the final reward during the search.
For each iteration, a reward is returned by the simulation stage and is updated during the backpropagation stage.
\[
    R_i= 
\begin{cases}
    \max(0, R_g - \textit{cost} - R_b),& \text{if } s_i \text{ is the goal state}\\
    \max(0, R_o(s_i) - \textit{cost} - R_b),              & \text{otherwise}
\end{cases}
\]

Traditionally, $Q$ retains the average reward values derived from simulation results, providing a robust estimation for action over millions of iterations. 
However, in our case, we aim to maintain the planning time within reasonable limits. 
Therefore, we introduced a priority queue to store simulation results, which serves as the $Q$ value in the algorithm.
For the purpose of calculation in~\autoref{eq:pmbs-ucb2}, we only preserve the top $k$ rewards, similarly in the \autoref{chap:more}.
There is a possibility that during a simulation, a subsequent action may transition the state to one with lower rewards, thereby negating the benefits of a preceding beneficial action within that simulation. 
To limit the search time, we choose to return the maximum reward encountered at the intermediate steps during the simulation instead of the final reward. 
Therefore, the returned reward from a simulation is given by $$\max(\beta \cdot \max_{i \in m-1}(R_i), R_m) \cdot \gamma^m,$$ where $\beta\in (0,1]$ is a scaling parameter and $m$ is the total steps used in the simulation.
$\gamma$ is the discount factor, encouraging the problem to be solved in the early stage if possible.

Due to limited space, we omit the pseudo-code of the parallel MCTS algorithm but note that all details for reproducing the algorithm have been fully specified. We call the resulting algorithm \emph{parallel Monte Carlo tree search for multi-primitive rearrangement} or \pmmr.

\section{Experimental Evaluation} 

We evaluated \hbfs and \pmmr methods for \prob in simulated environments and on a real robot. 
Regardless of whether it is a simulation or a real robot experiment, both algorithms perform \emph{percept-plan-act} loops until the task is solved or the budgeted time or maximum number of actions is exhausted. 
All experiments were conducted on an Intel i9-10900K (10 CPU cores) desktop PC and implemented in Python.
As a note, limited testing shows that using Intel i9-13900K (24 CPU cores) reduces the planning time by roughly half, demonstrating the effectiveness and scalability of employing parallelism (code in Python). 

\begin{figure}[h]
    \centering
    \includegraphics[width = 0.99\linewidth]{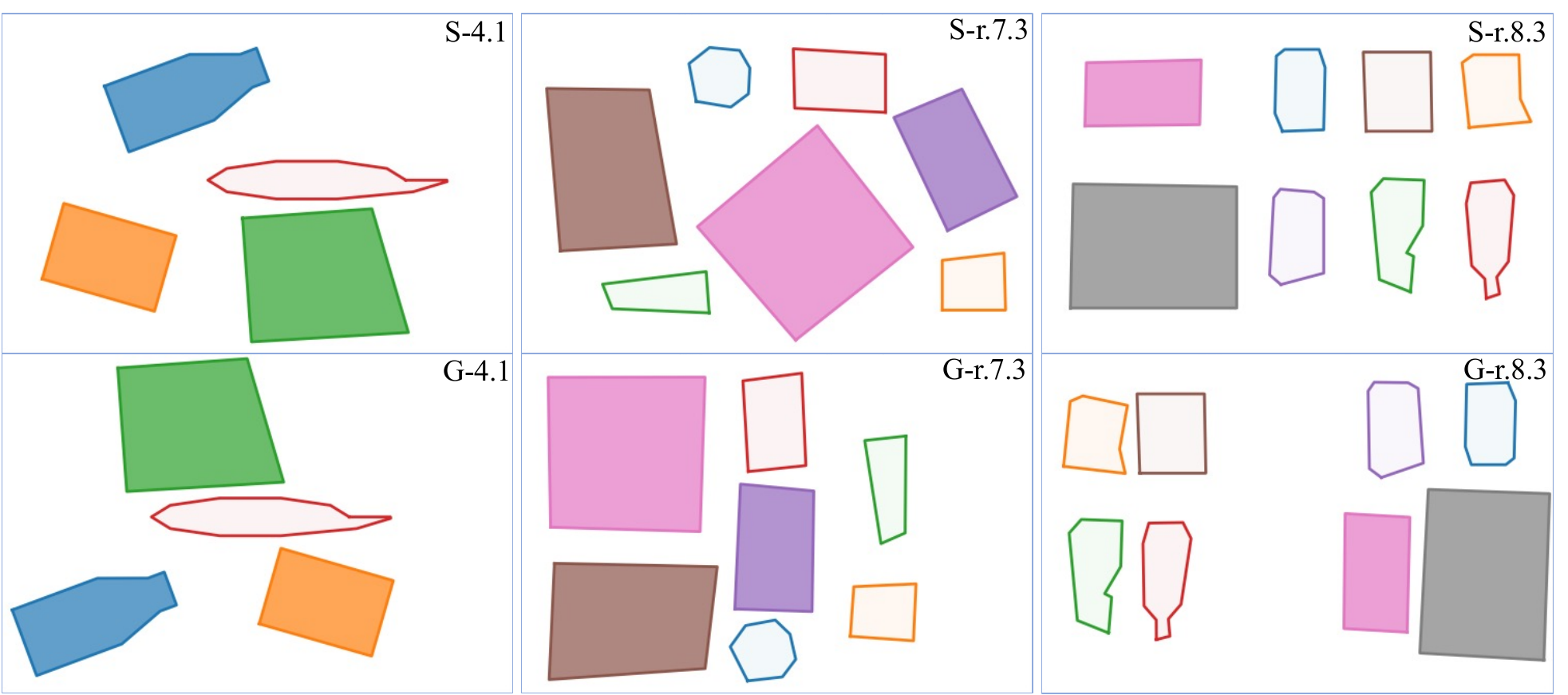}
    \caption{Example cases. The top row shows the start states and the bottom goal states. Lightly shaded objects can be pick-n-placed; heavily shaded objects must be manipulated using push. %
    Cases 4.1, r.7.3, and r.8.3 are evaluated and presented in~\autoref{fig:remp-sim}. Objects are distinguished by color.
    \label{fig:remp-cases}
    }
\end{figure}

\subsection{Simulation Studies}
Simulations are conducted in PyBullet~\cite{coumans2016pybullet}. 
A real robot setup, consisting of a Universal Robot UR5e $+$ OnRobot VGC-10 vacuum gripper, is replicated. 
The robot operates under end-effector position control; the workspace measures $0.78 \times 0.52 \si[per-mode=symbol]{\meter\squared}$.
$25$ feasible scenarios are created where all objects are confined within the workspace. Object sizes, shapes, and poses are randomly determined in each scenario (see \autoref{fig:remp-cases} for some examples). 
Cases that are trivial to solve (e.g., objects that happen to be mostly small) are filtered. 
The number of objects ranges from four to eight; five distinct cases are generated for each specified number of objects.

\begin{figure*}[ht!]
\vspace{1mm}
    \centering
    \begin{minipage}{\linewidth}
        \includegraphics[width = .99\linewidth]{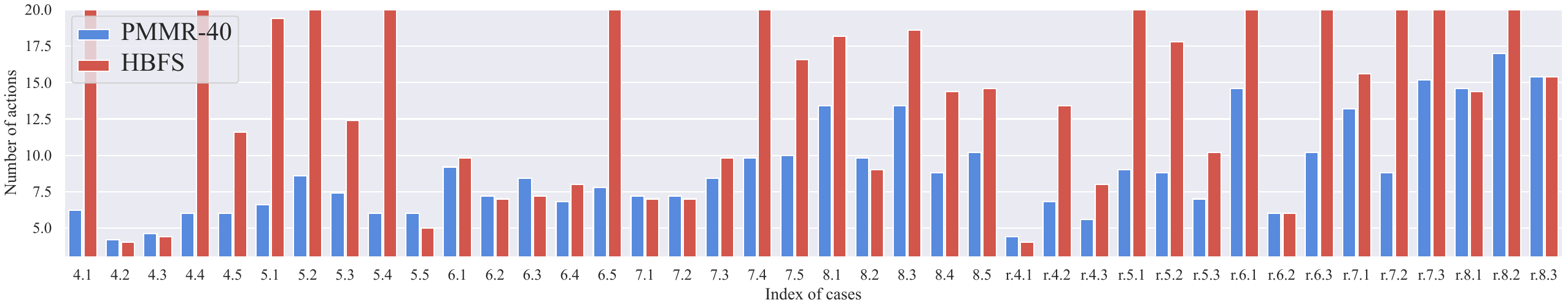}
    \end{minipage}
    \vspace{1mm}
    \begin{minipage}{\linewidth}
        \includegraphics[width = .99\linewidth]{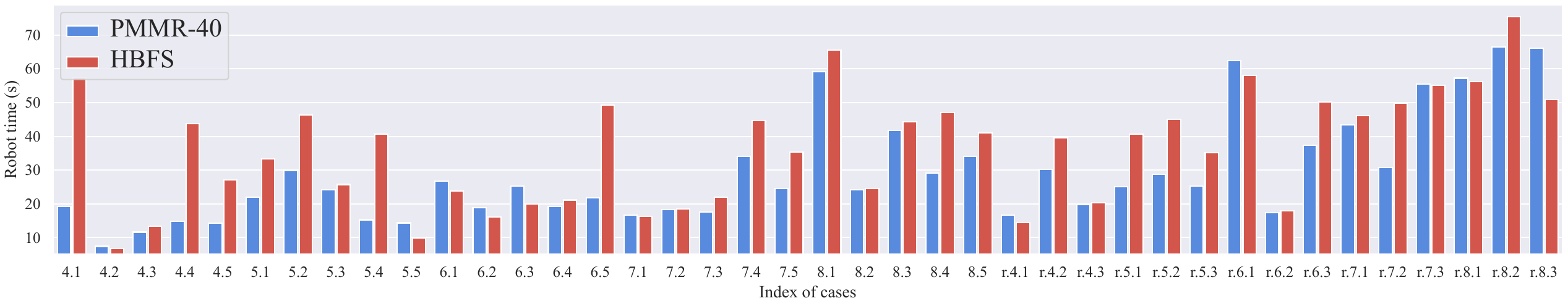}
    \end{minipage}
    \caption{\label{fig:remp-sim}
        As an expanded illustration of \autoref{tab:remp-simtable}, the upper plot lists the number of actions the robot executes to resolve individual cases. 
        The lower plot lists the robot's execution times in solving the individual cases following the computed plan.
        For the labels on the horizontal axis, the first digit indicates the number of objects contained within each case, while the second digit represents the index of the cases. Cases beginning with the prefix 'r' are the ones that are constructed for and executed by the real-robot setup.
    }
\end{figure*}

In evaluating \pmmr, each \emph{percept-plan-act} loop runs for a predetermined duration to identify the best next action until the problem is resolved or the maximum number of actions has been exhausted. %(which is case-dependent) 
If the latter occurs, it is treated as a failed case. 
%
%Moreover, the planning and robot execution times are capped based on the maximum steps. 
We denote the corresponding \pmmr method as \pmmr-X, where X is the maximum number of seconds allowed in a single iteration of the loop.
We settled on \pmmr-40 as the main \pmmr method used in the evaluation. 
In both simulation and real robot experiments, for \pmmr-40, we keep the top $k=100$ for $Q$ value, $c=1.5$ in \autoref{eq:pmbs-ucb2}.
The maximum depth of the MCTS tree $D$ is based on the number of objects $\mathcal{N}$: $D=2\mathcal{N} + 2$.
$\theta_{sim}$ is based on the depth $d$ of the node $\theta_{sim} = \max(-0.106 + 0.231d - 0.013d^2, 0.2)$.
These numbers are handpicked, representing that as the tree goes deeper, the probability of selecting a random action should be increased.
$r_o=0.7$ for object can be operated by $\mathcal{A}_{pp}$, the $R_g=2r_0 \mathcal{N}$. For an object that can be operated by $\mathcal{A}_{pt}$, the reward is given to $1.1r_o$.
We set $\beta=0.5$ and $\gamma=0.9$ to scale the reward.

\begin{table}[h]
    \centering
    \scalebox{0.99}{\begin{tabular}{c|c|c|c|c}
        & Robot Time & Completion & Num. of Actions & Plan Time     \\ \hline
        \pmmr-40 & $29.17$s & $98.00\%$ & $8.90$ & $264.99$s \\ \hline
        \hbfs & $36.22$s & $54.50\%$ & $13.72$ & $30.04$s \\ \hline
    \end{tabular}}
    % \vspace{2mm}    
    \caption{Summary of simulation results (25 cases) and real-robot experiments (15 cases) for \hbfs and \pmmr-40.}
    \label{tab:remp-simtable}
\end{table}

Individual experiment results for all cases are shown in \autoref{fig:remp-sim} (which also includes cases used for real-robot experiments, to be detailed in \autoref{sec:remp-real}).
Detailed experiment results are presented in \autoref{tab:remp-simtable}.
Here, \emph{robot time} refers to the cumulative time required for the robot to execute all actions, while \emph{completion} refers to the success rate. The number of actions quantifies the execution of atomic actions, represented by $\mathcal{A}_{pp}, \mathcal{A}_{pt}$. The \emph{plan time} is the total planning time.
Each case underwent five independent trials.

Failure often happens because the case requires more than 15 actions to solve (even for humans). A failure may be due to sampled actions not containing a solution or the search not being deep enough. Sometimes, the algorithm can recover from an early bad choice, but not always since the number of iterations is capped. 
We observe that, while \hbfs runs relatively fast in comparison to \pmmr-40, it frequently fails ($55\%$ success vs. $98\%$ for \pmmr-40) and uses many more actions (13.7 vs. 8.9 for \pmmr-40). 
Visually, as can be seen in the accompanying video, the actions generated by \pmmr-40 are much more human-like than those by \hbfs (the same holds for real robot experiments).

\subsection{Ablation Studies}
We investigated the impact of time budgets, the depth of tree search, and the base reward $R_b$ on solving \prob. The time budget is critical; an extended search duration tends to yield better results. However, it is necessary to balance planning time and solution quality. An insufficient search might sample highly suboptimal paths, leading to locally optimal actions. As depicted in \autoref{fig:remp-simablation} and \autoref{tab:remp-ablation-table} shows the correlation between planning time and solution quality, leading us to select \pmmr-40 for our main evaluation.

A shallow MCTS (\pmmr-40 ($D=3$), max MCTS tree depth of $3$) was included specifically to compare with \hbfs, which has three ``depth levels'' per iteration. 

In terms of reward design, as detailed in \autoref{sec:remp-methods}, we introduced a base reward $R_b$, which serves to normalize the reward to 0 at root. Without this adjustment, the tree search might commence with a non-zero reward signal, where a disadvantageous branch may still return a reward, causing the search to frequently explore such branches. By comparing \pmmr-40 (no-$R_b$) in \autoref{tab:remp-ablation-table} with \pmmr-40 in \autoref{tab:remp-simtable}, we observe that the introduction of $R_b$ aids the tree search.

\begin{table}[h]
\vspace{1mm}
    \centering
    \scalebox{0.94}{\begin{tabular}{c|c|c|c|c}
        & Robot Time & Completion & Num. of Actions & Plan Time     \\ \hline
        \pmmr-10 & $35.60$s & $94.00\%$ & $10.51$ & $103.61$s \\ \hline
        \pmmr-20 & $32.46$s & $96.00\%$ & $9.86$ & $155.68$s \\ \hline
        \pmmr-60 & $27.93$s & $99.50\%$ & $8.53$ & $358.44$s \\ \hline
        \pmmr-40 ($D=3$)& $57.83$s & $32.50\%$ & $16.62$ & $641.76$s \\ \hline
        \pmmr-40 (no-$R_b$)& $32.92$s & $94.00\%$ & $9.83$ & $270.65$s \\ \hline
    \end{tabular}}
    \caption{Ablation study results (averaged over 40 cases), for comparison with \autoref{tab:remp-simtable}.}
    \label{tab:remp-ablation-table}
\end{table}

\begin{figure}[h]
    \centering
    \includegraphics[width = \linewidth]{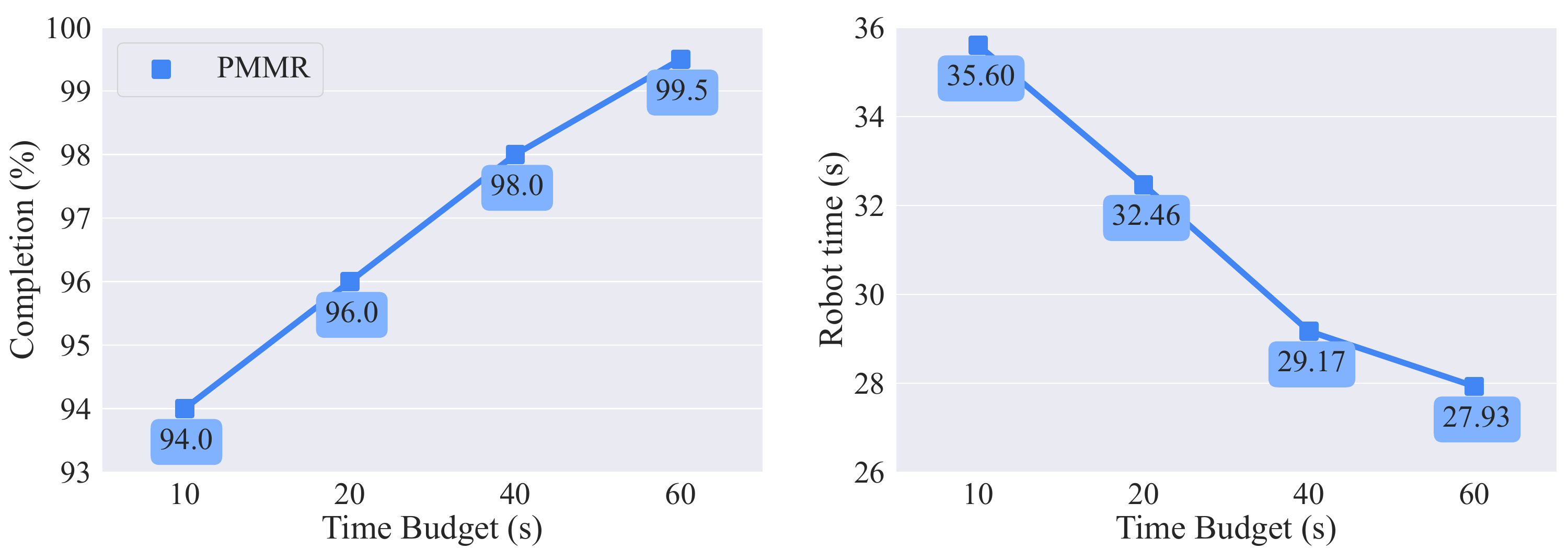}
    \caption{\label{fig:remp-simablation}
        \pmmr is evaluated with different time budgets. The reported values are averaged over 40 cases.
    } 
\end{figure}

\subsection{Real Robot Experiments}\label{sec:remp-real}

In simulation, we emulate the vacuum function by attaching the object to the end-effector using an extra link via a fixed joint. 
We employ two vacuum cups to provide sufficient suction power to ensure a robust connection in the real-world setup. 
The point of suction on the object is taken as its center, assuming this central area is flat.
Similar to simulation studies, the number of objects ranges from four to eight, and three distinct cases are generated for each number of objects.

\begin{figure}[h]
    \centering
    \includegraphics[width=0.97\linewidth]{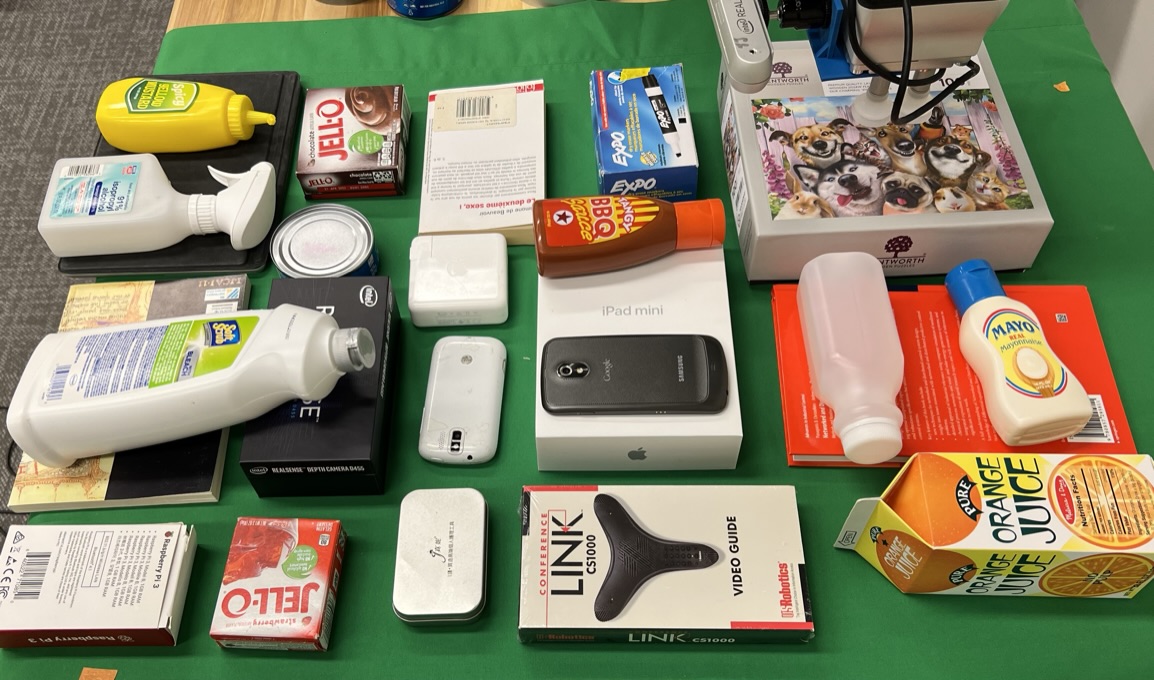}
    \caption{The full set of objects used in our real-robot experiments. 
    \label{fig:remp-all-objects}
    }
\end{figure}

A RealSense D455 camera is affixed to the robot's wrist, capturing the scene from a top-down perspective, and an orthogonal view is rendered from the point cloud.
We employ the Segment Anything Model~\cite{kirillov2023segment} to extract masks of the objects present on the table. 
Subsequently, OpenCV~\cite{opencv_library} is applied to determine the contours and approximate them into polygons, which are then used for planning.
An additional step determines each object's $SE(2)$ poses. 
For each case, the experiment is repeated at least three times.

Due to the small sim-to-real gap, we let the algorithm plan the entire sequence of manipulation actions at the beginning, which generally works well. 
If no solution is found using $20$ actions, we mark it as a failure; otherwise, the robot executes the planned actions. 
Robot time is not recorded for failure cases, hence its absence in \autoref{tab:remp-realtable}. 
For completeness, in cases where both \pmmr and \hbfs succeed at least once, \hbfs averages $15.15$ actions and \pmmr $9.56$, with robot (execution) times of $96.62$ seconds and $95.75$ seconds, respectively (but note that \hbfs fails much more frequently).
Results from individual benchmarks across 15 cases are presented in \autoref{fig:remp-realnum}. Each case was subjected to three independent trials.
In real-robot experiments, the cases were intentionally designed to be challenging.
A greedy action may exacerbate the problem, making it even more difficult to resolve. 
Consequently, the completion rate of \hbfs is significantly reduced.

\begin{table}[h]
    \centering
    \scalebox{0.96}{\begin{tabular}{c|c|c|c|c}
        & Robot Time & Completion & Num. of Actions & Plan Time     \\ \hline
        \pmmr-40 & $95.75$s* & $96.44\%$ & $9.56$ & $292.02$s \\ \hline
        \hbfs & $96.62$s* & $38.33\%$ & $15.15$ & $29.05$s \\ \hline
        \pmmr-40 (Sim) & $-$ & $94.67\%$ & $10.44$ & $306.41$s \\ \hline
        \hbfs (Sim) & $-$ & $45.33\%$ & $14.99$ & $22.18$s \\ \hline
    \end{tabular}}
    \caption{Experiment results of real robot trials across 15 cases, with time budgets constrained to a maximum of 40 seconds for a single MCTS run. The robot time is only considered in cases where both methods succeed at least once. Additionally, benchmarks from simulations covering 15 cases are included for sim-to-real gap comparisons. The robot time for \pmmr-40 and \hbfs, denoted with an asterisk, is recorded only for successful cases.}
    \label{tab:remp-realtable}
\end{table}

\begin{figure}[ht!]
    \centering
    \includegraphics[width =\linewidth]{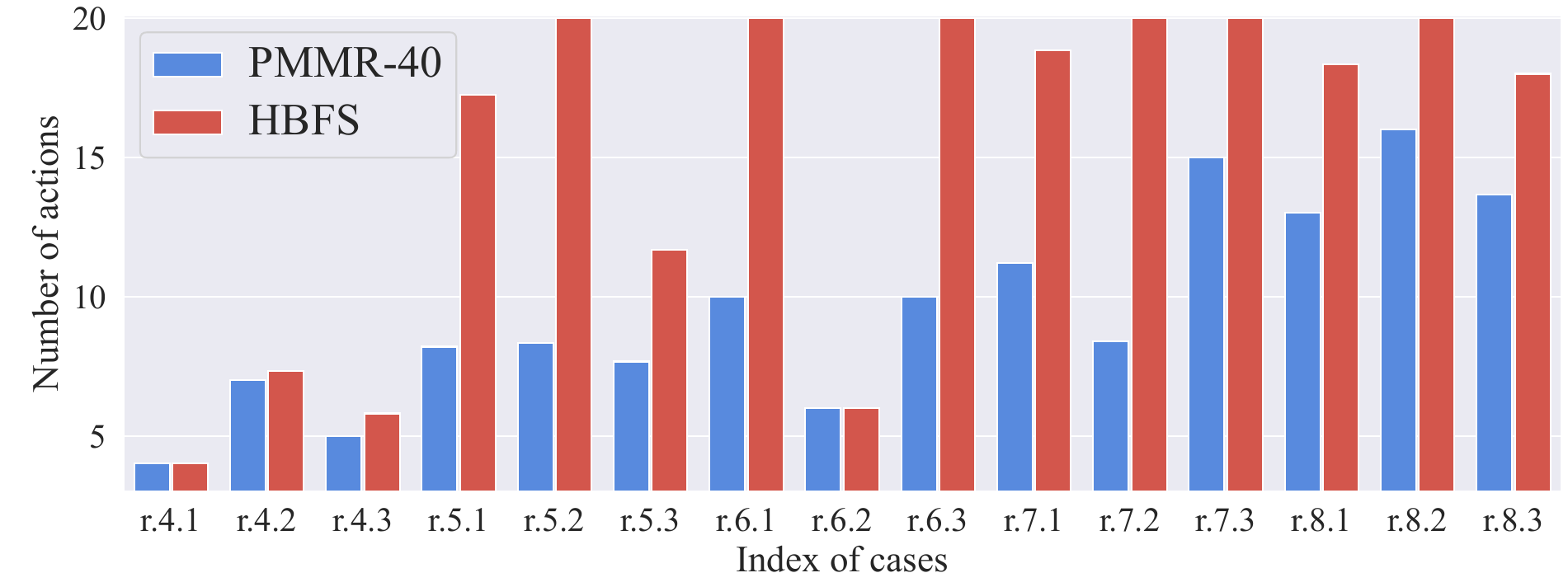}
    \caption{\label{fig:remp-realnum}
        As an expanded illustration of \autoref{tab:remp-realtable}, this plot illustrates the number of actions the robot executes to resolve individual cases. 
    } 
\end{figure}

\section{Summary}
We make the observation that humans frequently solve manipulation challenges using multiple types of manipulation actions. In contrast, there has been relatively limited research tackling planning high-quality resolutions for long-horizon manipulation tasks exploring the synergy of multiple manipulation actions. 
Inspired by how humans solve everyday manipulation tasks, in this paper, we proposed and studied the \emph{Rearrangement with Multiple Manipulation Primitives} (\prob) problem. 
To optimally solve \prob, we developed two effective methods, \hbfs and \pmmr. \pmmr is especially adept at solving difficult \prob instances with high success rates and producing high-quality solution sequences, capabilities confirmed through thorough simulation and real-robot experiments that included full percept-plan-act loops. 

    \chapter{Conclusion}\label{chap:conclusion}
\thispagestyle{myheadings}

In this dissertation, we have explored a spectrum of approaches for robotic manipulation in cluttered and long-horizon scenarios, with an emphasis on efficiently combining learned predictive models, search-based planning, and parallelization techniques. A central theme across the presented work is the pursuit of accurate interaction prediction and high-quality planning under uncertainty, grounded in both simulation and real-world experimentation.

With the introduction of Deep Interaction Prediction Network (DIPN) for push-and-grasp challenges, we established the importance of generating clear and reliable intermediate predictions that can be effectively used by downstream networks such as the Grasp Network (GN). This integrated method (DIPN+GN) exhibits strong generalization capabilities, demonstrates excellent sample efficiency, and outperforms prior state-of-the-art learning-based approaches. Notably, the methods train in a self-supervised manner, require no manual labeling or human input, and are robust to variations in object size, shape, color, and friction.

Building on these learned predictive models, the Visual Foresight Trees (VFT) framework introduces a synergy between DIPN and Monte Carlo Tree Search (MCTS) for long-horizon planning in object retrieval tasks from dense clutter. VFT has been shown to generate high-quality multi-step plans, though its performance is constrained by the substantial computational overhead. This drawback can be mitigated by parallelization—an approach that has been partially addressed by developing parallel MCTS strategies and will be essential for achieving real-time performance in robotic applications.

In parallel, we explored techniques for model-based simulation, including MORE (a simulator-driven approach), which uses a search-and-learn philosophy. Despite showing promising results, MORE and similar planning algorithms require either explicit object models or learned surrogates to simulate push outcomes. These requirements impose potential limitations on generalization to novel objects and introduce additional uncertainties when learned components are used in place of a physics engine.

To address the computational bottleneck inherent in MCTS for robotic tasks, we introduced PMBS, a novel parallel MCTS method that leverages GPU-enabled batched simulations, achieving an over $30\times$ speedup relative to a strong serial MCTS baseline. Real-world robot experiments confirm that PMBS transfers effectively from simulation to physical systems, enabling near real-time performance in complex long-horizon planning.

Finally, we proposed the rearrangement with multiple manipulation primitives (REMP) framework, inspired by how humans tackle daily manipulation tasks using diverse actions. Our methods, HBFS and PMMR, facilitate planning that encompasses multiple action types (e.g., pushing, grasping) and achieve high success rates with high-quality solution sequences in both simulation and real-robot trials.

Across all these methods, a unifying message is the feasibility and effectiveness of integrating learning, search, and parallelization to solve challenging robotic manipulation tasks. Nonetheless, data efficiency, robustness to model inaccuracies, and computational scalability remain ongoing concerns that warrant further attention.

In future research, we will (1) extend the range of manipulation actions to include non-horizontal pushes and non-vertical grasps (arbitrary 6D end-effector poses), enabling more versatile rearrangement; (2) further optimize parallelization to achieve near real-time planning, focusing on multi-threaded or GPU-accelerated strategies; and (3) pursue integrated learning-based frameworks for rollout policies and reward estimation, aiming to reduce reliance on explicit simulation and improve overall efficiency and robustness. These directions promise to enhance the adaptability and performance of robotic systems in increasingly complex, real-world scenarios.

\begin{theappendices}
    \chapter{Chapter 6 - PMBS Supplementary}
    \thispagestyle{myheadings}
    
    \subsection{Grasp Classifier Implementation Details}
For our implementation of the grasp classifier (GC), we used Isaac Gym to collect grasp training data.
Random objects are first sampled on the workspace, and then we discretize the workspace into a grid, where each point is the $(x,y)$ of a grasp action $a^g$.
We also discretize rotation into $K$ angles uniformly.
All robots in simulator will pick one grasp action and check the distance between two fingers as a signal of successful grasping.
For each depth image, it is associated with hundreds of grasps. If the target can be grasped in at least $n$ attempts, then the label is 1, and 0 otherwise ($n$ = 5).
We used two days of generating 20000 training data (a depth image focus centered on the target object and a label of can it be grasped) without human annotation.
It is evaluated on test data with $93.45\%$ accuracy if the $R_c^{*}$ equals 0.7.
The batch size is $256$, learning rate is $0.1$, epochs is $90$, momentum is $0.9$.
We have successfully reduced the grasp evaluation time from $0.26$ to $0.003$ per image, making the parallel MCTS possible.

\subsection{Grasp Network Implementation Details}
For deciding whether to perform further push actions or to make an attempt to retrieve the target object, we resort to a \emph{grasp network} (GN) that is fast and amenable to parallelization.
GN is based on \emph{fully convolutional networks} (FCNs)~\cite{zeng2018learning, huang2021dipn} and customized to estimate the grasp probability for the target object~\cite{huang2021visual, huang2022interleaving}.
It takes an RGB-D image $o_t$ as input and outputs dense pixel-wise values $P(o_t)\in [0,1]^{H\times W\times K}$. 
$H$ and $W$ are the height and width of the $o_t$.
To account the gripper orientation, we 
discretize $\theta$ of $a^g$ into $K=16$ angles, in multiples of ($22.5^{\circ}$), so $o_t$ has been rotated $K$ times.
GN presented in~\cite{huang2021dipn} is trained to estimate the grasp success rate for all objects.
Further, binary mask of target object $M$ is imposed using Mask R-CNN~\cite{he2017mask}, to truncate the values as $P_m(o_t)=P(o_t)\cap M(o_t)$.
In proposed system, we only interested the highest grasp probability.
If $\max_{h,w,k} P_m(o_t)$ is greater than the preset threshold $P^*$ (it is $0.75$ in our case), then, the robot should grasp at location $h, w$ with $k$ as orientation of gripper.
The backbone of GN is ResNet-18 FPN~\cite{he2016deep,DBLP:journals/corr/LinDGHHB16}, with convolution layers and bilinear-upsampling layers as described in~\cite{huang2021visual, huang2022interleaving}.
We used the pre-trained Grasp Network from~\cite{huang2022interleaving}.
The Grasp Network is a plug-in component for the system, can be replaced by other advanced methods as if a grasp probability and grasp action can be provided.

\subsection{Real-to-Sim-to-Real Comparison}

We solve the task using a physics simulator. While there exists a real-to-sim and sim-to-real gaps, it is sufficiently small for this type of task, even when considering pose estimation errors that affect object localization in both the real and simulated environments.

From the \autoref{fig:pmbs-case1} and \autoref{fig:pmbs-case2}, in the first row and first column, we capture an image of the real-world scene, perform pose estimation, and reconstruct the scene in the simulator (first column, second row). Planning is conducted within the simulator, where the estimated push action and its resulting state are shown in the first-row images. This process is iteratively repeated until the target object is successfully grasped. The discrepancy between the third-row images and the first-row images in the next column illustrates the difference between the planned/estimated results and the actual execution results of the real robot.

The simulator is capable of providing highly accurate physics simulations. However, in some cases, it may yield non-accurate yet reasonable physics approximations, which are still useful for task execution.

\begin{figure}[ht!]
    \centering
    \includegraphics[width=\linewidth]{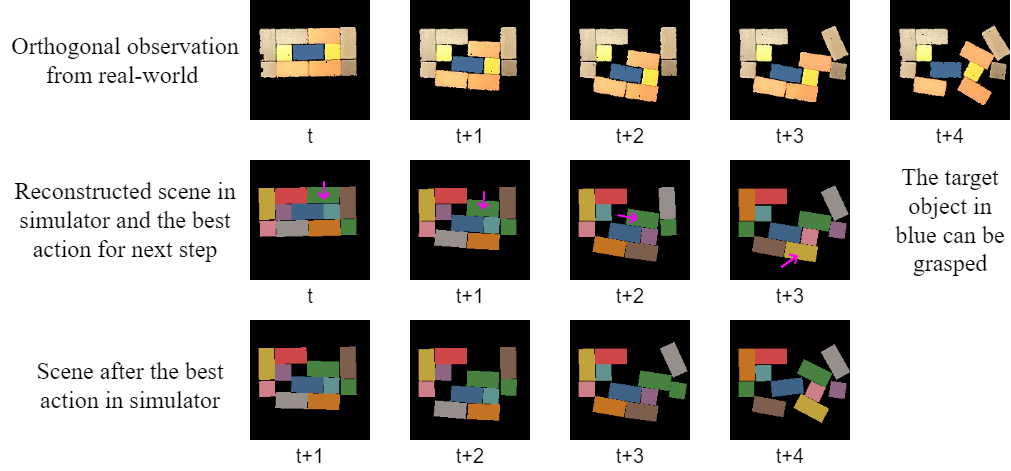}
    \caption{\label{fig:pmbs-case1}
        Case study one of real to sim to real gap. Simulator provides accurate physics simulations.
    } 
\end{figure}

\begin{figure}[ht!]
    \centering
    \includegraphics[width=\linewidth]{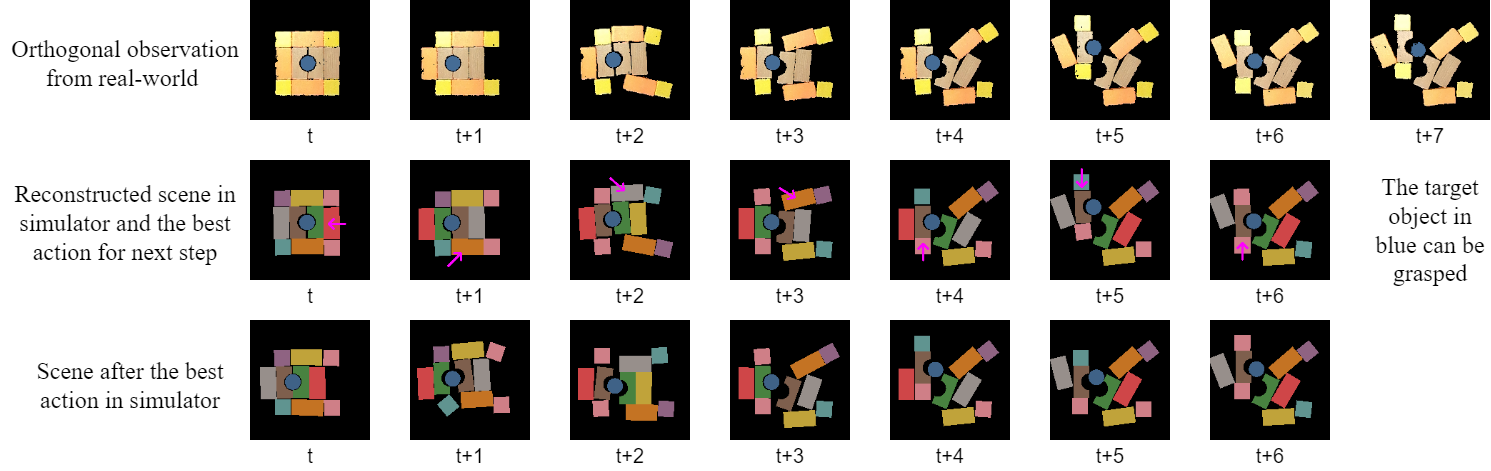}
    \caption{\label{fig:pmbs-case2}
        Case study two of real to sim to real gap. Simulator provide non-accurate but reasonable physics simulations.
    } 
\end{figure}

    \chapter{Chapter 7 - REMP Supplementary}
    \thispagestyle{myheadings}

\begin{figure}[ht!]
    \centering
    \includegraphics[width=0.6\linewidth]{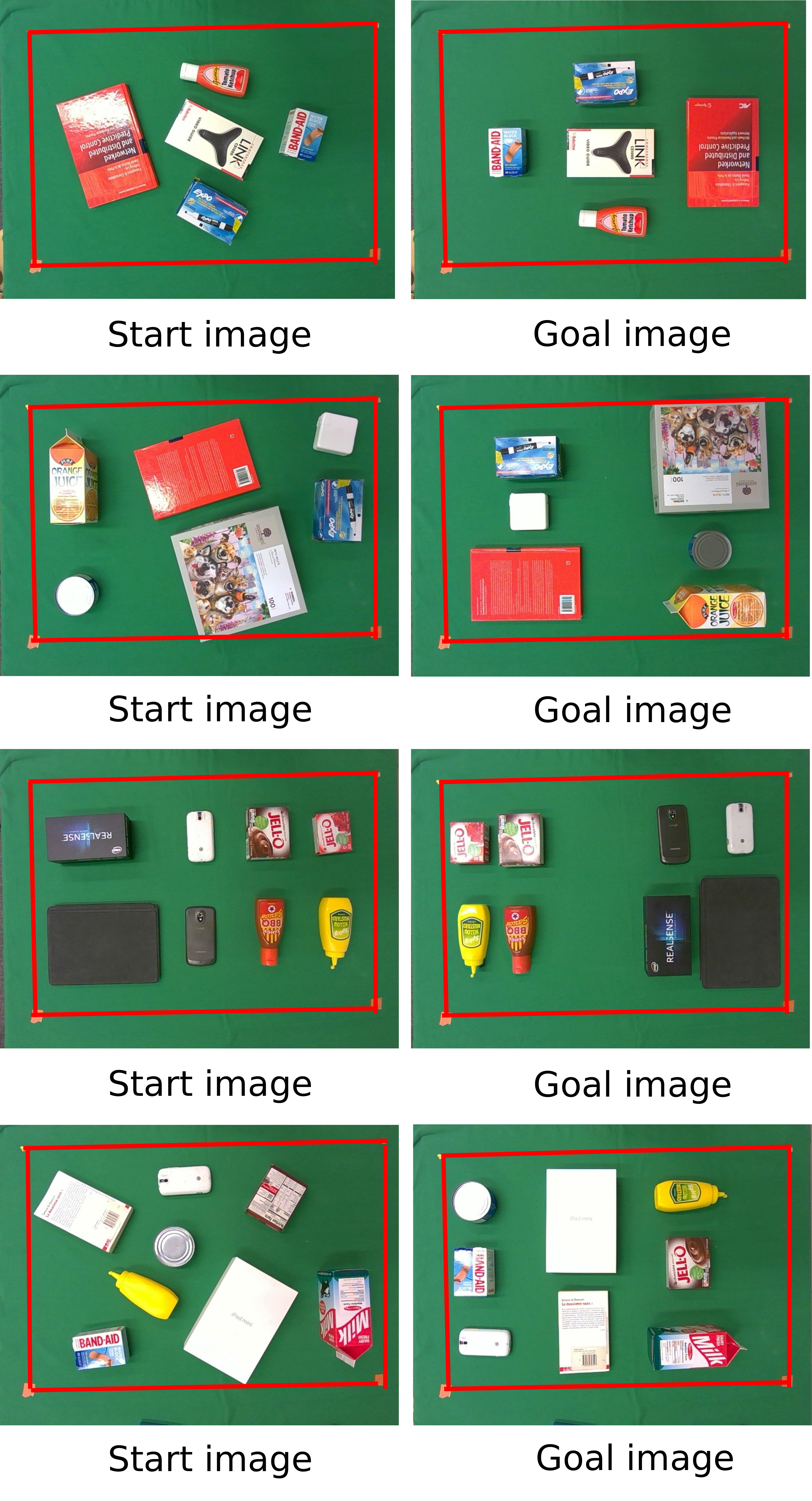}
    \caption{\label{fig:remp-real}
        Some cases in real world setup.
    } 
\end{figure}
    
\end{theappendices}
    \begin{publications}
\thispagestyle{myheadings}

%% Please use \labelcref instead of \ref when refer to your publications!

\begin{enumerate}[leftmargin=*,start=1,label=\textbf{P\arabic*}, ref={P\arabic*}]
    \item \label{pubs:huang2021dipn} 
    \textbf{Baichuan Huang}, Shuai D. Han, Abdeslam Boularias, and Jingjin Yu. 
    "Dipn: Deep interaction prediction network with application to clutter removal." 
    In \textit{2021 IEEE International Conference on Robotics and Automation (ICRA)}, 
    pp. 4694--4701, 2021. IEEE.
    
    \item \label{pubs:huang2021visual} 
    \textbf{Baichuan Huang}, Shuai D. Han, Jingjin Yu, and Abdeslam Boularias. 
    "Visual foresight trees for object retrieval from clutter with nonprehensile rearrangement." 
    \textit{IEEE Robotics and Automation Letters}, 
    vol. 7, no. 1, pp. 231--238, 2021. IEEE.
    
    \item \label{pubs:huang2022interleaving} 
    \textbf{Baichuan Huang}, Teng Guo, Abdeslam Boularias, and Jingjin Yu. 
    "Interleaving monte carlo tree search and self-supervised learning for object retrieval in clutter." 
    In \textit{2022 International Conference on Robotics and Automation (ICRA)}, 
    pp. 625--632, 2022. IEEE.

    \item \label{pubs:huang2022parallel} 
    \textbf{Baichuan Huang}, Abdeslam Boularias, and Jingjin Yu. 
    "Parallel monte carlo tree search with batched rigid-body simulations for speeding up long-horizon episodic robot planning." 
    In \textit{2022 IEEE/RSJ International Conference on Intelligent Robots and Systems (IROS)}, 
    pp. 1153--1160, 2022. IEEE.

    \item \label{pubs:huang2024toward} 
    \textbf{Baichuan Huang}, Xujia Zhang, and Jingjin Yu. 
    "Toward optimal tabletop rearrangement with multiple manipulation primitives." 
    In \textit{2024 IEEE International Conference on Robotics and Automation (ICRA)}, 
    pp. 10860--10866, 2024. IEEE.

    \item \label{pubs:huang2023earl} 
    \textbf{Baichuan Huang}, Jingjin Yu, and Siddarth Jain. 
    "EARL: Eye-on-hand reinforcement learner for dynamic grasping with active pose estimation." 
    In \textit{2023 IEEE/RSJ International Conference on Intelligent Robots and Systems (IROS)}, 
    pp. 2963--2970, 2023. IEEE. This work, while part of the author's PhD research efforts, is not discussed in detail within this dissertation.
\end{enumerate}

\end{publications}

    \makeBibliography
\end{thesisbody}

\end{document}